\documentclass[review]{elsarticle}

\usepackage{natbib}
\usepackage{geometry}
\usepackage{fleqn}
\usepackage{graphicx}
\usepackage{caption}
\usepackage{subcaption}
\usepackage{enumerate}
\usepackage{amssymb}
\usepackage[utf8]{inputenc}
\usepackage{algorithm}
\usepackage{algpseudocode}

\begin{document}
\let\today\relax
\begin{frontmatter}

\title{ Three-dimensional planar model estimation using multi-constraint knowledge based on k-means and RANSAC\tnoteref{t1}}
\tnotetext[t1]{This study was supported in part by the University of Alicante, Valencian Government and Spanish government under grants GRE11-01, GV/2013/005 and DPI2013-40534-R.}

\author[alc]{Marcelo Saval-Calvo\corref{cor2}}
\ead{msaval@dtic.ua.es}

\author[alc]{Jorge Azorin-Lopez\corref{cor1}}
\ead{jazorin@dtic.ua.es}

\author[alc]{Andres Fuster-Guillo\corref{cor1}}
\ead{fuster@dtic.ua.es}

\author[alc]{Jose Garcia-Rodriguez \corref{cor1}}
\ead{jgarcia@dtic.ua.es}

\cortext[cor1]{Corresponding author}
\cortext[cor2]{Principal corresponding author}

\address[alc]{University of Alicante, Department of Computer Technology, \\ C/ San Vicente - Alicante s/n
03690, San Vicente del Raspeig (Alicante)}

\begin{abstract}
Plane model extraction from three-dimensional point clouds is a necessary step in many different applications such as planar object reconstruction, indoor mapping and indoor localization. Different RANdom SAmple Consensus (RANSAC)-based methods have been proposed for this purpose in recent years. In this study, we propose a novel method-based on RANSAC called Multiplane Model Estimation, which can estimate multiple plane models simultaneously from a noisy point cloud using the knowledge extracted from a scene (or an object) in order to reconstruct it accurately. This method comprises two steps: first, it clusters the data into planar faces that preserve some constraints defined by knowledge related to the object (e.g., the angles between faces); and second, the models of the planes are estimated based on these data using a novel multi-constraint RANSAC. We performed experiments in the clustering and RANSAC stages, which showed that the proposed method performed better than state-of-the-art methods.
\end{abstract}


\end{frontmatter}

\section{Introduction}
Many different applications of three-dimensional (3D) computer vision (such as scene registration and reconstruction, localization and mapping) use large volumes of data to extract a model. In particular, plane estimation is a common problem in various applications, including planar object reconstruction, mapping using planar surfaces to localize a robot in an environment and minimizing the effects of noisy data. Recently, new low-cost RGB-D sensors have been developed to provide colour and depth information simultaneously (e.g., Microsoft Kinect, PrimeSense Carmine and Asus Xtion). In general, these sensors estimate the depth using structured light techniques, specifically a speckle pattern in the infrared spectrum. At present, these sensors are being used in various research projects \cite{Morell2014} because of their affordability, and they have also opened up new areas of study and application. However, the data obtained from these sensors tend to be noisy, thereby resulting in errors in the results achieved by these processes in many cases. Thus, techniques need to be developed to minimize the effects of noise. 

One of the most popular methods for estimating a model from a set of 3D data is the RANdom SAmple Consensus (RANSAC). This method or paradigm was proposed by Fischler and Bolles in \cite{Fischler1981} for fitting a model to experimental data by selecting a random subset of input values, estimating a model from these values and then evaluating the quality of the model based on the overall dataset. This process is applied iteratively until convergence, or for a predefined number of iterations. Finally, the best model is selected (i.e., that with best fit to the overall point set). 

One of the most important advantages of RANSAC is that it is not constrained to a specific dimensionality. For example, it has been used in various situations such as two-dimensional (2D) line fitting \cite{Aly2008}\cite{Borkar2012} for smart cars to estimate the directions of road markings and to guide the car between them. Another common use is for removing outliers during image matching. Raguram et al. \cite{Raguram2008} reviewed the use of RANSAC variants for matching outlier removal in 2D features. RANSAC was also applied to 3D registration by Su et al. \cite{Su2013} and Henry et al. \cite{Henry2014}. Furthermore, Hassner et al. \cite{Hassner2013} employed RANSAC to align 3D data with 2D images. Zhou et al. \cite{Zhou2013} used RANSAC to obtain camera parameters by removing outliers during the calibration process. 

However, the RANSAC algorithm has a random step that affects repeatability. Thus, Hast et al. \cite{Hast2013} proposed a variation of RANSAC by introducing a recalculation hypothesis step where only the inliers of the previous step are used, thereby allowing the method to find the optimal set even with a high percentage of outliers.

The use of RANSAC has been studied widely for plane model fitting. Tarsha-Kurdi et al. \cite{Tarsha-kurdi2007} compared the use of the Hough transform and RANSAC for 3D roof plane estimation. Kim et al. \cite{Kim2012} used RANSAC for plane detection during stereo matching in scene reconstruction. Mufti et al. \cite{Mufti2012} focused on finding planes (walls and the ground plane) in moving scenarios during autonomous navigation, where they used spatio-temporal information to evaluate the consensus set and to estimate the probability of a point belonging to a specific plane.

In many cases, multiple planes have to be estimated in a scene. Gallup et al. \cite{Gallup2010} extracted planar and non-planar regions using iterative RANSAC, where they selected a random set of points, estimated a plane using RANSAC, obtained the data that fit this plane and then repeated the process with the remaining data until the remaining points could not be used to find a plane. Zuliani et al. \cite{Zuliani2005} presented multiRANSAC as a modification of the traditional RANSAC for multiple models, where the data that fitted the models were estimated using disjoint consensus sets. Schnabel et al. \cite{Schnabel2007} proposed the use of multiple primitives in order to find the maximum number of inliers that fit different primitives (planes, spheres, cylinders, cones and tori) in an iterative manner. Sinha et al. \cite{Sinha2009} estimated planes used vanishing points and 3D line segments estimated from several views of a scene. RANSAC was applied to the remaining points that had not been assigned previously to any model. Isack and Boykov \cite{Isack2012} proposed a method called PEARL that combines model sampling with the iterative re-estimation of inliers and model parameters. This method does not employ a predefined number of models because it uses traditional RANSAC for their initialization.

Three previous studies are highly related to the method proposed in the present study: \cite{Gallo2011}, \cite{Zhou2011} and \cite{Zhou2013a}. These approaches pre-cluster the data in order to find better planes, instead of simply selecting random data and trying to find the best match for a planar model. The first method is a CC-RANSAC variant for fitting multiple surfaces, which employs pre-clustered data to allow the RANSAC algorithm to obtain better results. This method assumes ground plane situations with steps, curbs or ramps. In these situations, the traditional method would fail to obtain a single plane if it crosses two or more patches. In order to avoid this problem, the CC-RANSAC variant method pre-clusters the data using the connected 8-neighbours components. The second and third methods, i.e., \cite{Zhou2011} and \cite{Zhou2013a}, improve the previous method by adding vector normal information to allow the estimation of each cluster in the clustering and patch-joining steps. In our proposed method, we also add scene knowledge to improve the results, where we add constraints in the clustering stage as well as during RANSAC model fitting. This improves the results greatly when the signal-to-noise ratio is low, whereas other methods fail to cluster correctly and thus the RANSAC stage also fails. 

Other techniques have been proposed for model fitting, e.g., Mercer Kernel statistical learning techniques were used to estimate model fitting by Chin et al. \cite{Chin2009}, while Zhou et al. \cite{Zhou2010} used particle swarm optimization for multiple model fitting.

After considering the state-of-the art methods for model estimation from a point cloud, it clear that a challenging problem still remains when high accuracy is required, mainly with noisy data, including that captured using RGB-D sensors. In this study, we propose a method for estimating the planes that best fit the model of a planar object, which we call multiplane model estimation (MME). The inputs comprise a 3D set of points and a group of constraints, an initial clustering (a point cloud clustering, PCC) is performed using a new variant of k-means based on the knowledge extracted from the model. Next, a variant of RANSAC with multiple constraints (MC-RANSAC) is applied to estimate the best planes that fit each cluster and that preserve the object model.

The remainder of this paper is organized as follows. In Section~\ref{sec:MME}, we explain the method, where Subsection~\ref{sec:PCC} describes PCC clustering and Subsection~\ref{sec:MCRANSAC} presents the MC-RANSAC approach. In Section~\ref{sec:exp}, we present the results of experimental evaluations of the PCC (Subsection~\ref{sec:exp:PCC}) and MC-RANSAC (Subsection~\ref{sec:exp:MCRANSAC}) methods. Finally, we give our conclusions, including descriptions of the main advantages of the proposed method and the most interesting results obtained.

\section{Multiplane Model Estimation}\label{sec:MME}
\begin{figure}[h!]
  \centering
    \includegraphics[width=\textwidth]{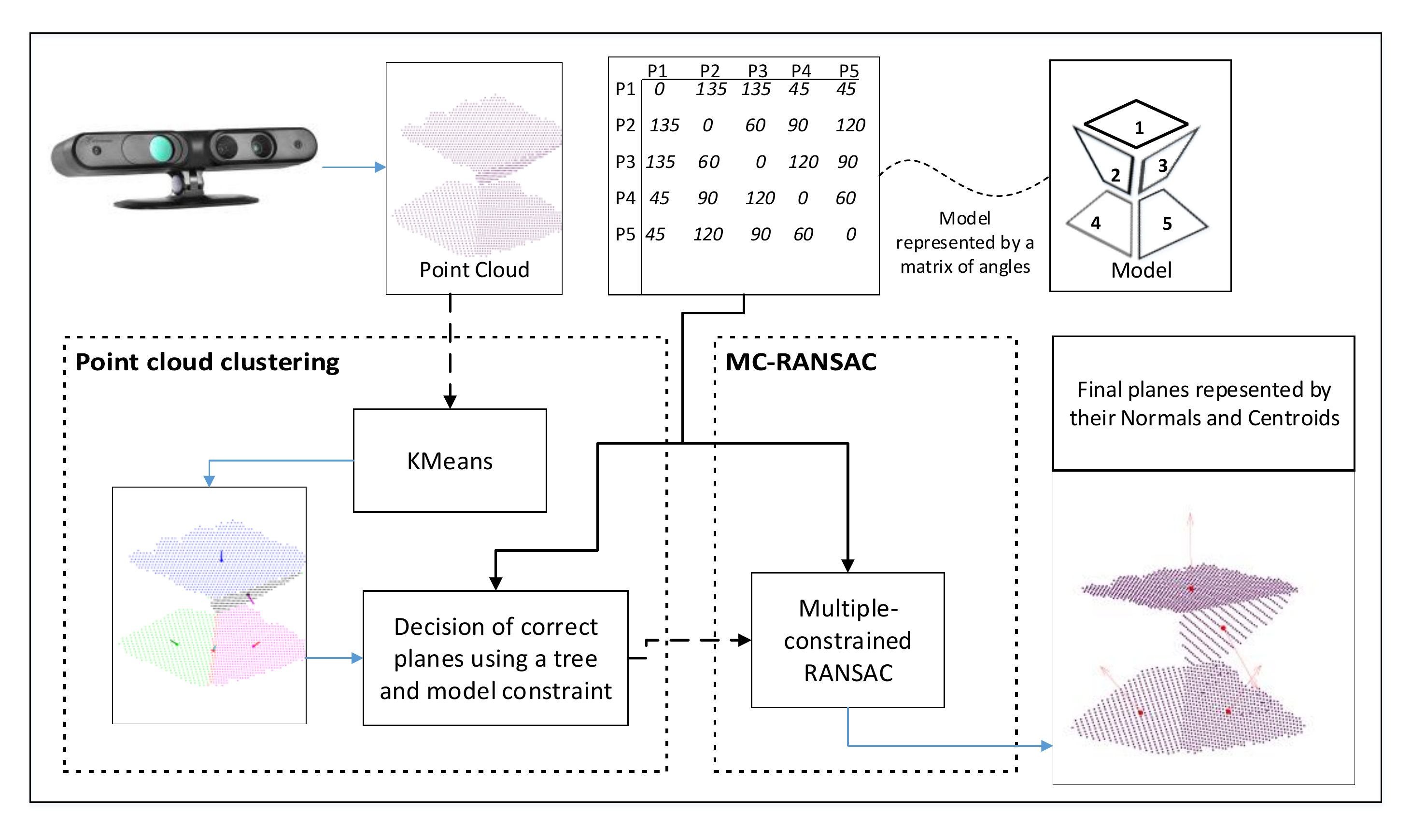}
  \caption{General overview of the process. There are two inputs: a point cloud and a group of constraints. The model is represented using a matrix of angles, which are the constraints. First, the points are clustered and the correct clusters are detected. Next, a MC-RANSAC extracts the planes of the remaining points.}
  \label{fig:MME:scheme}
\end{figure}

In this section, we explain the proposed MME method in detail. Using 3D spatial data as the inputs, the MME method employs a group of constraints to estimate the planes that best fit the 3D point clouds. This method is divided into two main steps: PCC based on the k-means algorithm and a search tree (see Section~\ref{sec:PCC}); and a proposed variant of the RANSAC method called MC-RANSAC (see Section~\ref{sec:MCRANSAC}), which estimates the models from clustered points. Figure \ref{fig:MME:scheme} shows a diagram of the MME, with the PCC and MC-RANSAC steps. The hyphened arrows represent real points transferred between steps and the thick arrows are constraints in the model. The algorithm has two inputs: a point cloud and the model constraints. It is important to stress that the model shown in the figure is actually represented in the method by the angles between each pair of planes (e.g., Table~\ref{tab:MME:modmat}). The point cloud is clustered using traditional k-means based on point and normal vector information, where the clusters are evaluated using the model constraints. The points classified as correct clusters are sent to MC-RANSAC to finally estimate the planes again using the model constraints. 

Finally, it is important to note that the MME method is suitable for N-dimensions because k-means and RANSAC can work in N-dimensions. In this study, we focus on 3D planes in depth, but the general method is explained in Section~\ref{sec:MCRANSAC}.

\subsection{Point Cloud Clustering}\label{sec:PCC}

\begin{figure}[h!]
  \centering
    \includegraphics[width=0.8\textwidth]{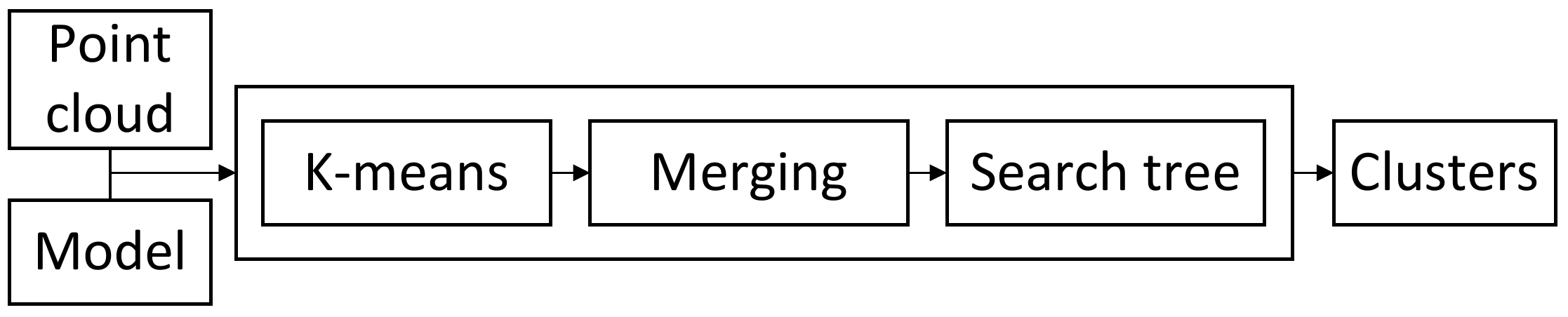}
  \caption{PCC process scheme.}
  \label{fig:PCC:scheme}
\end{figure}

In this subsection, we explain the PCC step of the MME method. A short summary is provided in Fig.~\ref{fig:PCC:scheme}, which shows that PCC employs a raw 3D point cloud and a table of constraints on the model (e.g., Table~\ref{tab:MME:modmat}) as inputs. 

The full algorithm is summarized briefly in Algorithm~\ref{alg:MME}.

\begin{algorithm}[H]
\caption{General PCC algorithm. First, k-means estimates the clusters based on the points and normals. Next, the planes with similar orientations are joined. The clusters that obtain the best solutions corresponding to the constraints are calculated using a tree search technique. }
\label{alg:MME}
\algblock{If}{EndIf}
\algcblock[If]{If}{ElsIf}{EndIf}
\algcblock{If}{Else}{EndIf}

\begin{algorithmic}[1]
\State input: \{Point cloud; model constraints\}
\State output: \{Point cloud clusters that meet the constraints\}
\State ClustersAux = k-means(points,normals,numClusters);
 \For{each cluster idC1 and idC2 in ClustersAux}
 	\If{ClustersAux(idC1)$\approx$ClustersAux(idC2)}
\State 		ClustersAux(idC1) = ClustersAux(idC2);
 	\EndIf
 \EndFor
\State Calculate angle correspondence matrices A and B;
 \For{each plane in B}
\State 	Estimate similar planes in A using Eq.~\ref{fml:similar}
 \EndFor
 \While{ possible solutions remain}
\State	read current candidate for the level;
	\If{new node meets the constraints}
\State		go to next level;
		\If{is the last level}
\State			PossibleSolutions.add(branch);
		\EndIf
	\Else
\State		go back to the previous level and change the candidate;
	\EndIf
 \EndWhile
 
 \end{algorithmic}
\end{algorithm}

The method estimates the clusters that best fit the model in three main steps. First, a k-means algorithm estimates the clusters based on the 3D points and normal vectors. Second, similar planes (e.g., similar angles, distances and colours) are merged, where we used the plane orientation in the present study. Finally, a tree search technique finds the best cluster. The best solution is used as the input for the next step (as explained in Subsection~\ref{sec:MCRANSAC}). 

For each point in the point cloud, the normal vector is estimated using its neighbourhood. The size of the neighbourhood affects the homogeneity of the orientation of the normals in a plane. The normals will be smooth for a large neighbourhood (i.e., all of the normals in a plane will point in the same direction), but the edges of the objects will also be smoothed, and thus they are less descriptive. By contrast, if the neighbourhood is small, the normals will be affected greatly by noise and they will be less uniform for a single plane surface. In the present study, we selected experimental values to obtain a smoothed result that retained sharp edges (i.e., the number of neighbours was seven or 11 points). 

We normalize the 3D points and normals within the range [-1,1] to equalize the weights in both cases in the proposed method. Next, we apply the well-known k-means algorithm to obtain an initial clustering that considers the spatial location and orientation. This method needs a predefined number of clusters; thus, at least the maximum number of planes visible from a single point of view should be used by k-means, although we suggest the use of a higher number in order to manage noise. For example, if we are looking for a cube, the maximum number of planes from a single view will be three (top or bottom and two sides), and thus we use $3 + \lceil3*perc\rceil$, where this percentage, $perc$, $0.4$, was estimated experimentally in our study.

After we obtain the initial clusters, it is necessary to merge and/or reject some of them. The number of clusters initialized is higher than the maximum number of planes, so it might appear that there are various clusters where there should only be one plane, or the planes may be incorrect due to noise. To merge the clusters, we evaluate the similarity of the orientations of the normals of the clusters, which corresponds to the average of all the normals that belong to each cluster. The clusters are joined if the angle is below a threshold. Centroids are not used in this step because they can produce incorrect solutions. For example, if two small clusters are close to an edge on both sides, they are very close each other, but they do not have to be joined.

The next step is the most critical feature of this part of the algorithm. After merging, we might still have incorrect clusters, which must be rejected using the model provided by the table of constraints. Thus, we propose the use of model constraints and a tree search technique to determine which of the clusters best represents the model planes. First, we explain the model constraints. The model is defined by the relationship between the planes related to the object, e.g., distances, sizes or colours. In this study, we propose the use of the angles among the planes to describe the model of an object, where two angle correspondence matrices are used to represent the angles between each pair of planes: the first corresponds to the model constraints ($A_{nxn}$, Table~\ref{tab:MME:modmat}, where $n$ is the maximum number of planes visible in a single view of the model) and the other represents the angles between the previously estimated clusters ($B_{mxm}$, Table~\ref{tab:MME:objmat}, where $m$ is the number of clusters). We refer to these as the Model A and \textit{Object} B. Matrix A must contain the constraints for the maximum number of possible visible planes, which makes the algorithm invariant to position and scale.  

\begin{table}[h!]
\centering
\begin{tabular}{|c||c|c|c|c|c|}
\hline
	A & $PA_1$ & $PA_2$ & $PA_3$ & \ldots & $PA_N$ \\
\hline
\hline
	$PA_1$ & 0  & $\alpha_{(1,2)}$ & $\alpha_{(1,3)}$ & \ldots  & $\alpha_{(1,n)}$ \\
    $PA_2$ &      $\alpha_{(2,1)}$  & 0 & $\alpha_{(2,3)}$ & \ldots  & $\alpha_{(2,n)}$ \\
    $PA_3$ &      $\alpha_{(2,1)}$  & $\alpha_{(3,2)}$ & 0 & \ldots  & $\alpha_{(2,n)}$ \\
    \vdots & \vdots  & \vdots & \vdots  & \vdots  & \vdots \\
    $PA_n$ & $\alpha_{(n,1)}$  & $\alpha_{(n,2)}$ & $\alpha_{(n,3)}$ & \ldots  & 0 \\
\hline
\end{tabular}
\caption{Model angle correspondence matrix.}
\label{tab:MME:modmat}
\end{table}

\begin{table}[h!]
\centering
\begin{tabular}{|c||c|c|c|c|c|}
\hline
	B & $PB_1$ & $PB_2$ & $PB_3$ & \ldots & $PB_M$ \\
\hline
\hline
	$PB_1$ & 0  & $\beta_{(1,2)}$ & $\beta_{(1,3)}$ & \ldots  & $\beta_{(1,m)}$ \\
    $PB_2$ &      $\beta_{(2,1)}$ & 0 & $\beta_{(2,3)}$ & \ldots  & $\beta_{(2,m)}$ \\
    $PB_3$ &      $\beta_{(3,1)}$ & $\beta_{(3,2)}$ & 0 & \ldots  & $\beta_{(3,m)}$ \\
    \vdots & \vdots  & \vdots & \vdots  & \vdots  & \vdots \\
    $PB_m$ & $\beta_{(m,1)}$  & $\beta_{(m,2)}$ &$\beta_{(m,3)}$ & \ldots & 0 \\
\hline
\end{tabular}
\caption{Object angle correspondence matrix.}
\label{tab:MME:objmat}
\end{table}

In order to evaluate the clusters that describe the planes and those that should be rejected, different situations can be handled by evaluating the correspondences between A and B, as follows.
\begin{itemize}
	\item A and B are not necessarily aligned because $PA_i$ in A does not correspond to $PB_i$ in B.
	\item A and B could have a different number of planes because A has the maximum visible planes whereas B may be a view with fewer planes. For example, a cube would have A with $n = 3$ but if the current view is from the front, B could be $m = 2$ for the front and top sides.
	\item B could have incorrect planes due to noise effects.
\end{itemize}

\begin{figure}[H]
  \centering
    \includegraphics[width=0.7\textwidth]{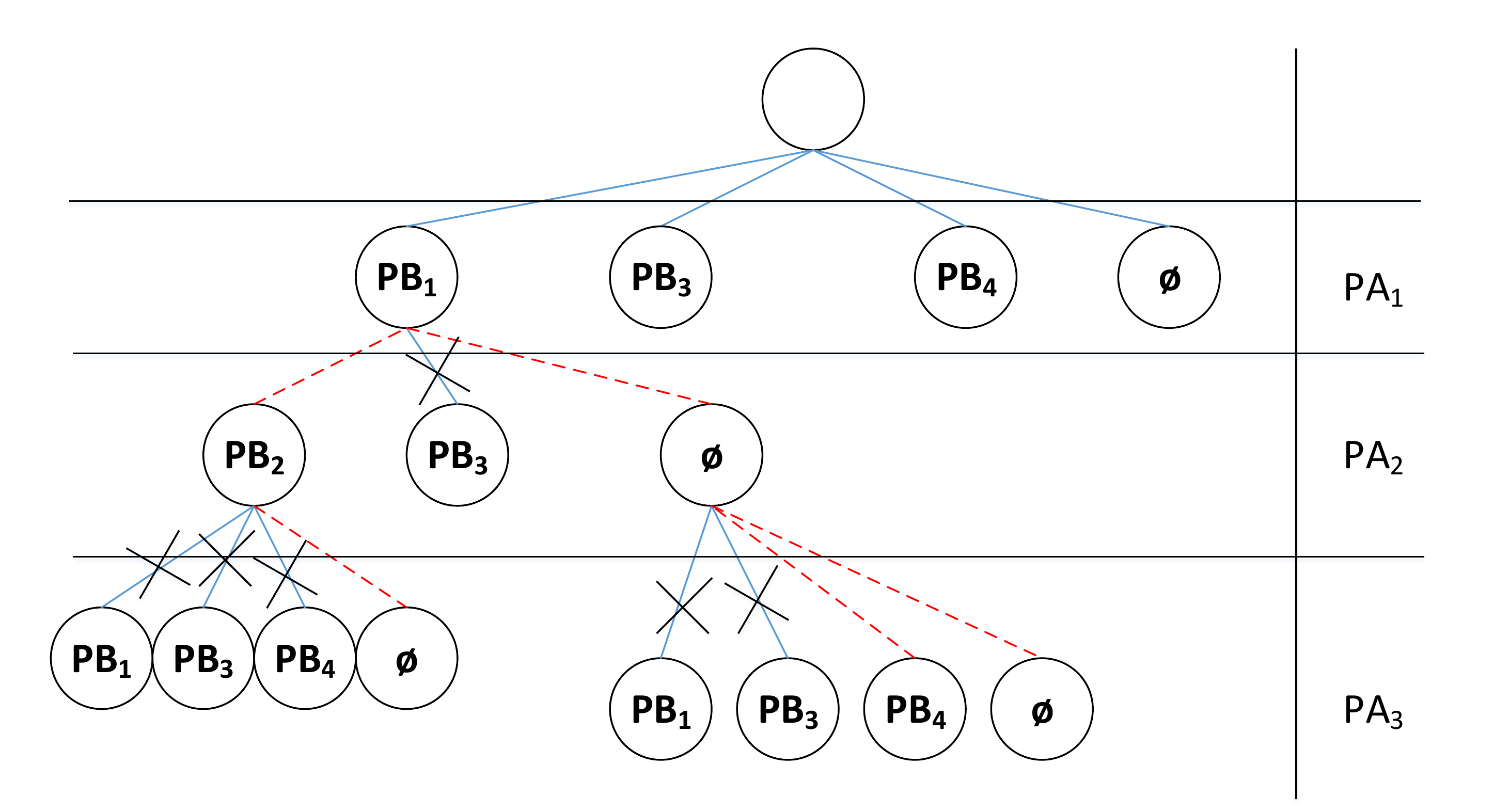}
  \caption{A partial tree for the example considered. The hyphened lines represent a good subset of planes, which preserve the constraints.}
  \label{fig:MME:tree}
\end{figure}

To consider all of these situations, we need to identify the best group of clusters in B that fits A using a tree search technique. The tree of solutions (Figure \ref{fig:MME:tree}) is obtained with $n$ levels, where each belongs to a plane in the model. The nodes at every level correspond to every possible value in B. The number of branches generated by this tree is $n^m$, so it is necessary to use reduction techniques. Thus, we propose to evaluate the most similar planes from B to A, which involves finding the clustered planes that are most similar to those in the model based on comparisons of their angles, where we apply Eq.~\ref{fml:similar} to each plane in B for this purpose:

\begin{equation}\label{fml:similar}
Similar(PB_x) = max(\sum^{N}_{y=1}count(PB_x \approx PA_y))
\end{equation}
where $count(PB_x \approx PA_y)$ is the number of values (angles) in $PA_y$ that differ by less than a threshold $PB_x$. For example, using Table~\ref{tab:MME:exobjmat}, we may assume a threshold of 5 (which means that any difference in the angles that exceeds this value will not be considered), and thus the plane $PB_1$ will have two similar values with $PA_1$, one with $PA_2$, and two with $PA_3$. Next, we keep the maximum values, so $PB_1$ could be related to $PA_1$ or $PA_3$, as found in the model. Using this formula, we obtain a table of possible correspondences between A and B, which maximize the likelihood of $Object$ and $Model$, thereby reducing the solution space. Table~\ref{tab:MME:exsimil} shows the results obtained for the example in Table~\ref{tab:MME:exobjmat}.

\begin{table}[H]
\centering
\begin{tabular}{c c}

\begin{tabular}{|c||c|c|c|}
\hline
	A & $PA_1$ & $PA_2$ & $PA_3$ \\
\hline
\hline
	$PA_1$ & 0 & 45 & 90  \\
    $PA_2$ & 45 & 0 & 45 \\
    $PA_3$ & 90 & 45 & 0 \\
\hline
\end{tabular}
&
\begin{tabular}{|c||c|c|c|c|}
\hline
	B & $PB_1$ & $PB_2$ & $PB_3$ & $PB_4$\\
\hline
\hline
	$PB_1$ & 0  & 44 & 70 & 91  \\
    $PB_2$ & 44 & 0  & 46 & 73  \\
    $PB_3$ & 70 & 43 & 0  & 80  \\
    $PB_4$ & 91 & 73 & 80 & 0   \\
\hline
\end{tabular}
\end{tabular}
\caption{Example of Model and Object angle correspondence matrices. The left table shows the angles between the planes in the model and the right table presents the angles between the planes of the clusters. In this case, B has more planes than A. Moreover, the best result includes $PB_1$ and $PB_2$, which correspond to $PA_1$ and $PA_2$, respectively.}
\label{tab:MME:exobjmat}
\end{table}

\begin{table}[H]
\begin{center}
\begin{tabular}{|c|c|c|c|}
\hline
	$PB_1$ & $PB_2$ & $PB_3$ & $PB_4$ \\
\hline
\hline
	$PA_1$ & $PA_2$ & $PA_1$ & $PA_1$   \\
    $PA_3$ &        & $PA_2$ & $PA_3$  \\
           &        & $PA_3$ &   \\
\hline
\end{tabular}\caption{Possible correspondences between planes in A and B.}\label{tab:MME:exsimil}
\end{center}
\end{table}

Table~\ref{tab:MME:exobjmat} shows the $Model$ and $Object$ angle correspondence matrices, which demonstrates that there are more planes in the $Object$, $PB_{1..4}$ than $PA_{1..3}$. Using Eq.~\ref{fml:similar}, the search space is reduced to obtain Table~\ref{tab:MME:exsimil}. Finally, the best results that maximize the number of correct planes in B will be $PB_1$ and $PB_2$, which correspond to $PA_1$ and $PA_2$, respectively. In a correct combination, \textit{Model} and \textit{Object} are similar in terms of their constraint angles. Thus, the angle difference between the selected plane in B and the corresponding plane in A is equal to or below a threshold (in this case, that used in Eq.~\ref{fml:similar}).


After this reduction of the search space, it is not always possible to obtain the response within a reasonable time. Thus, we evaluate the branch while the nodes are being inserted into the tree by using a backtracking algorithm to detect incoherent values at an early stage. This evaluation is performed by comparing the angles of the new node with all the predecessors in the tree. If the values in $Object$ ($B$ matrix) are similar to those in the $Model$ matrix, the new node is accepted and inserted. When all of the levels are filled, the branch is selected as a possible candidate among a good set of planes. Not all of the planes in B correspond to the planes in A, so it is necessary to add an empty value to the decision tree (Figure~\ref{fig:MME:tree}), which allows the algorithm to handle incorrect and missing planes. Finally, if more than one solution is obtained, the solution where more points are added to all clusters in the branch is returned. For the example shown in Figure~\ref{fig:MME:tree}, $(PB_1 \cup PB_2 \cup \O)$, $(PB_1 \cup \O \cup PB_4)$, and $(PB_1 \cup \O \cup \O) $ are possible solutions. If we assume that cluster $PB_1$ has 100 points, cluster $PB_2$ has 200 points and cluster $PB_4$ has 100 points, then the result will be $(PB_1 \cup PB_2 \cup \O)$ because the total number of points added to the cluster for that branch will be 300. 

\subsection{Multi-Constraint RANSAC} \label{sec:MCRANSAC}

\begin{figure}[h!]
  \centering
    \includegraphics[width=0.8\textwidth]{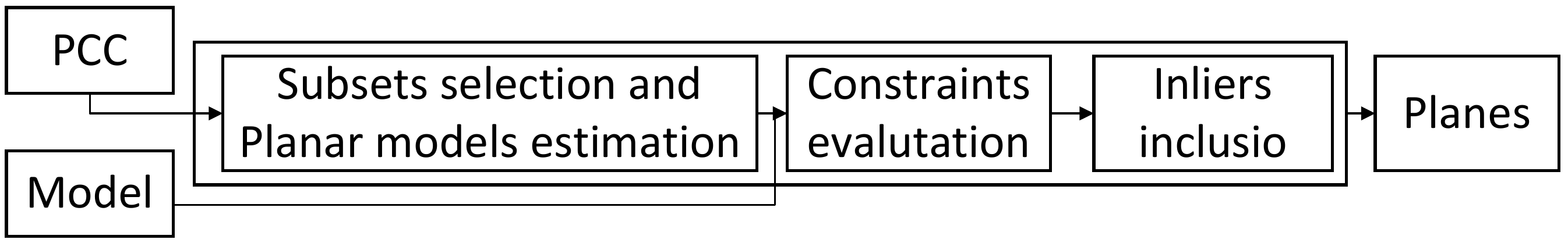}
  \caption{MC-RANSAC process scheme.}
  \label{fig:MCRANSAC:scheme}
\end{figure}

In this subsection, we present the proposed MC-RANSAC method for estimating the planes that best fit the input data using prior knowledge related to the object model. This method is based on the original RANSAC paradigm and it also employs the idea proposed in \cite{Gallo2011} of pre-clustering the input data. In the previous section, we proposed a method for this purpose, but any clustering method can be used because MC-RANSAC is independent of the pre-clustering process. Three main steps are considered (see Figure~\ref{fig:MCRANSAC:scheme}), including the two main steps in the general RANSAC method: estimating the model using a random subset of data and evaluating the inliers for the remaining data. In our method, there are several input data groups, i.e., one for each cluster, so one subset is calculated for each group of data and the evaluation of inliers is also applied to each group. Furthermore, a middle step is introduced where the constraints are evaluated. We assume that the union of all planes yields a globally correct figure, i.e., there are no incorrect groups. Therefore, if all of the constraints are not satisfied, the whole method returns an error status. This may mean that is not possible to obtain this model using these data or that the pre-clustering results are incorrect.

In this study, the method is evaluated using 3D planes where the input data comprise 3D points and the groups are the previously calculated clusters (see subsection \ref{sec:PCC}). The incorrect groups are rejected in the PCC step, so we do not have to handle incorrect planes in MC-RANSAC. 

The MC-RANSAC input comprises a set of clustered data and the model constraints that correspond to the clusters. For the example shown in Tables \ref{tab:MME:exobjmat} and \ref{tab:MME:exsimil}, we only use $PA_1$ and $PA_2$ columns and rows in the matrices. The first step involves selecting a subset of data from each group, where the mathematical model of each individual subset is estimated. In the specific case of 3D planes, least squares estimation can be employed to extract the models.

After estimating the models, we propose to introduce a step where the constraints are evaluated. These constraints are the angles in the model matrix that correspond to the input clusters because we need to find the planes that fit the original model. The constraint between each pair of planar models (not to be confused with the model of constraints) is evaluated. If the angles describe the constraints, we compare the angle between each pair of planes with the corresponding model angle. If one does not agree, this group of planes is rejected and the algorithm starts again. It is not always possible to achieve a perfect fit for the constraints, so a threshold for the allowed deviation from the perfect fit is introduced, which determines the accuracy of the results.

The final step of the MC-RANSAC is the evaluation of inliers. The planar models are extracted from a subset of points, so the remainder are evaluated as inliers, where an iterative method is applied. For each group, every point that is not used in the initial subset is incorporated into the inlier subset. The model for this new subset is then calculated and the constraints are tested again. If the model does not comply with the restrictions, the point is removed as an outlier. It should be noted that this step is the most time-consuming part of the algorithm because it has to be applied to all of the data in all of the groups. Thus, in order to reduce the temporal complexity, our proposed method only evaluates the minimum number of data per group that are considered sufficient to ensure the quality of the model.

The overall process is iterated several times to change the initial random subsets and to obtain the final models that fit more of the input data while preserving the constraints.

In some cases, as shown in the experimental evaluations later in this study, the PCC clustering process might successfully satisfy the constraints in the previous step, but the points do not actually belong to correct planes. In subsection \ref{sec:exp:PCC}, Figure \ref{fig:exp:k-means_wrong} shows an example of this situation. Therefore, MC-RANSAC must detect any errors and feed back into the MME general method to evaluate whether recalculation of the clusters is necessary.

The full algorithm is summarized briefly in Algorithm~\ref{alg:MCRANSAC}.

\begin{algorithm}[h!]
\caption{General MC-RANSAC algorithm. First, the plane model is calculated for each subset. Next, the constraints defined by the corresponding matrices determine whether the plane models describe the actual model if they are all satisfied. The inliers and outliers in each group are checked by adding each point to the accepted set and recalculating the plane. All of the constraints are tested using this new plane and a point is accepting as an inlier if they are all satisfied. }
\label{alg:MCRANSAC}
\algblock{If}{EndIf}
\algcblock[If]{If}{ElsIf}{EndIf}
\algcblock{If}{Else}{EndIf}

\begin{algorithmic}[1]

\State input: \{Data$\_$in:$Cluster_{1..l}$; Model constraints corresponding to the $Cluster_{1..l}$\}
\State output: \{Model$\_$out:$Cluster_{1..l}$\}
\For{i := each group in Data$\_$in}
	\State InitialSubset(i) := random(Data$\_$in(i));
	\State ModelInitial(i) := estimate the model for InitialSubset(i);
\EndFor
\State Evaluate the constraints(ModelInitial(i=$1..l$), Model constraints);
\If{Constraints OK} 
	\For{i := each group in Data$\_$in}
		\For{j := each data in Data$\_$in(i)}
			\State InitialSubset(i).add(Data$\_$in(i)(j));
			\State ModelInitial(i) := estimate the model for InitialSubset(i);
			\State Evaluate the constraints(ModelInitial(i), Model Constraints);
			\If{NOT Constraints OK}
				\State InitialSubset(i).remove(Data$\_$in(i)(j));
			\EndIf
		\EndFor
	\EndFor
\EndIf

 \end{algorithmic}
\end{algorithm}

%

\section{Experimental}\label{sec:exp}

The experiments used to validate the proposed method comprised three different parts. First, we conducted various experiments in order to evaluate the behaviour of PCC with different levels of noise and different objects. Second, we also evaluated MC-RANSAC with different levels of noise and objects. Finally, we performed a combined analysis of PCC and MC-RANSAC to demonstrate the overall benefits of the MME method. 
\begin{figure}[!t]
\centering
\begin{tabular}[t]{c c}
	\includegraphics[width=0.4\textwidth]{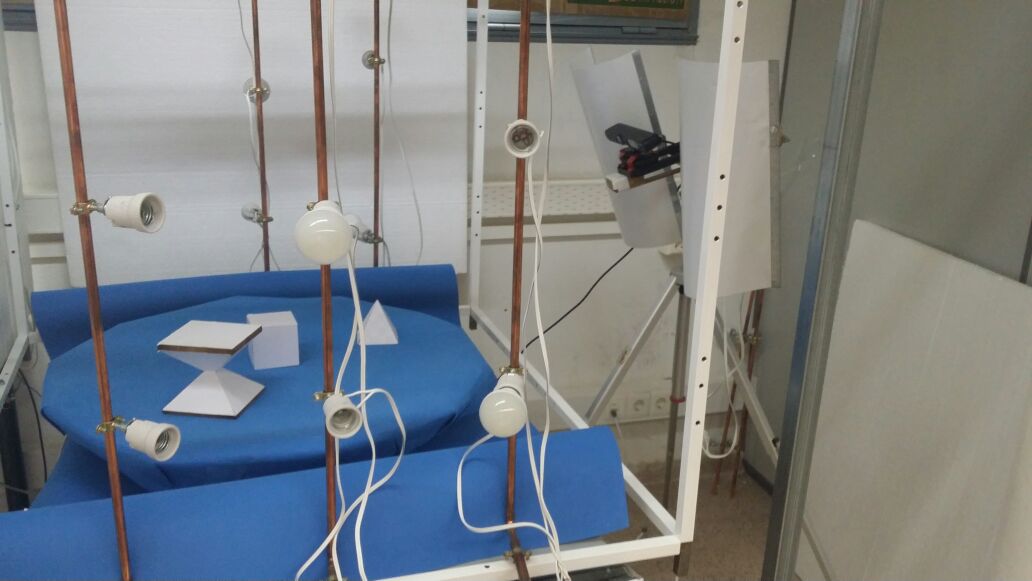}
	& \includegraphics[width=0.4\textwidth]{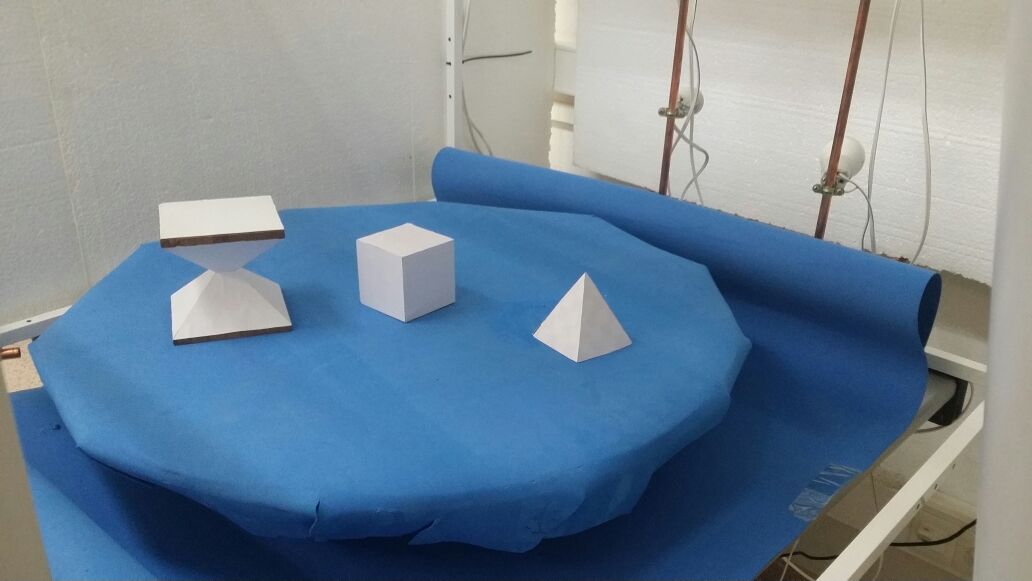}
\end{tabular}
\caption{Set-up of the experiment using three different objects and a camera.}
\label{fig:exp:setup}
\end{figure}

\begin{table}[!b]
\begin{tabular}[t]{r c}
\begin{tabular}[b]{|c||c|c|c|}
\hline
	A & $PA_1$ & $PA_2$ & $PA_3$ \\
\hline
\hline
	$PA_1$ & 0 & 90 & 90  \\
    $PA_2$ & 90 & 0 & 90 \\
    $PA_3$ & 90 & 90 & 0 \\
\hline
\end{tabular} 
& 
\includegraphics[height = 3cm]{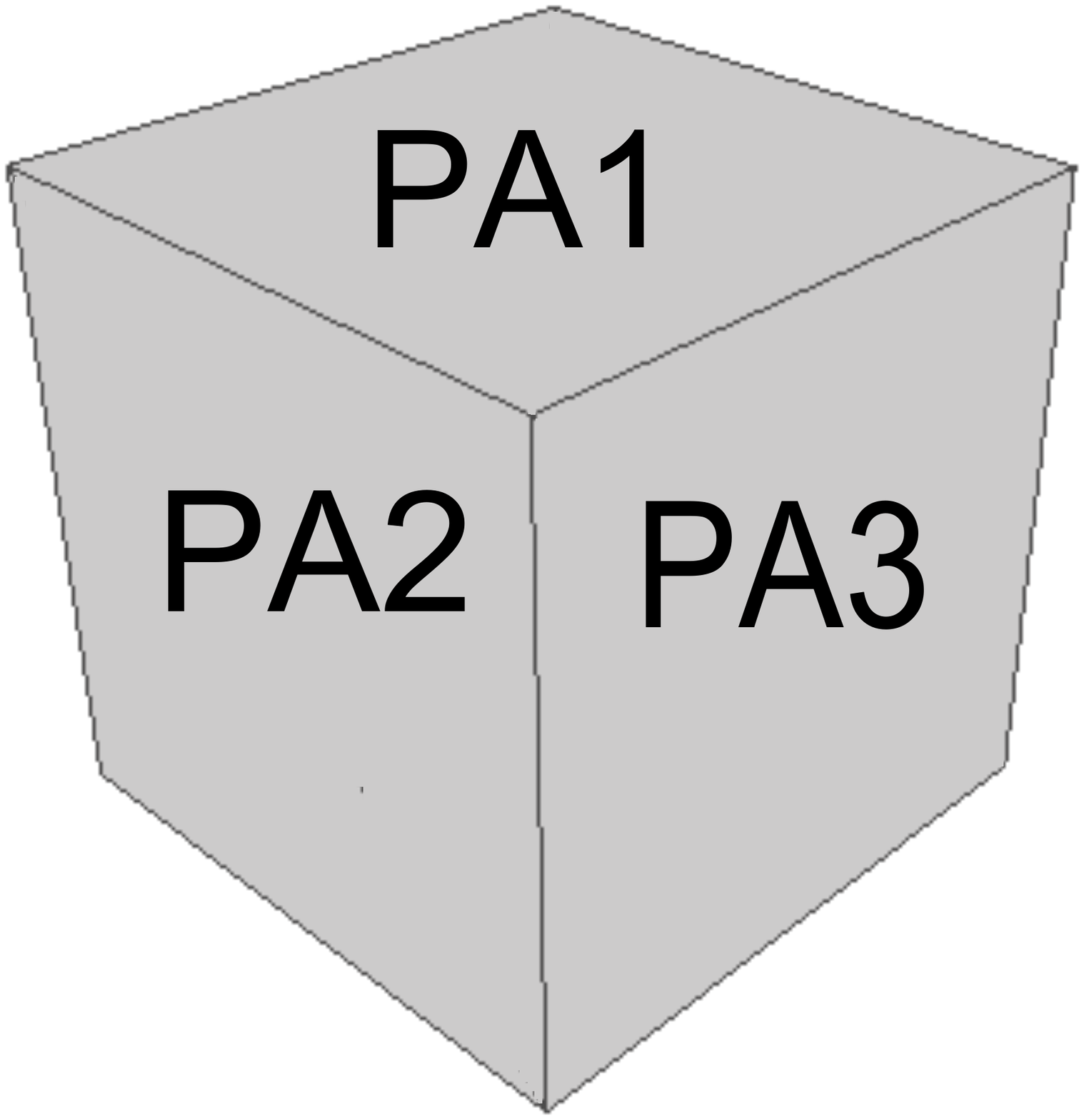}

\\

\begin{tabular}[b]{|c||c|c|}
\hline
	A & $PA_1$ & $PA_2$ \\
\hline
\hline
	$PA_1$ & 0 & 80  \\
    $PA_2$ & 80 & 0  \\
\hline
\end{tabular}
&
\includegraphics[height = 3cm]{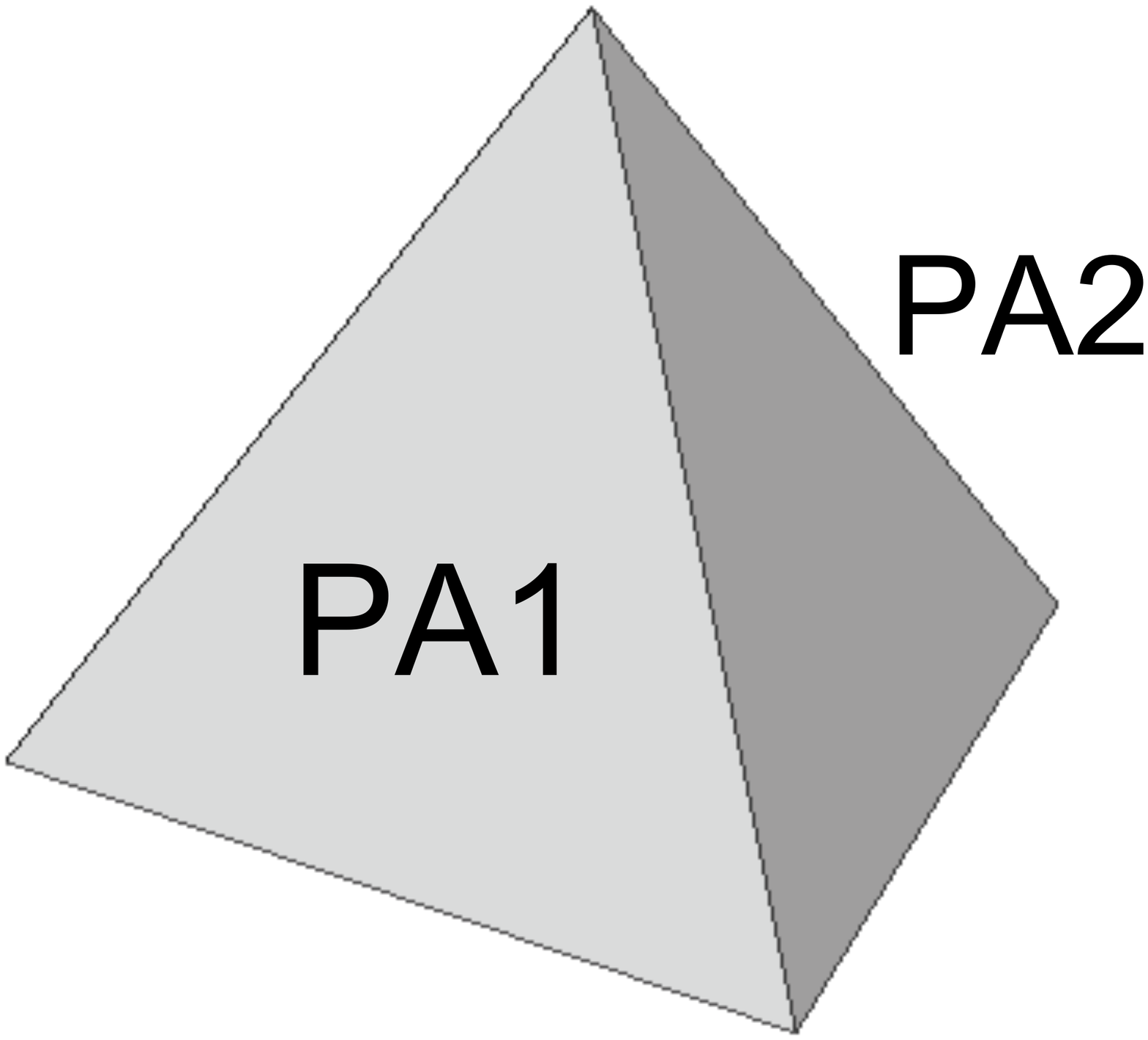}

\\

\begin{tabular}[b]{|c||c|c|c|c|c|}
\hline
	A & $PA_1$ & $PA_2$ & $PA_3$ & $PA_4$ & $PA_5$ \\
\hline
\hline
	$PA_1$ & 0 & 135 & 135 & 45 & 45  \\
    $PA_2$ & 135 & 0 & 60 & 90 & 120 \\
    $PA_3$ & 135 & 60 & 0 & 120 & 90 \\
    $PA_4$ & 45 & 90 & 120 & 0 & 60 \\
    $PA_5$ & 45 & 120 & 90 & 60 & 0\\
\hline
\end{tabular}
&
\includegraphics[height = 4cm]{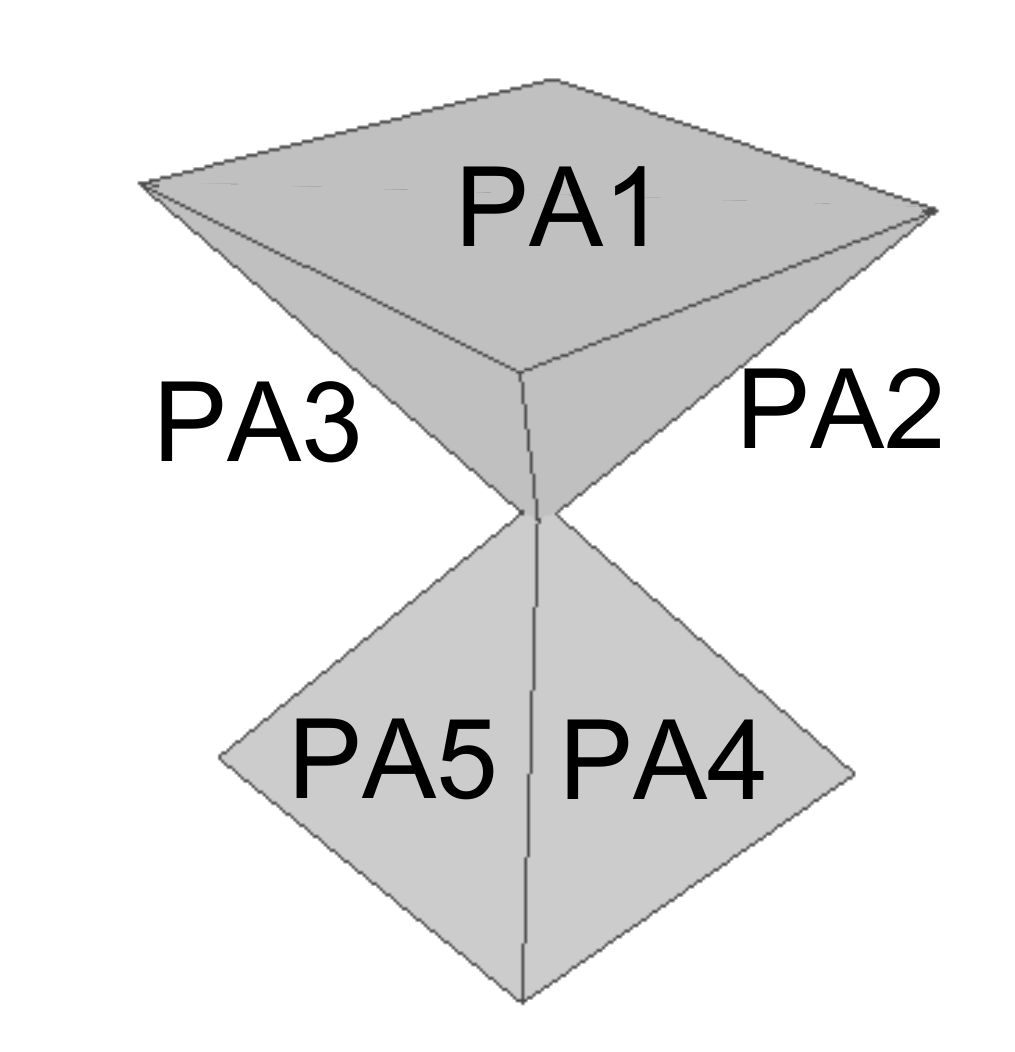}

\end{tabular}
\caption{Object angle correspondence matrices for models of a cube, pyramid and double pyramid.}
\label{tab:exp:model}
\end{table}

The proposed method was evaluated using synthetic and real data, i.e., a cube, a pyramid and a double pyramid joined at the tips (Table \ref{tab:exp:model}). The synthetic data were obtained by simulating a Microsoft Kinect sensor with Blensor \cite{Blensor2011}. This tool allowed us to create different objects, which were represented by planes, and we could move the camera to obtain different points of view. Some white Gaussian noise with different levels of dispersion was added to the synthetic data to evaluate the ability of the algorithms to handle noise. This noise was parameterized with the mean and the standard deviation according to a Gaussian probability distribution, where the mean indicated the location affected by noise and the standard deviation indicated the dispersion of the noise. The mean value $\mu$ was zero in all of the tests and the standard deviation $\sigma$ had values of $1\cdot10^{-5}$, $4\cdot10^{-5}$ and $6\cdot10^{-5}$. The Gaussian noise was applied to the depth information to simulate noise in the RGB-D sensors because the depth information is affected more significantly by incorrect values. The models represented the maximum number of faces that could be seen from a single viewpoint, as explained in Section \ref{sec:PCC}. To evaluate all possibilities, this experiment included eight points of view for each object. 

Microsoft Kinect was used to evaluate the algorithms with real data, where the objects were similar to those simulated with Blensor. A turntable was used to obtain different views (see Figure~\ref{fig:exp:setup}). The set-up also included a computer to perform experiments, where the system comprised Windows 7, an Intel i5 processor and 8 GB of RAM. The code was implemented in Matlab vR2013b. 

Table \ref{tab:exp:model} shows the angle correspondence matrices (constraints) for the three objects.

\subsection{PCC experiment}\label{sec:exp:PCC}

First, we evaluated the \textit{PCC method} (Subsection \ref{sec:PCC}). This part of the system employs the k-means method to cluster the points initially using information related to the 3D points and normal vectors. Initially, k-means needs the number of clusters, so for each object, we estimated the maximum number of faces visible in a view and we added 40\% of this number, e.g., for the cube, the maximum number of visible faces was three, so the number of clusters: $nClust = 3 + \lceil 0.4*3 \rceil = 5$. This facilitated the handling of noisy data. Moreover, in Eq. \ref{fml:similar}, $count(PB_x \approx PA_y)$ used a threshold with a value of 20, which was estimated experimentally as the best using our dataset (synthetic and real data).

\begin{figure}[!htb]
	\centering
	\begin{subfigure}[b]{0.2\textwidth}
      	\includegraphics[width=\textwidth]{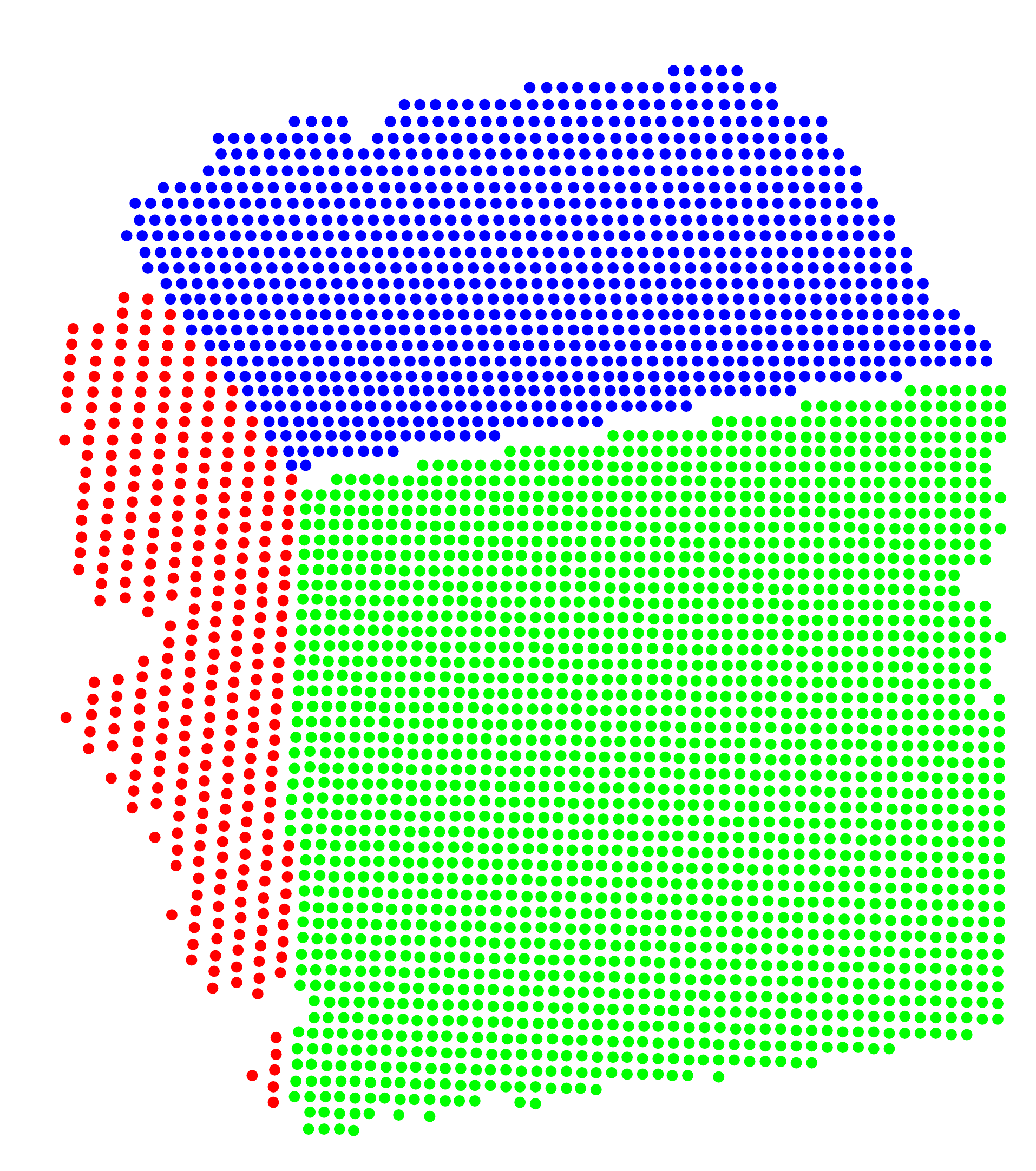} 
      	\label{fig:cube:1}
	\end{subfigure}
	\begin{subfigure}[b]{0.2\textwidth} 	
      	\includegraphics[width=\textwidth]{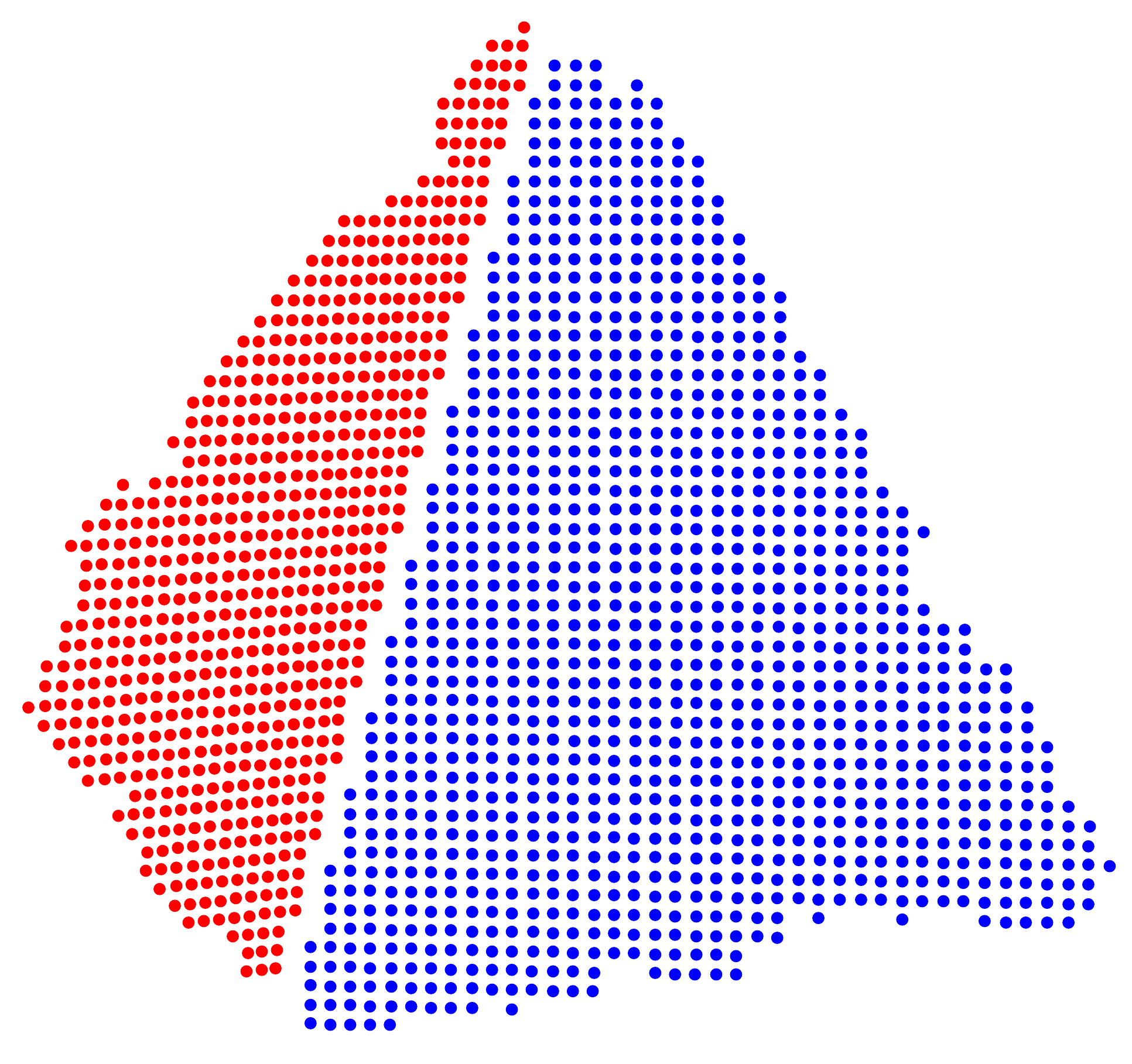}
      	\label{fig:pyr:1}
	\end{subfigure}
	\begin{subfigure}[b]{0.2\textwidth} 	
      	\includegraphics[width=\textwidth]{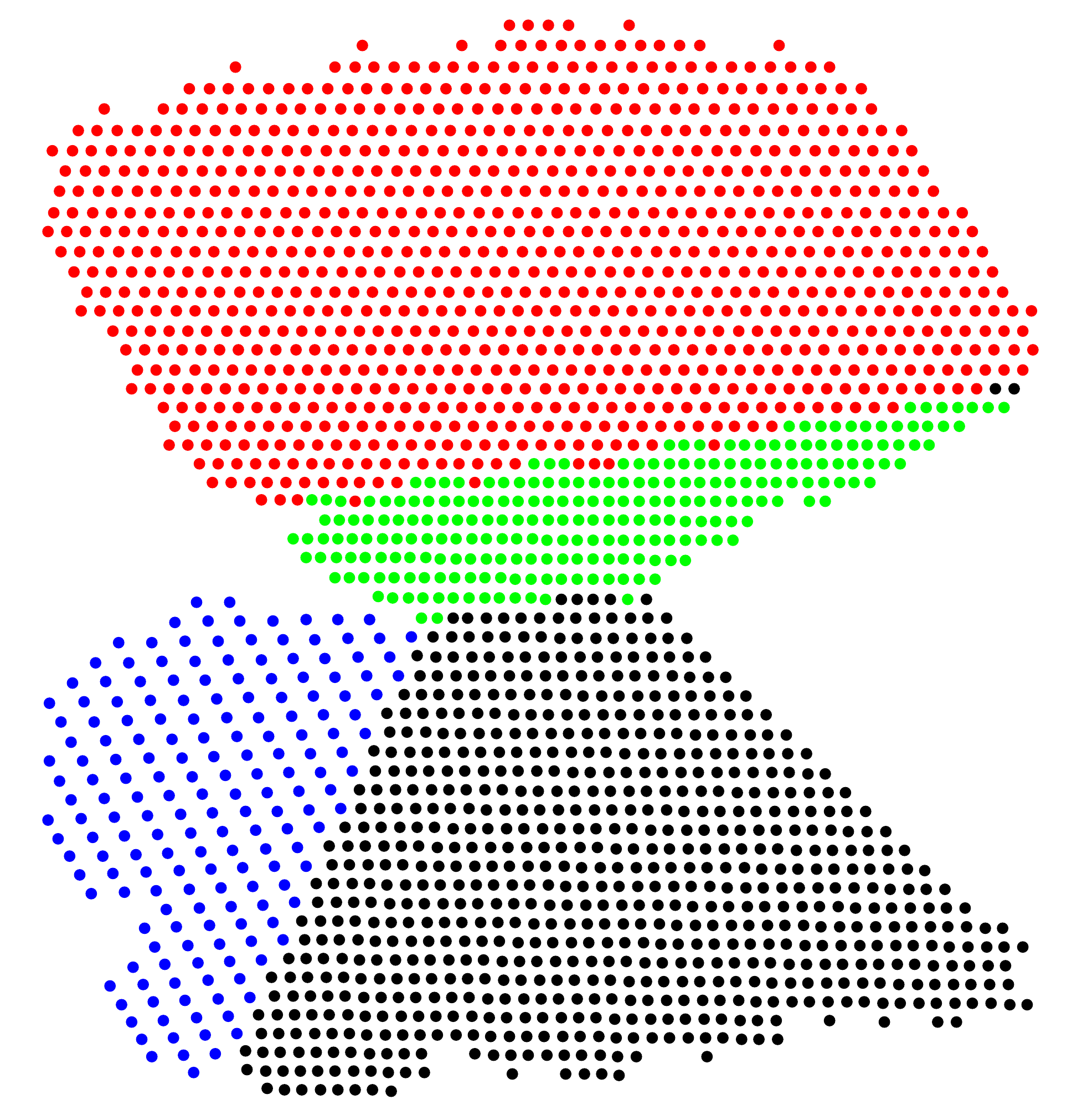}
      	\label{fig:dobpyr:1}
	\end{subfigure}

	\begin{subfigure}[b]{0.2\textwidth}
      	\includegraphics[width=\textwidth]{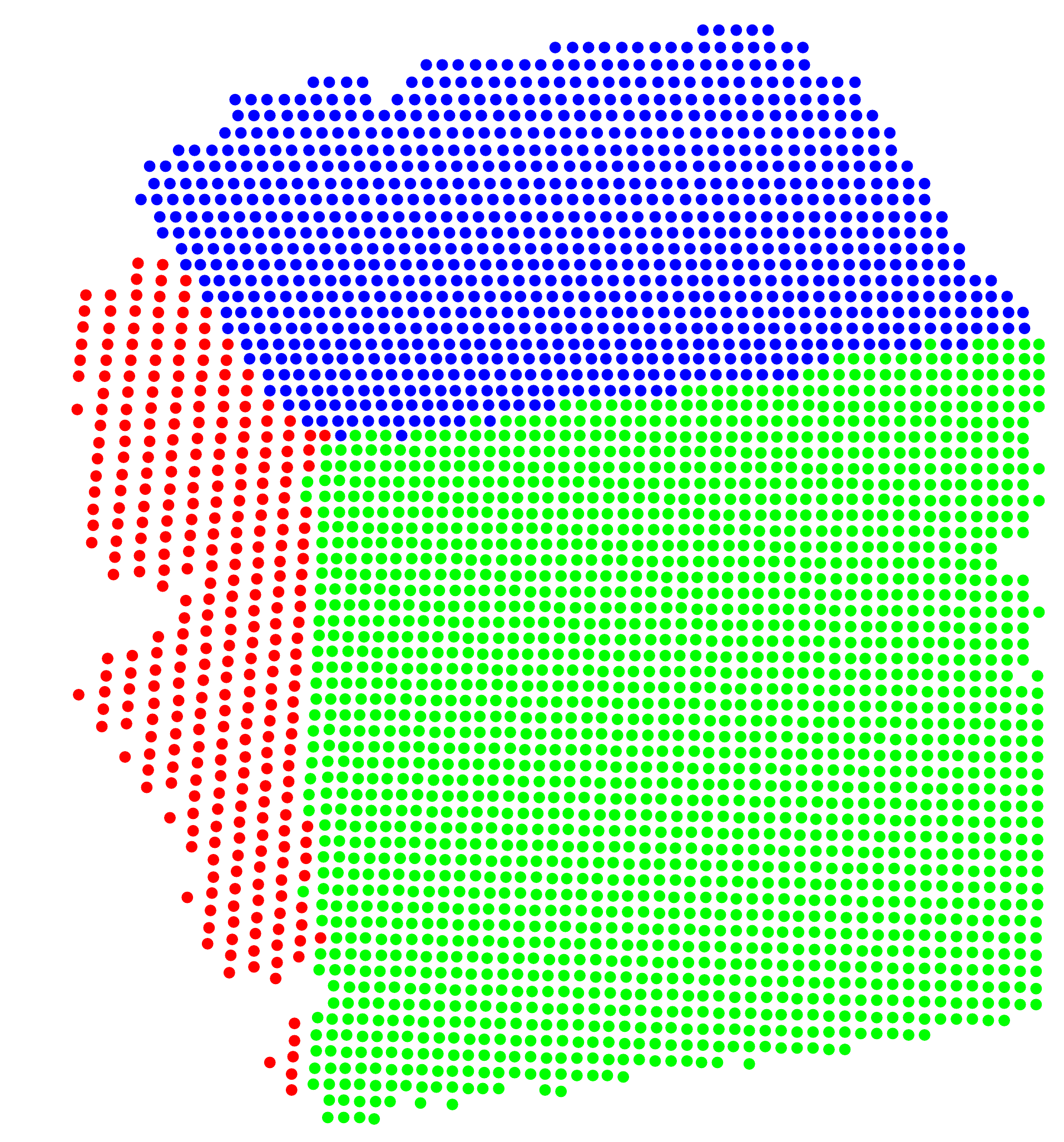} 
      	\label{fig:cube:2}
	\end{subfigure}
	\begin{subfigure}[b]{0.2\textwidth} 	
      	\includegraphics[width=\textwidth]{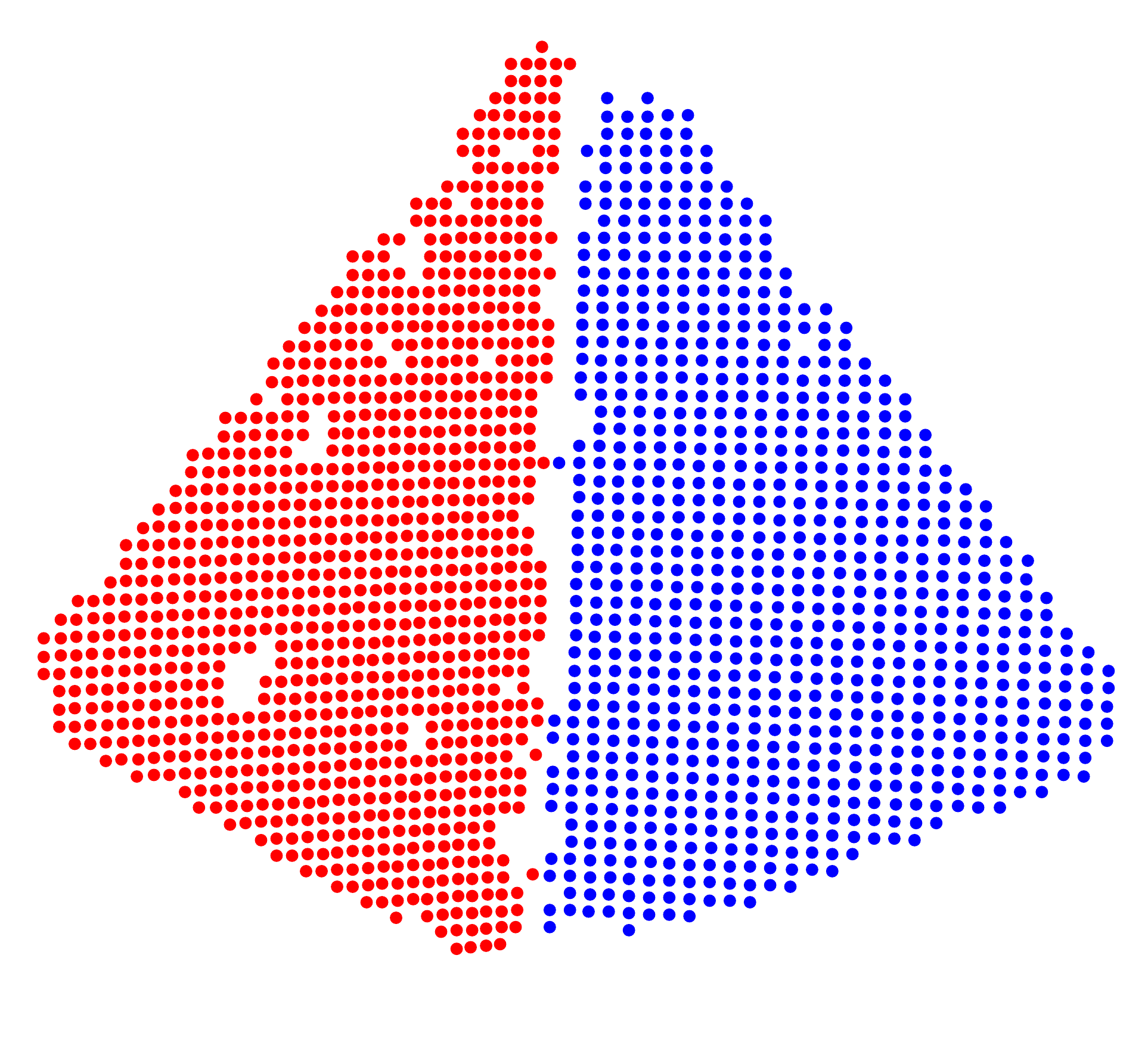}
      	\label{fig:pyr:2}
	\end{subfigure}
	\begin{subfigure}[b]{0.2\textwidth} 	
      	\includegraphics[width=\textwidth]{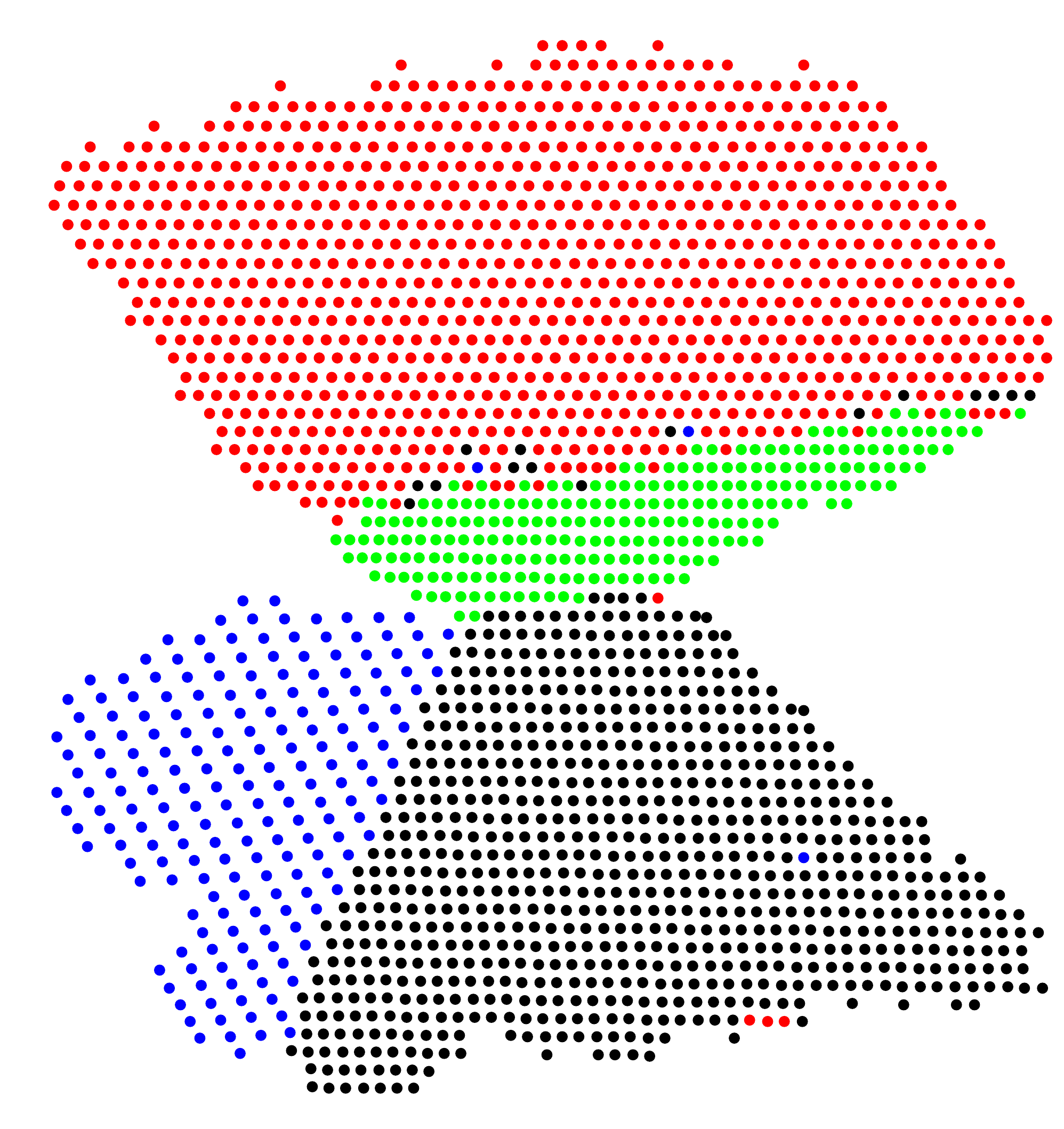}
      	\label{fig:dobpyr:2}
	\end{subfigure}

	\begin{subfigure}[b]{0.2\textwidth}
      	\includegraphics[width=\textwidth]{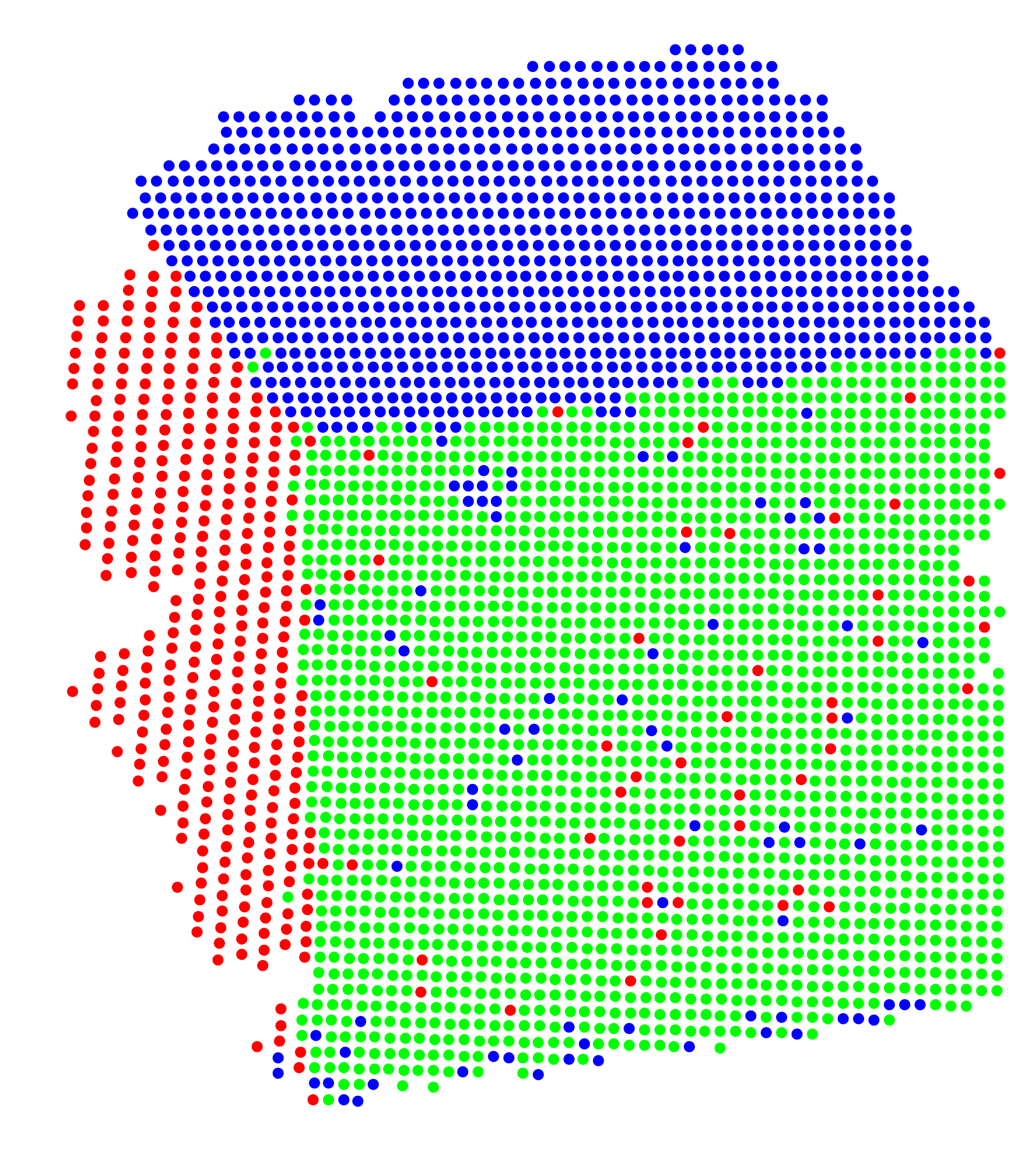} 
      	\label{fig:cube:3}
	\end{subfigure}
	\begin{subfigure}[b]{0.2\textwidth} 	
      	\includegraphics[width=\textwidth]{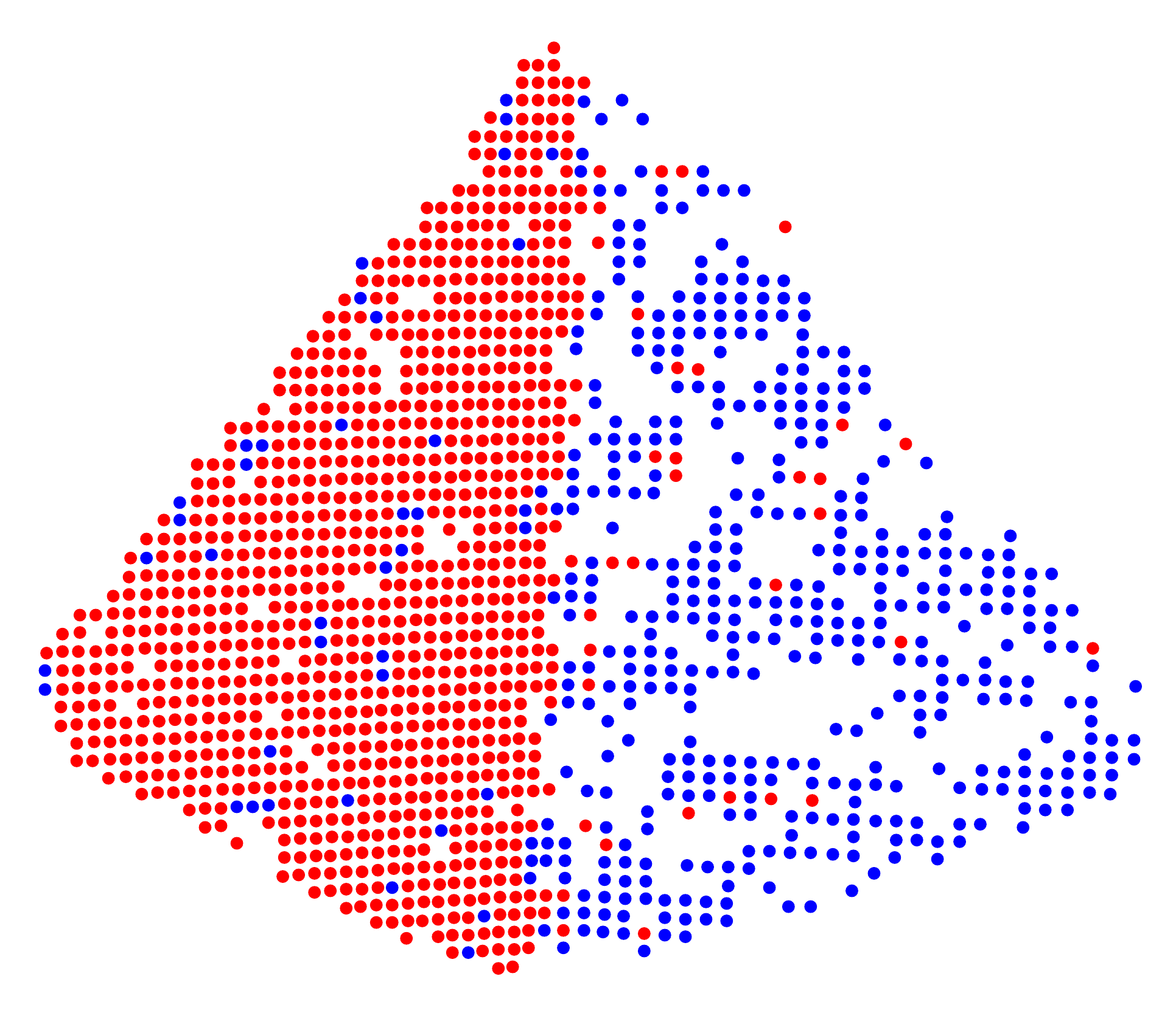}
      	\label{fig:pyr:3}
	\end{subfigure}
	\begin{subfigure}[b]{0.2\textwidth} 	
      	\includegraphics[width=\textwidth]{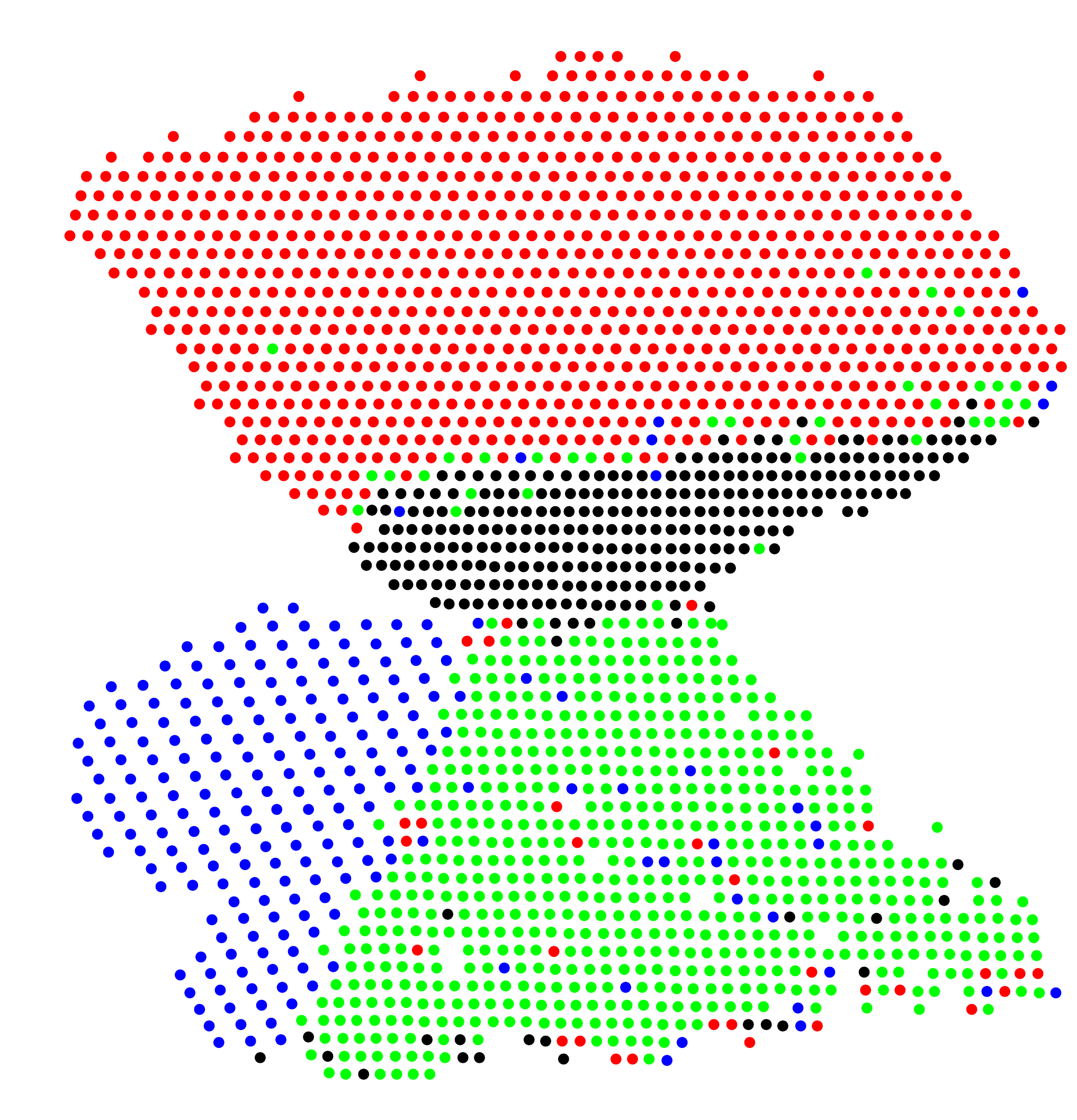}
      	\label{fig:dobpyr:3}
	\end{subfigure}

	\begin{subfigure}[b]{0.2\textwidth}
      	\includegraphics[width=\textwidth]{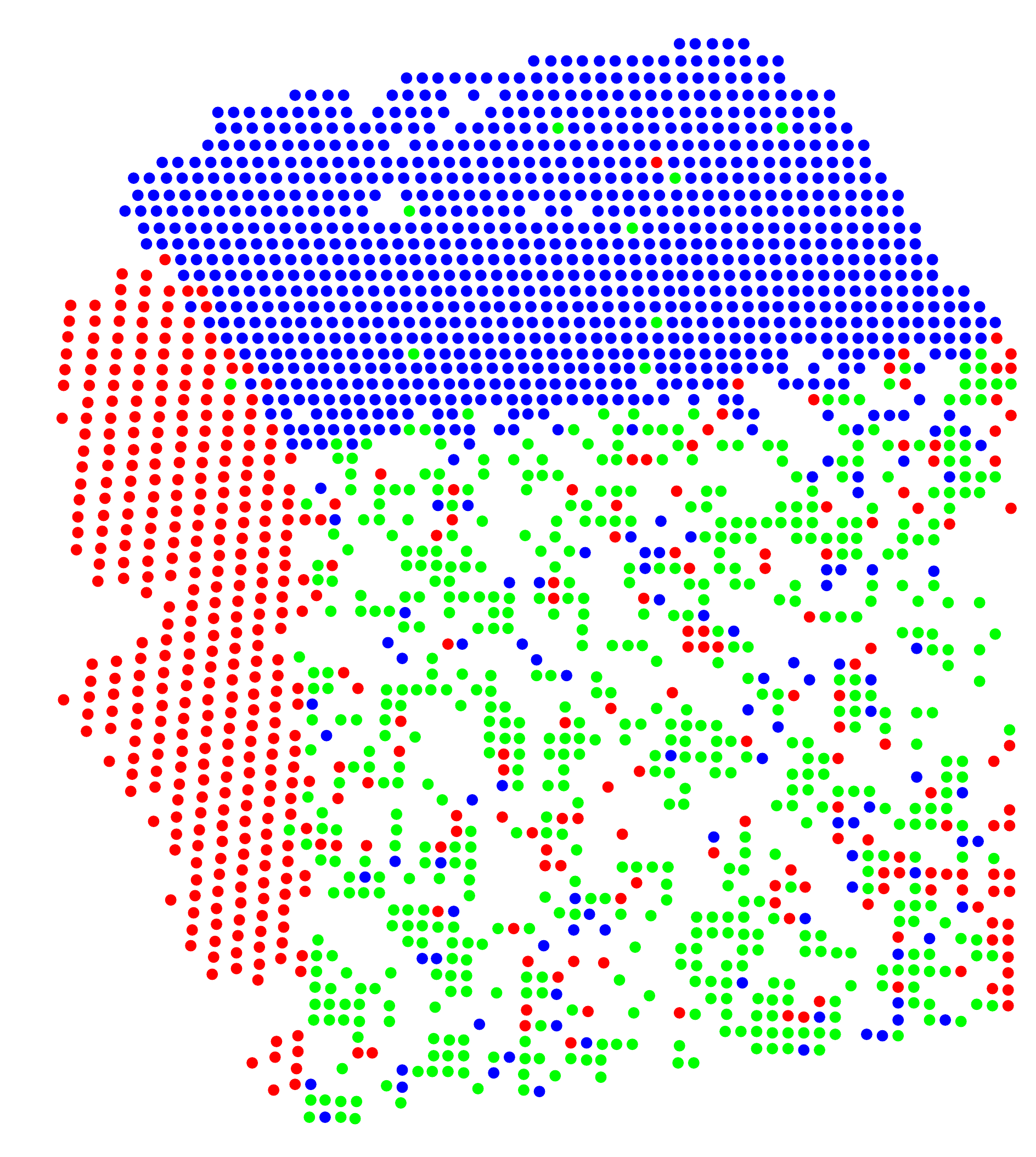} 
      	\label{fig:cube:4}
	\end{subfigure}
	\begin{subfigure}[b]{0.2\textwidth} 	
      	\includegraphics[width=\textwidth]{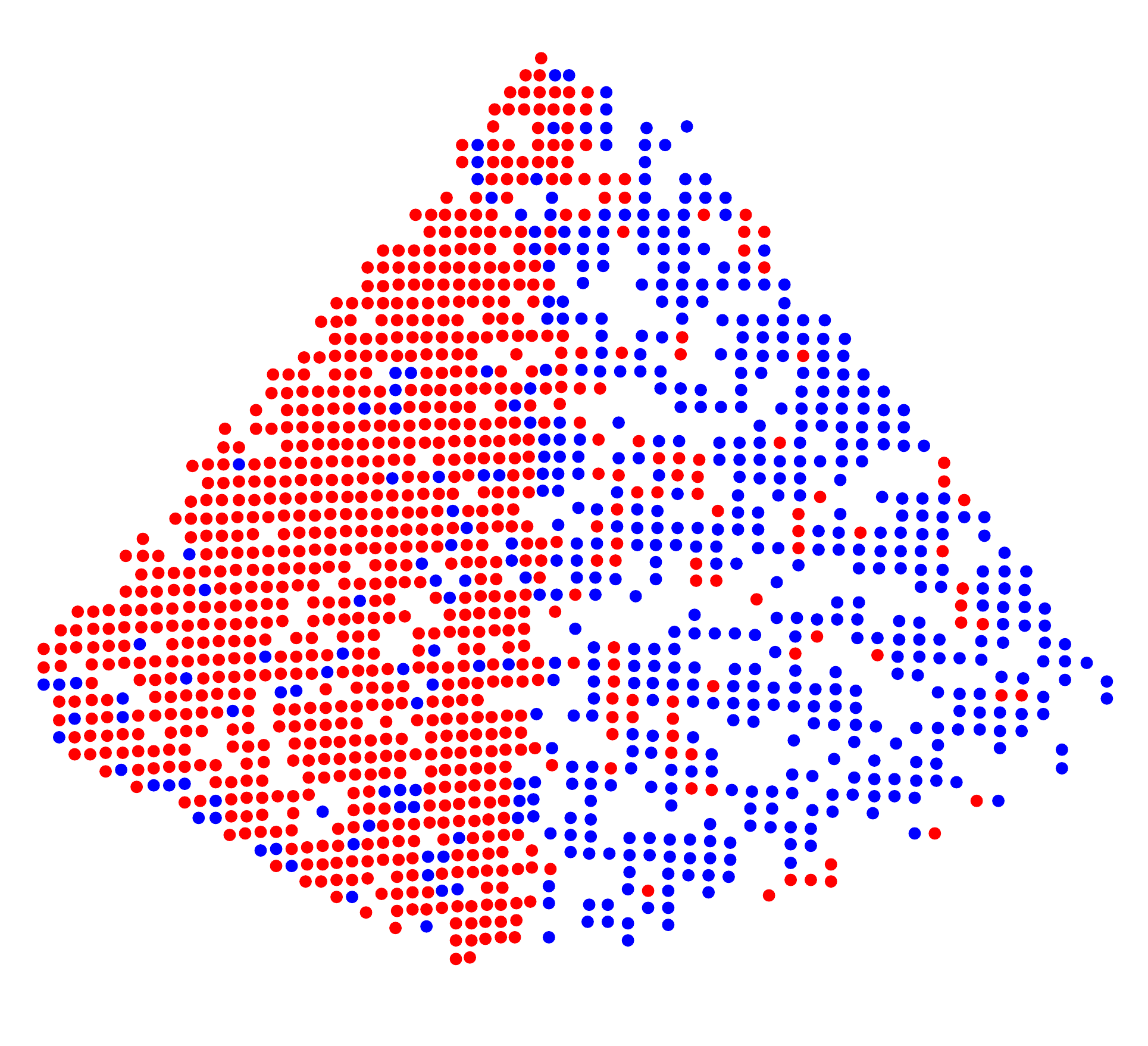}
      	\label{fig:pyr:4}
	\end{subfigure}
	\begin{subfigure}[b]{0.2\textwidth} 	
      	\includegraphics[width=\textwidth]{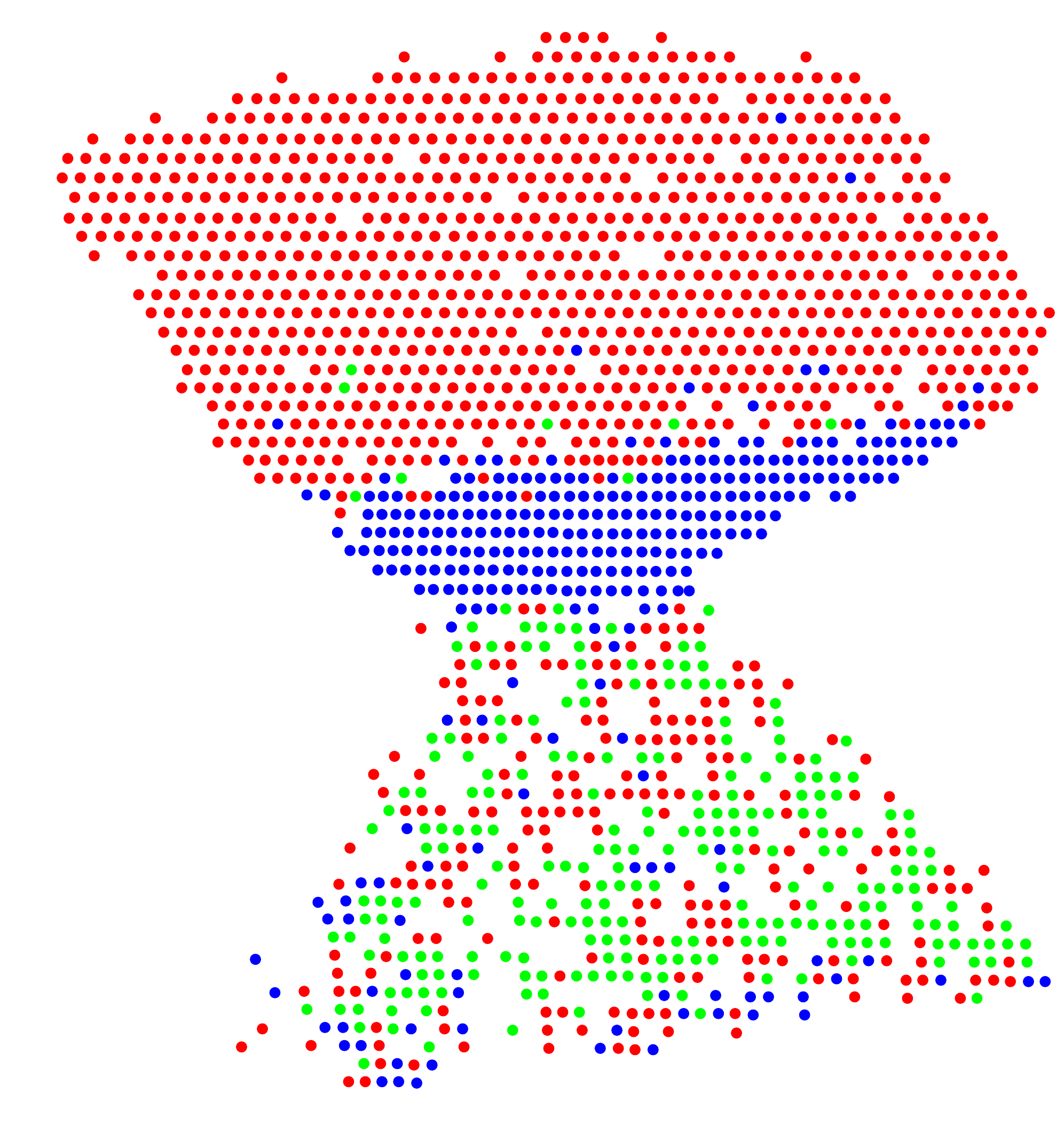}
      	\label{fig:dobpyr:4}
	\end{subfigure}

	\caption{Representations of synthetic objects with different amounts of white Gaussian noise. The first row shows the data without noise. From the second row to the last row, sigma was varied among $1\cdot10^{-5}$, $4\cdot10^{-5}$ and $6\cdot10^{-5}$.}
	\label{fig:exp:k-means}
\end{figure}

\begin{figure}[!htb]
	\centering
	\begin{subfigure}[b]{0.2\textwidth}
      	\includegraphics[width=\textwidth]{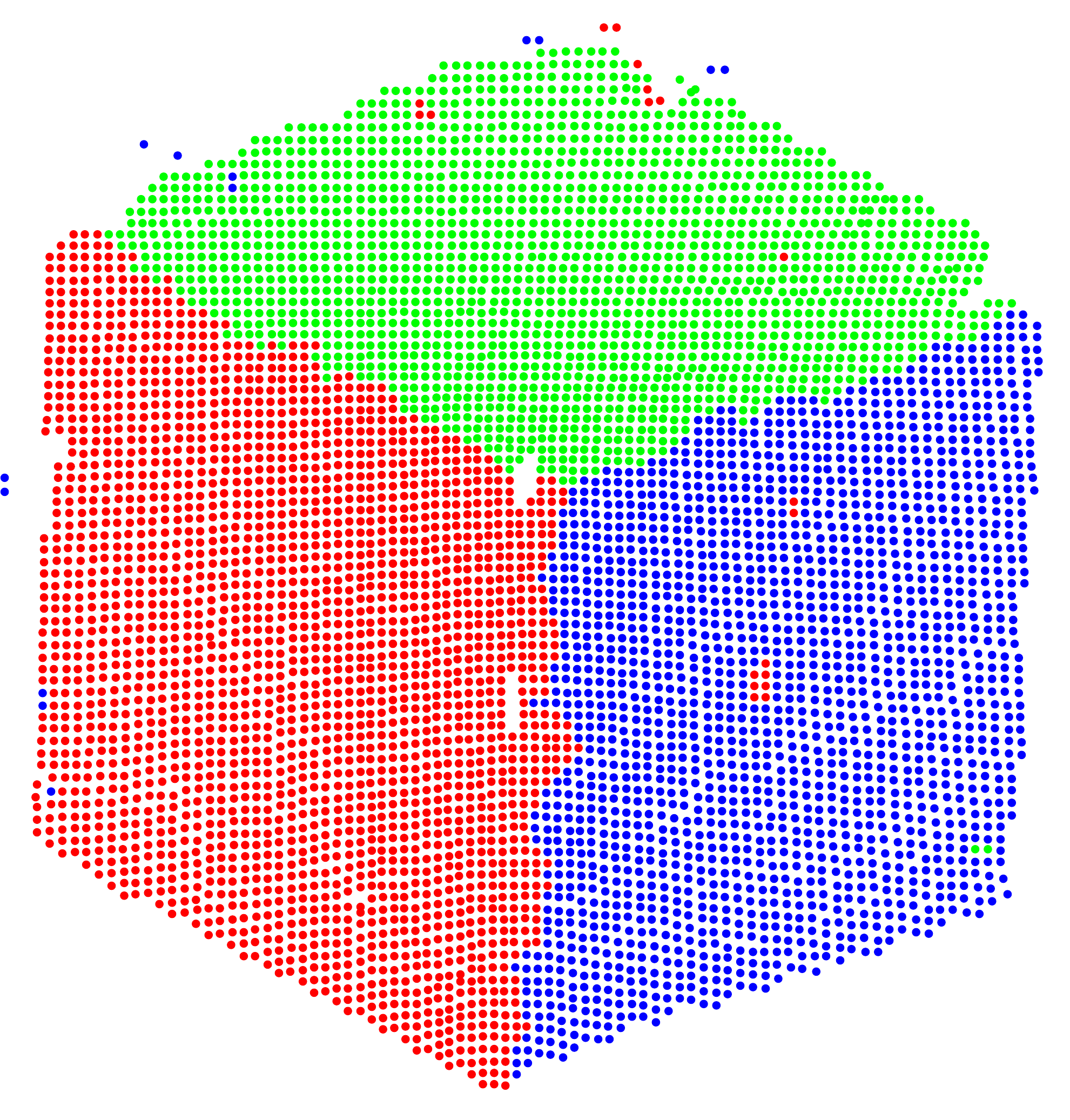} 
      	\label{fig:cube:kinect}
	\end{subfigure}
	\begin{subfigure}[b]{0.2\textwidth} 	
      	\includegraphics[width=\textwidth]{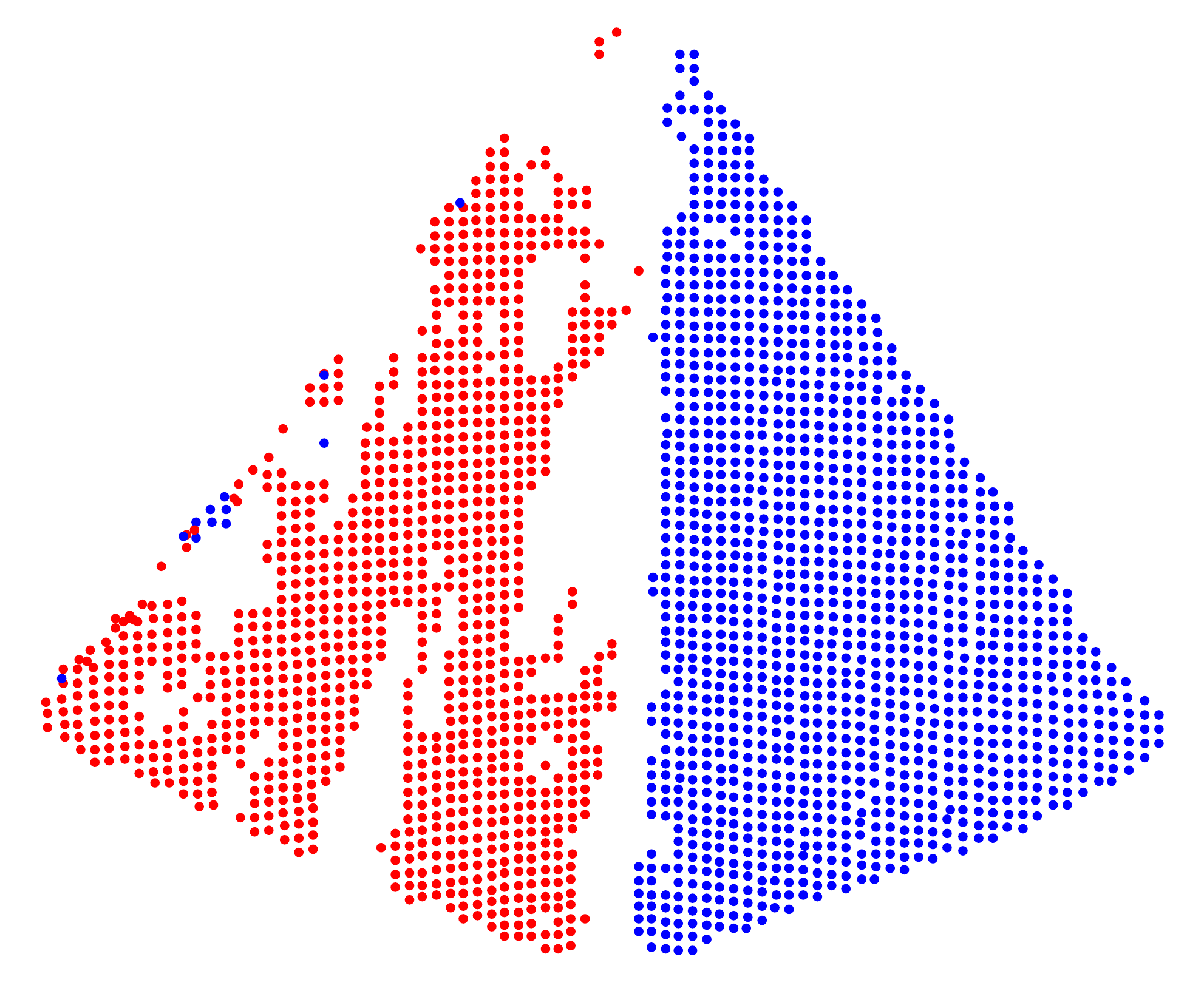}
      	\label{fig:pyr:kinect}
	\end{subfigure}
	\begin{subfigure}[b]{0.2\textwidth} 	
      	\includegraphics[width=\textwidth]{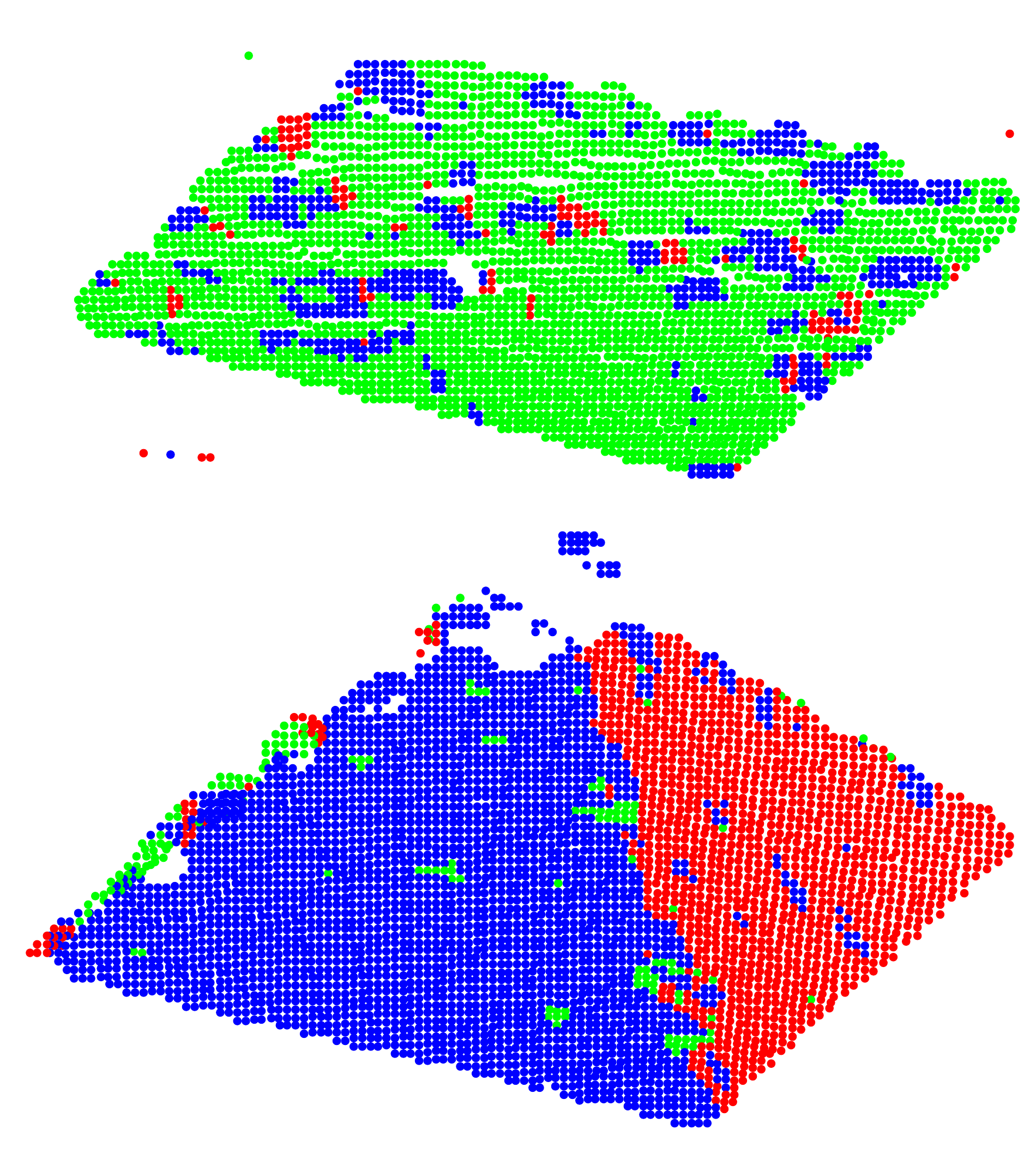}
      	\label{fig:dobpyr:kinect}
	\end{subfigure}

	\caption{Results obtained by PCC using real data acquired by the Kinect sensor.}
	\label{fig:exp:k-means_kin}
\end{figure}

Figure \ref{fig:exp:k-means} shows the results obtained using this part of the algorithm for synthetic objects with different levels of Gaussian noise. The noise was applied to the depth values, so the planes perpendicular to the viewpoint were affected more significantly by the noise. Figure \ref{fig:exp:k-means_kin} shows the results obtained for the data acquired from the Microsoft Kinect sensor. The third object, i.e., the double pyramid joined at their tips, lacked planes compared with the synthetic data because of the method employed by the sensor to obtain depth information. These sensors use a speckle pattern \cite{Khoshelham12} to calculate the depth information. Thus, for surfaces that are parallel to the camera, it is difficult to obtain a good disparity and distance values. The proposed algorithm does not require continuity in data and it can obtain the corresponding clusters. Visually, the results obtained by PCC can be used to estimate the data that best fit with the model we want to estimate. 

We performed several tests to quantitatively evaluate the effects of noise on the \textit{PCC} method. For each object in the synthetic and real data, the algorithm was run five times for each view (i.e., eight views from different angles) and noise level: five times four levels of noise, as well as with no noise for each of the eight viewpoints. Table \ref{tab:exp:numpoints} shows the average number of points for each object in the \textit{points} columns. The maximum number of faces visible for each object is shown in the \textit{faces} columns. The first row is the average number of points for synthetic data and the second row shows the number of real data acquired by the Kinect sensor. The objects are the cube in the first main column, the pyramid in the second, and the double pyramid in the third. 

\begin{table}[h!]
\centering
\begin{tabular}{|c||c|c|c|c|c|c|}
\hline
& \multicolumn{2}{|c|}{Cube} & \multicolumn{2}{|c|}{Pyramid} & \multicolumn{2}{|c|}{Double Pyramid} \\
\hline
& points & faces & points & faces & points & faces \\
\hline
Synthetic& 3227 &  3 & 1761 &  2 & 2286 & 5 \\

Kinect& 5616 &  3 & 2872 &  2 & 8033 &  5 \\
\hline
\end{tabular}
\caption{Average number of points in each object for all views. The first row shows the values for the synthetic data and the second row shows the values for the Kinect data. }
\label{tab:exp:numpoints}
\end{table} 

\subsubsection{Analysis of inliers}
The average ratios for selected data (inliers) relative to the initial data from the raw point cloud are presented in Table \ref{tab:exp:k-means_ratio}. The first four rows correspond to the synthetic data. In general, the ratio tended to decrease as the noise increased. In the final row with the results for the real data, there is a high ratio of inliers, which is a very important consideration in the next step (see Section \ref{sec:MCRANSAC}), i.e., more model estimates are possible when more data have been clustered. 

\begin{table}[h!]
\centering
\begin{tabular}{|c||c|c|c|}
\hline
	& Cube & Pyramid & Double Pyramid \\
\hline
\hline
	$\sigma = 0$ & 0.9863 & 0.9711 & 0.9580\\

    $\sigma = 1\cdot10^{-5}$ &  0.9930 & 1.0000 & 0.9660 \\

    $\sigma = 4\cdot10^{-5}$ & 0.9961 & 0.7299 & 0.8678 \\ 
    
    $\sigma = 6\cdot10^{-5}$ & 0.7235 & 0.7303 & 0.6711 \\

\hline

    $Kinect$ & 0.9964 & 0.9159 & 0.9646 \\

\hline
\end{tabular}
\caption{Ratio of total points in the point cloud relative to those obtained by PCC. }
\label{tab:exp:k-means_ratio}
\end{table}

In order to assess the performance of the clustering method in more detail, we determined the average number of points per face for each object in the eight views. We had the ground truth clustering results (i.e., those obtained by the method with no noisy data), so we compared each of the points to evaluate whether they had been clustered correctly. The results of this experiment are shown in Tables \ref{tab:exp:k-means_cube}, \ref{tab:exp:k-means_pyr} and \ref{tab:exp:k-means_dpyr}, for the cube, pyramid and double pyramid, respectively.
 
\begin{table}[h!]

   \scalebox{1}{
     \centering
    \begin{tabular}{|r|r r r||r r r||r r r||r r r|}
    \hline
 & \multicolumn{3}{c}{\textbf{Ground truth}} & \multicolumn{3}{c}{\textbf{$\sigma = 1\cdot10^{-5}$}} &  \multicolumn{3}{c}{\textbf{$\sigma = 4\cdot10^{-5}$}} & \multicolumn{3}{c}{\textbf{$\sigma = 6\cdot10^{-5}$}} \\
    \hline
    view  & F. 1 & F. 2 & F. 3 & F. 1 & F. 2 & F. 3 & F. 1 & F. 2 & F. 3 & F. 1 & F. 2 & F. 3 \\
    1     & 2011  & 930   &       & 1870  & 932   &       & 713   & 800   &       & 445   & 864   & \textbf{\textit{368}} \\
    2     & 1982  & 919   &       & 2024  & 906   &       & 589   & 777   &       & 487   & 864   & \textbf{\textit{372}} \\
    3     & 1987  & 938   &       & 2012  & 947   &       & 1106  & 839   &       & 733   & 806   &  \\
    4     & 1946  & 926   & 80    & 1933  & 937   & 111   & 535   & 862   & 248   & 381   & 755   & 370 \\
    5     & 1931  & 940   & 307   & 1921  & 953   & 326   & 637   & 918   & 382   & 565   & 885   & 419 \\
    6     & 1768  & 900   & 587   & 1796  & 926   & 589   & 1099  & 824   & 592   & 521   & 862   & 579 \\
    7     & 1629  & 896   & 851   & 1651  & 920   & 841   & 596   & 826   & 834   & 568   & 819   & 753 \\
    8     & 1505  & 920   & 1018  & 1501  & 894   & 1024  & 721   & 810   & 800   & 719   & 717   & 676 \\
    \hline
    \end{tabular}}%
   
    \caption{Results obtained per face for point clustering of the cube. Each row represents a view. Each group of three columns denotes a different level of noise. The first three columns show the number of points clustered for the ground truth (results without noise). The other groups are for $\sigma = 1\cdot10^{-5}$, $\sigma = 4\cdot10^{-5}$ and $\sigma = 6\cdot10^{-5}$, respectively. The two values in bold in the final column indicate a face that was classified incorrectly in these views. This error is explained at the end of this subsection. }
  \label{tab:exp:k-means_cube}
\end{table}%

 \begin{table}[h!]
   \centering
     \begin{tabular}{|r|rr||rr||rr||rr|}
     \hline
     \textbf{} & \multicolumn{2}{c}{\textbf{Ground truth}} & \multicolumn{2}{c}{\textbf{$\sigma = 1\cdot10^{-5}$}} & \multicolumn{2}{c}{\textbf{$\sigma = 4\cdot10^{-5}$}} & \multicolumn{2}{c}{\textbf{$\sigma = 6\cdot10^{-5}$}} \\
     \hline
     view  & Face 1 & Face 2 & Face 1 & Face 2 & Face 1 & Face 2 & Face 1 & Face 2 \\
     1     & 1141  & 368   & 1157  & 357   & 543   & 426   & 521   & 460 \\
     2     & 1161  & 389   & 1171  & 375   & 539   & 499   & 539   & 506 \\
     3     & 1126  & 445   & 1137  & 430   & 556   & 485   & 440   & 610 \\
     4     & 1045  & 527   & 1049  & 512   & 558   & 617   & 523   & 619 \\
     5     & 1020  & 600   & 1020  & 651   & 578   & 634   & 493   & 644 \\
     6     & 902   & 735   & 905   & 735   & 496   & 728   & 534   & 665 \\
     7     & 803   & 833   & 806   & 927   & 535   & 629   & 580   & 717 \\
     8     & 752   & 880   & 780   & 950   & 466   & 725   & 537   & 609 \\
     \hline
     \end{tabular}%
     \caption{Results obtained per face for point clustering of the pyramid. Each row represents a view. Each group of three columns denotes a different level of noise. The first three columns show the number of points clustered for the ground truth (results without noise). The other groups are for $\sigma = 1\cdot10^{-5}$, $\sigma = 4\cdot10^{-5}$ and $\sigma = 6\cdot10^{-5}$, respectively.}
   \label{tab:exp:k-means_pyr}%
 \end{table}

\begin{table}[h!]
  \centering
  \scalebox{1}{
   \begin{tabular}{|r|rrrrr||rrrrr|}
    \hline
     & \multicolumn{5}{c}{\textbf{Ground truth}} & \multicolumn{5}{c}{\textbf{$\sigma = 1\cdot10^{-5}$}} \\
   	\hline
    view  & F. 1 & F. 2 & F. 3 & F. 4 & F. 5 & F. 1 & F. 2 & F. 3 & F. 4 & F. 5 \\
    1     & 980   & 192   & 685   & 295   &       & 870   & 198   & 652   & 297   &  \\
    2     & 981   & 203   & 705   & 306   &       & 881   & 215   & 671   & 295   &  \\
    3     & 962   & 200   & 713   & 257   &       & 897   & 198   & 709   & 263   &  \\
    4     & 969   & 211   & 677   & 225   &       & 870   & 214   & 664   & 246   &  \\
    5     & 922   & 228   & 686   &       & 186   & 858   & 245   & 680   &       & 191 \\
    6     & 916   & 229   & 680   &       & 257   & 849   & 228   & 678   &       & 267 \\
    7     & 882   & 212   & 669   &       & 329   & 817   & 186   & 679   & 42    & 320 \\
    8     & 856   & 187   & 639   &       & 395   & 754   & 180   & 663   &       & 368 \\
    \hline
      & \multicolumn{5}{c}{\textbf{$\sigma = 4\cdot10^{-5}$}} & \multicolumn{5}{c}{\textbf{$\sigma = 6\cdot10^{-5}$}} \\
    \hline
    view  & F. 1 & F. 2 & F. 3 & F. 4 & F. 5 & F. 1 & F. 2 & F. 3 & F. 4 & F. 5 \\
    1     & 826   & 226   & 194   & 300   &       & 786   & 259   &       & 235   &  \\
	2     & 1160  &       & 171   & 295   &       & 824   &       & 273   & 292   &  \\
    3     & 840   &       & 249   & 266   &       & 565   & 230   & 545   &       &  \\
    4     & 959   & 207   & 185   & 192   &       & 189   &       & 228   & 177   &  \\
    5     & 857   & 254   & 257   &       & 213   & 805   &       & 248   &       &  \\
    6     & 969   & 222   & 217   &       & 268   & 1108  & 273   &       &       & 246 \\
    7     & 1013  &       & 253   & 81    & 259   & 1315  & 208   & 253   & 84    &  \\
    8     & 844   & 199   & 288   &       &       & 737   & 219   & 299   &       &  \\
	\hline
    \end{tabular}}%
    \caption{Results obtained per face for point clustering of the double pyramids. Each row represents a view. Each group of three columns denotes a different level of noise. The first three columns show the number of points clustered for the ground truth (results without noise). The other of groups are for $\sigma = 1\cdot10^{-5}$, $\sigma = 4\cdot10^{-5}$ and $\sigma = 6\cdot10^{-5}$, respectively. The two values in bold in the final column indicate a face that was classified incorrectly for these views. This error is explained at the end of this subsection. }
  \label{tab:exp:k-means_dpyr}%
\end{table}

Different conclusions can be reached based on Tables \ref{tab:exp:k-means_cube}--\ref{tab:exp:k-means_dpyr}. It was more difficult to obtain a result when more clusters had to be fitted. For example, the pyramid figure required a maximum of two planes, so the \textit{k} number of clusters could be three or four, depending on the percentage that we want to add (in this case, 40\%). However, the double pyramid had five planes and adding this percentage required seven clusters with k-means. Thus, there was a lower probability of each point being placed in the correct cluster for the second shape. The constraints used to improve the classification only consider the angles between the planes and not their positions. Therefore, when the model had several clusters and noisy data, the possibility of finding an incorrect plane that satisfied a constraint in the model increased. This was particularly notable when the number of plane models exceeded those viewed from the camera (i.e., when two sides of the cube were acquired, although the cube model had three faces (see Fig. \ref{fig:exp:k-means_wrong}).    

\subsubsection{Processing time}
In terms of the processing time (Table \ref{tab:exp:k-means_time}), the performance of the algorithm varied with respect to noise. In general, the time increased in proportion to the amount of noise, but the time was highly dependent on the number of points. If we compare the results in this table with Table \ref{tab:exp:numpoints}, which involved more input data, the processing time was higher. For example, the synthetic cube had 3227 points and the time was in the order of 0.3 s, whereas the pyramid had 1761 points and the corresponding time was lower, i.e.,  approximately 0.1 s. Another interesting conclusion that can be obtained from this data is that using similar data, the time tended to be greater when more planes had to be estimated. The real pyramid had 2872 points and the synthetic double pyramid had 2286, but the time was greater for the latter due to the number of planes in the model. These two main conclusions applied to the time required for the double pyramid with the Kinect data (last column and last row of Table~\ref{tab:exp:k-means_time}), where the number of points was the highest at 8033, and so was the number of planes, i.e., five. Therefore, this example required the longest processing time in this experiment.

\begin{table}[h!]
\centering
\begin{tabular}{|c||c|c|c|}
\hline
	& Cube & Pyramid & Double Pyramid \\
\hline
\hline
    $\sigma = 0$ &  0.2151 s & 0.0662 s & 0.2303 s \\

    $\sigma = 1\cdot10^{-5}$ &  0.3465 s & 0.1093 s & 0.2773 s \\

    $\sigma = 4\cdot10^{-5}$ & 0.4482 s & 0.1267 s & 0.4048 s \\

	$\sigma = 6\cdot10^{-5}$ & 0.3996 s & 0.1153 s & 0.4303 s \\

\hline

    $Kinect$ & 0.4531 s & 0.1149 s & 1.1034 s \\

\hline
\end{tabular}
\caption{Average time (s) required by the PCC algorithm with respect to the object and noise. }
\label{tab:exp:k-means_time}
\end{table}

\subsubsection{Discussion of clustering}
\begin{figure}[!htb]
	\centering
	\begin{subfigure}[b]{0.3\textwidth} 	
      	\includegraphics[width=\textwidth]{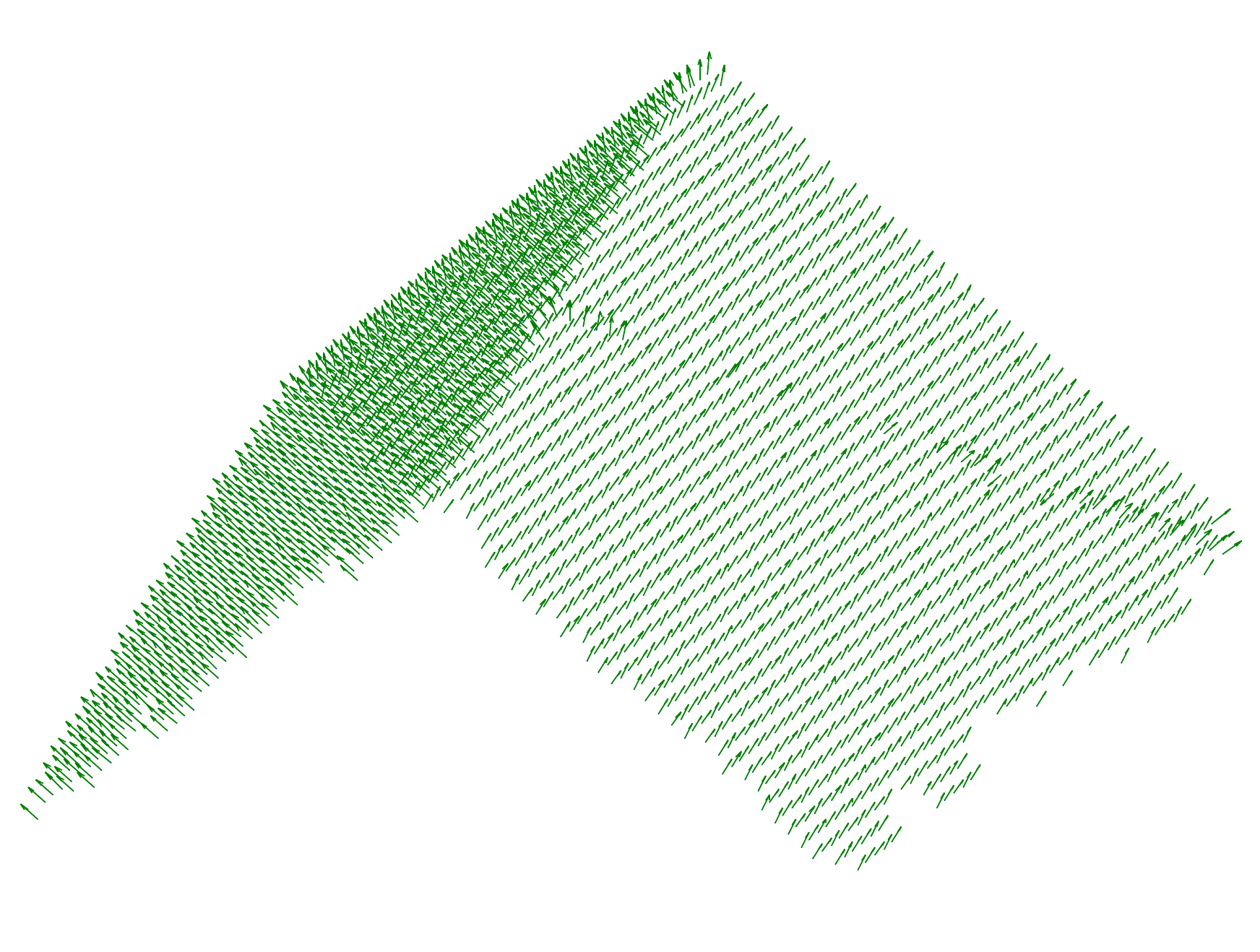}
      	\label{fig:exp:normals1}
	\end{subfigure}
	\begin{subfigure}[b]{0.3\textwidth} 	
      	\includegraphics[width=\textwidth]{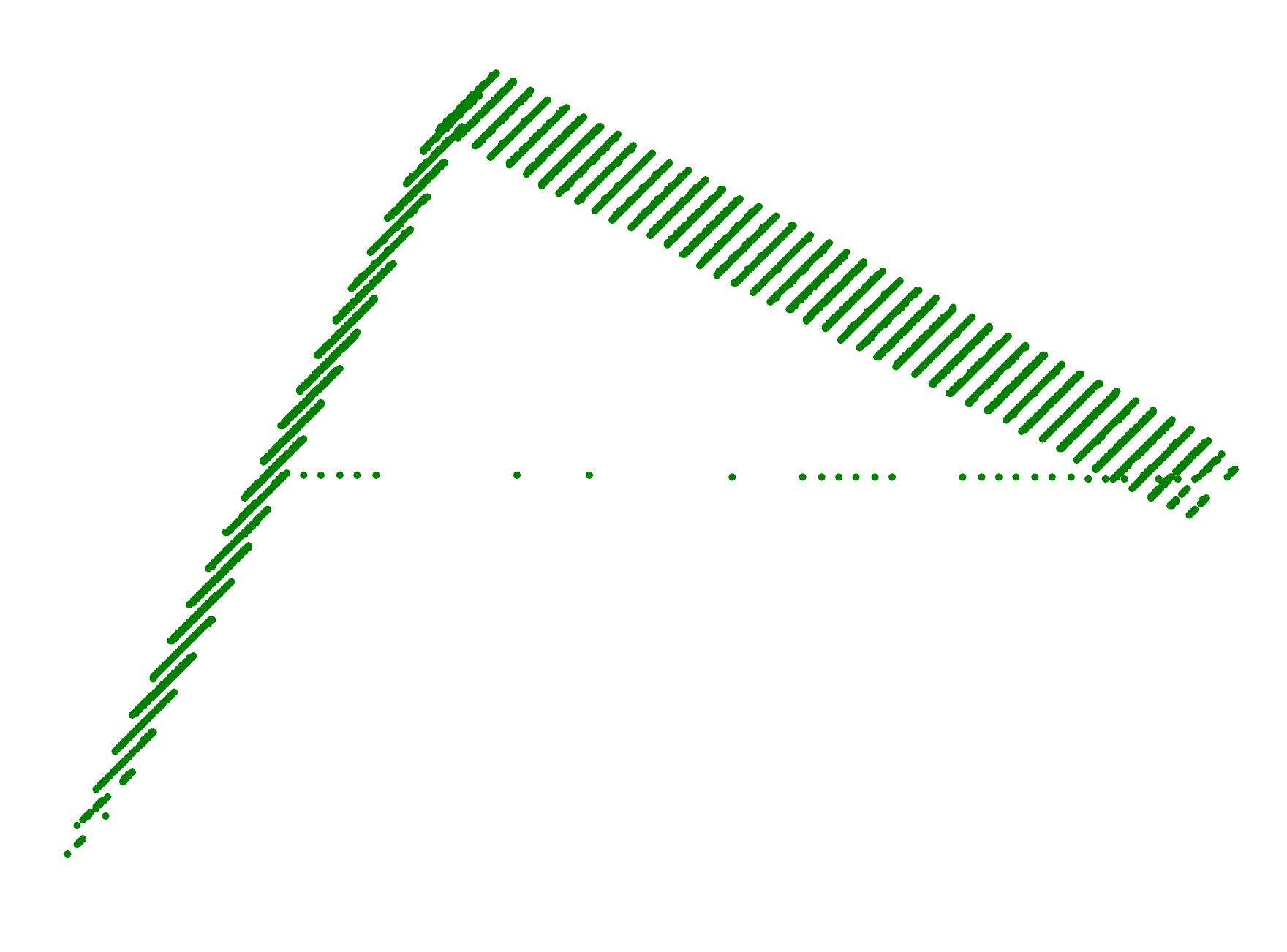}
      	\label{fig:exp:normals2}
	\end{subfigure}	
	\begin{subfigure}[b]{0.2\textwidth} 	
      	\includegraphics[width=\textwidth]{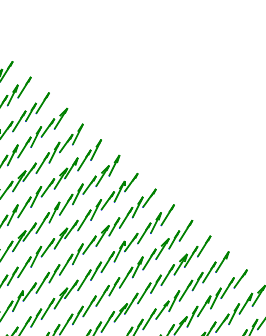}
      	\label{fig:exp:normals3}
	\end{subfigure}
	
	\begin{subfigure}[b]{0.3\textwidth} 	
      	\includegraphics[width=\textwidth]{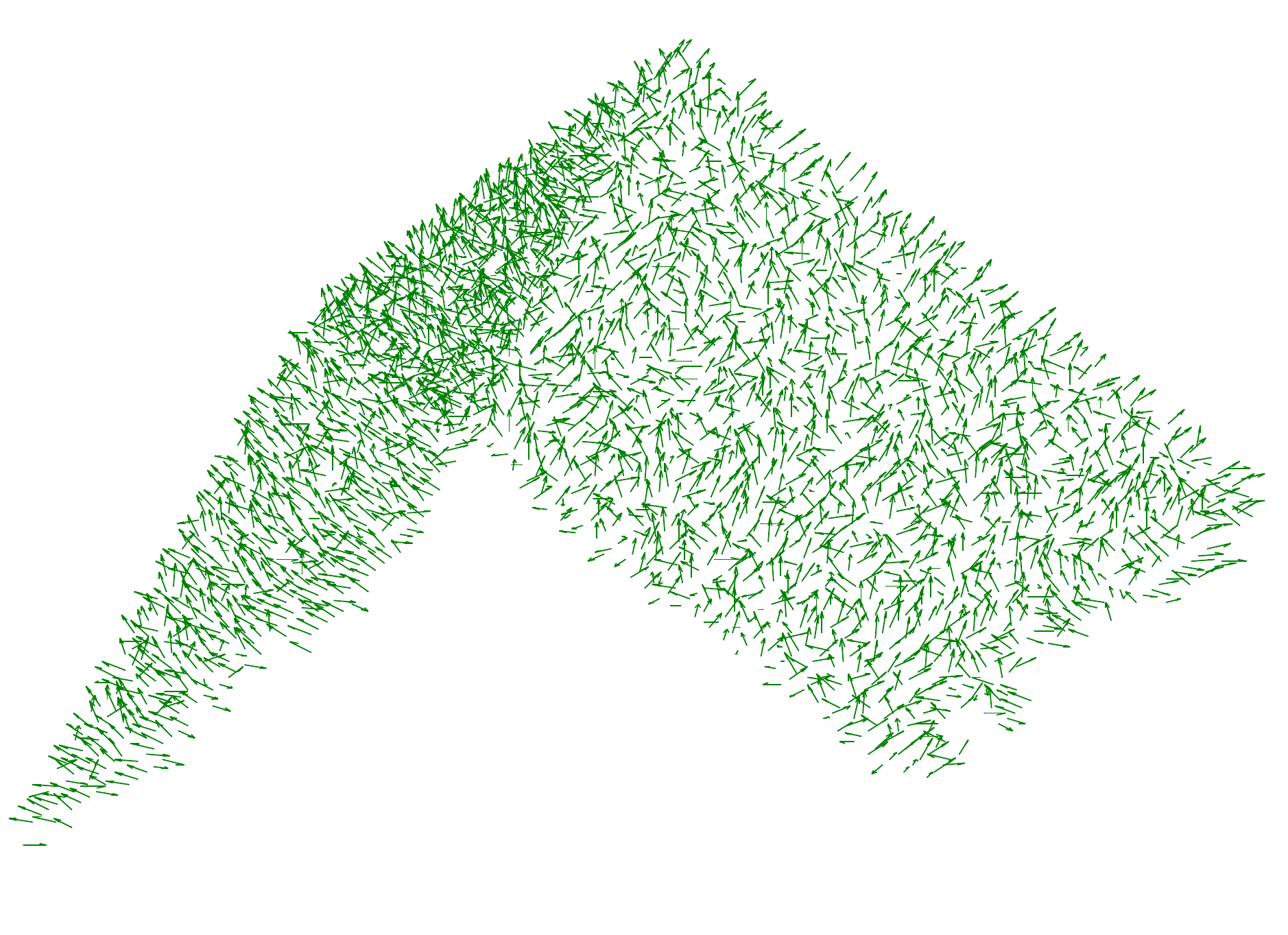}
      	\label{fig:exp:normals4}
	\end{subfigure}
	\begin{subfigure}[b]{0.3\textwidth} 	
      	\includegraphics[width=\textwidth]{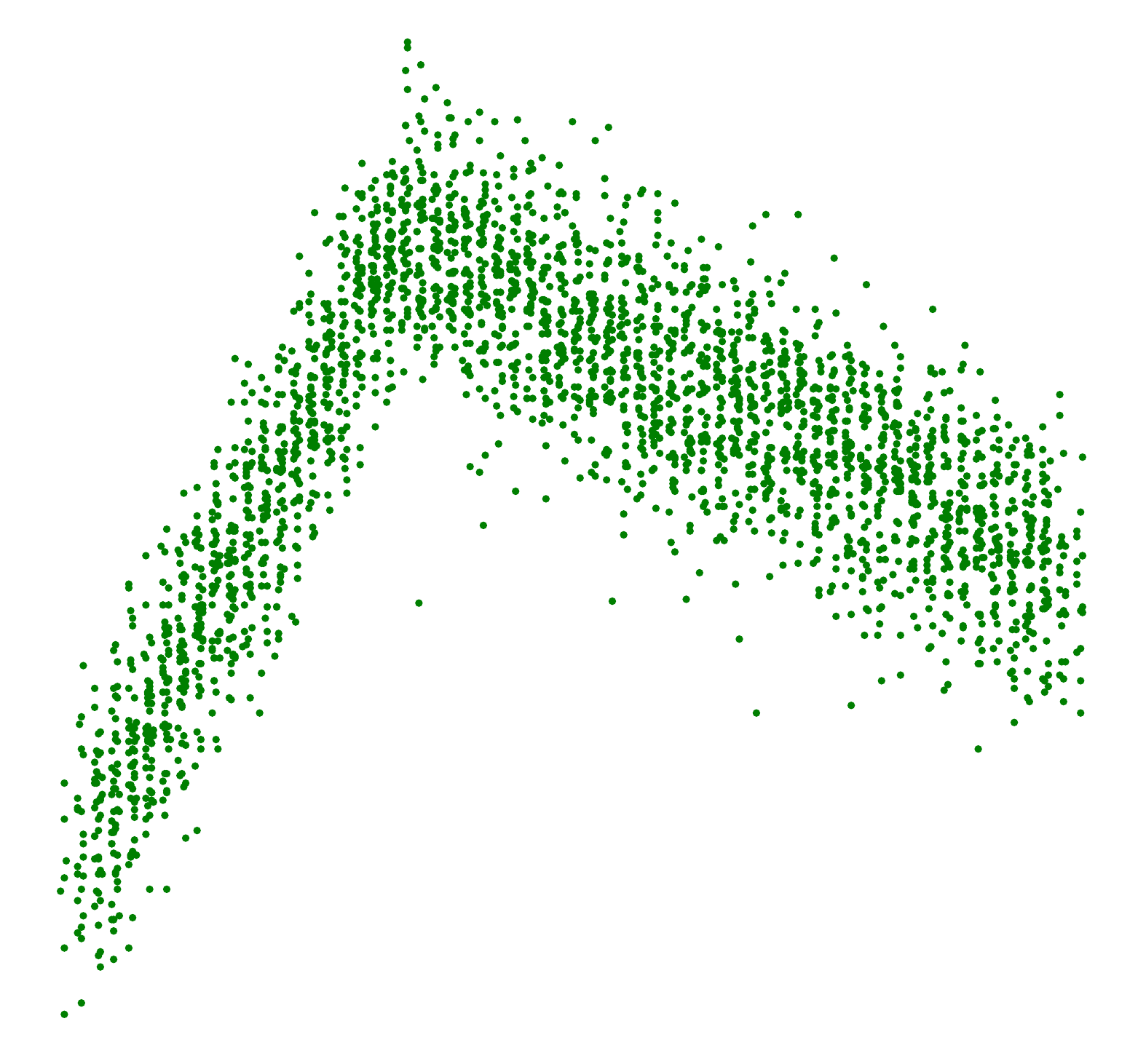}
      	\label{fig:exp:normals5}
	\end{subfigure}
	\begin{subfigure}[b]{0.2\textwidth} 	
      	\includegraphics[width=\textwidth]{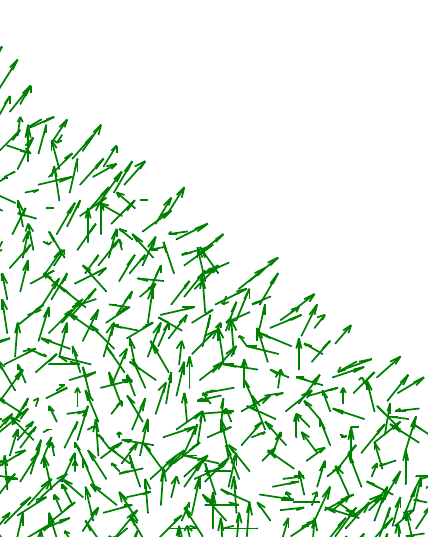}
      	\label{fig:exp:normals6}
	\end{subfigure}
	
	\caption{The normals associated with the data for the cube. The first row lacks noise and in the second, $\sigma = 4\cdot10^{-5}$. The variation in the normal orientation of the second shape produced clustering errors.}
	\label{fig:exp:normals}
\end{figure}

In this experiment, we assumed that the angles obtained by the method were due to the constrained step. The clusters returned by the method had to satisfy the constraints. However, when a large amount of noise was present, the clusters could be calculated incorrectly. To calculate the orientation of each cluster, we used the mean value of all the normals that belonged to the cluster. The normals associated with each point in the cluster were calculated using a neighbourhood, and thus the normals varied greatly if the noise level was high, even for close neighbours, as shown in Figure \ref{fig:exp:normals}. This figure shows the normals estimated in an ideal situation (left top) and the associated point cloud (middle top), as well as with a large amount of noise (left down) with the associated point cloud (middle down). The final two images show details of the normals in both situations.

\begin{figure}[!htb]
	\centering
	\begin{subfigure}[b]{0.3\textwidth} 	
      	\includegraphics[width=\textwidth]{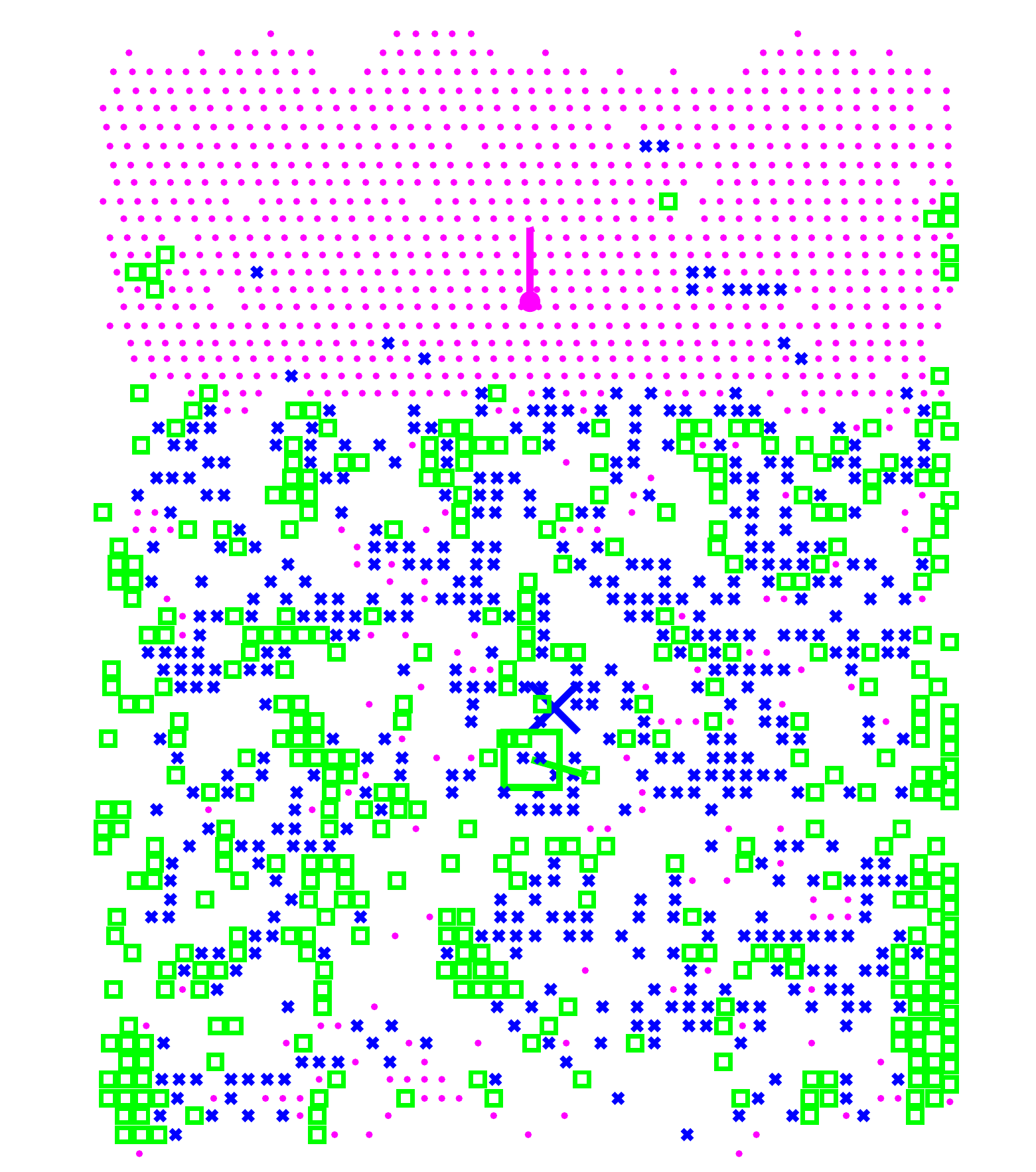}
      	\label{fig:exp:k-means_wrong1}
	\end{subfigure}
	\begin{subfigure}[b]{0.3\textwidth} 	
      	\includegraphics[width=\textwidth]{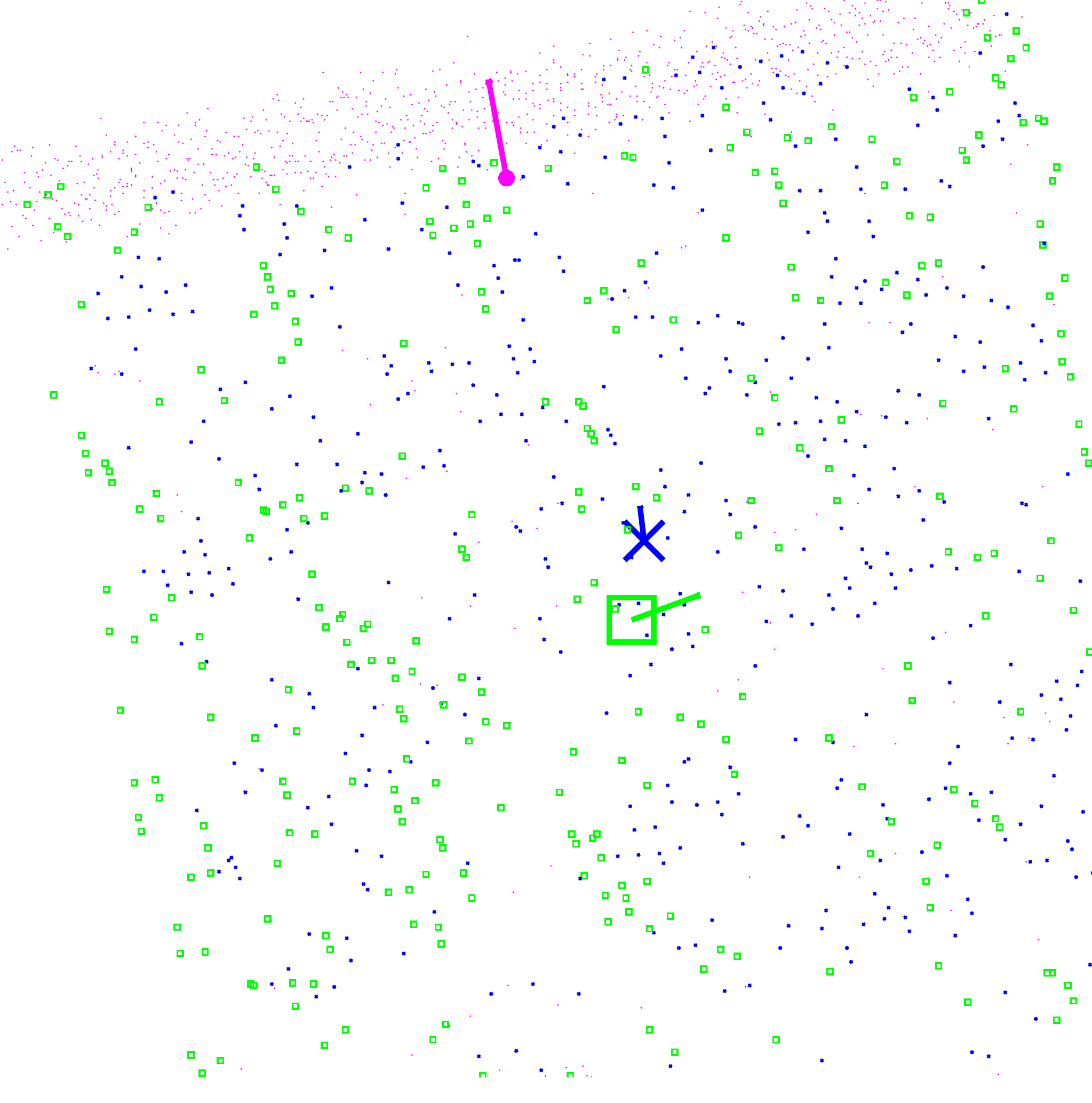}
      	\label{fig:exp:k-means_wrong2}
	\end{subfigure}

	\caption{Incorrect clustering due to noise, although the constraints are satisfied. The clusters satisfy the 90 degree constraint, but it is possible to see that the two clusters belong to the same plane. This situation occurs when the noise is excessively high and it deforms the surface greatly.}
	\label{fig:exp:k-means_wrong}
\end{figure}

For a specific amount of noise, the proposed clustering method for point clouds obtained an appropriate result. However, with a higher level of noise, special situations could occur such as those where the clusters satisfied the constraints of the model but they were actually incorrect. This situation is illustrated in Fig. \ref{fig:exp:k-means_wrong}. The left image shows the clustering results with three groups, where they are represented by pink dots oriented vertically. The other two clusters (blue "x" and green squares) belong to the same face but their normals are 88 degrees apart (image on the right). Thus, given the constraints on the cube model (Table \ref{tab:exp:model}) where all the angles must be 90 degrees, this shape satisfies the requirements. In the right image, it is possible to see the cluster centroids and normals. Hence, this cluster satisfies the constraints in the same manner as the right group, although their points actually fail to form the three faces of a cube. 

\subsection{MC-RANSAC experiments} \label{sec:exp:MCRANSAC}
In the second set of experiments, we tested the MC-RANSAC method (Section \ref{sec:MCRANSAC}). In order to obtain reliable evaluations of the results, the clustered faces used in this section were those extracted without noise, i.e., the ground truth from the previous section. As in the PCC experiments, we used eight views for each object and different levels of noise were added.

\begin{figure}[!htb]
	\centering
	\begin{subfigure}[b]{0.2\textwidth}
      	\includegraphics[width=\textwidth]{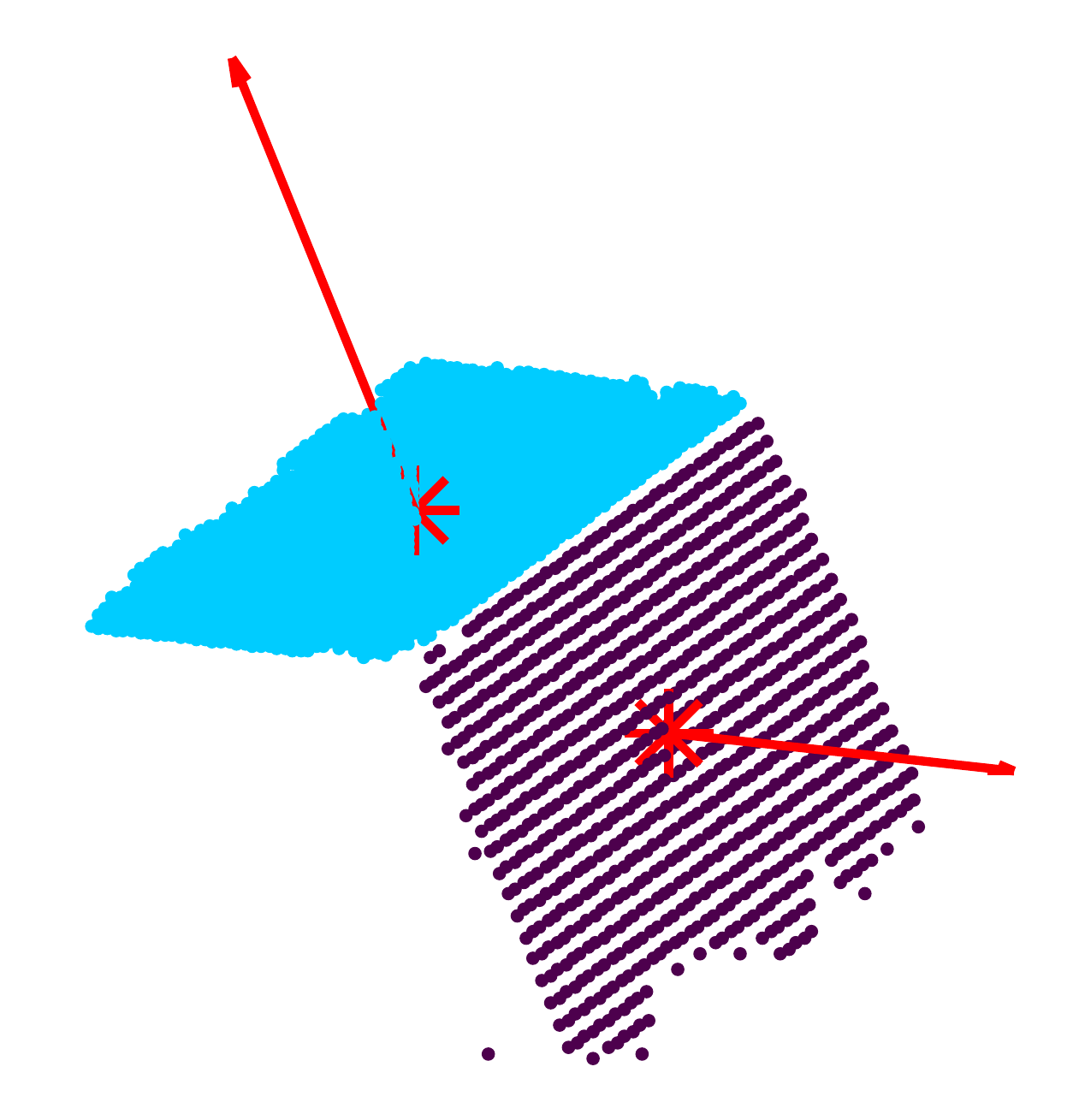} 
      	\label{fig:exp:ransac_ini1}
	\end{subfigure}
		\begin{subfigure}[b]{0.2\textwidth} 	
      	\includegraphics[width=\textwidth]{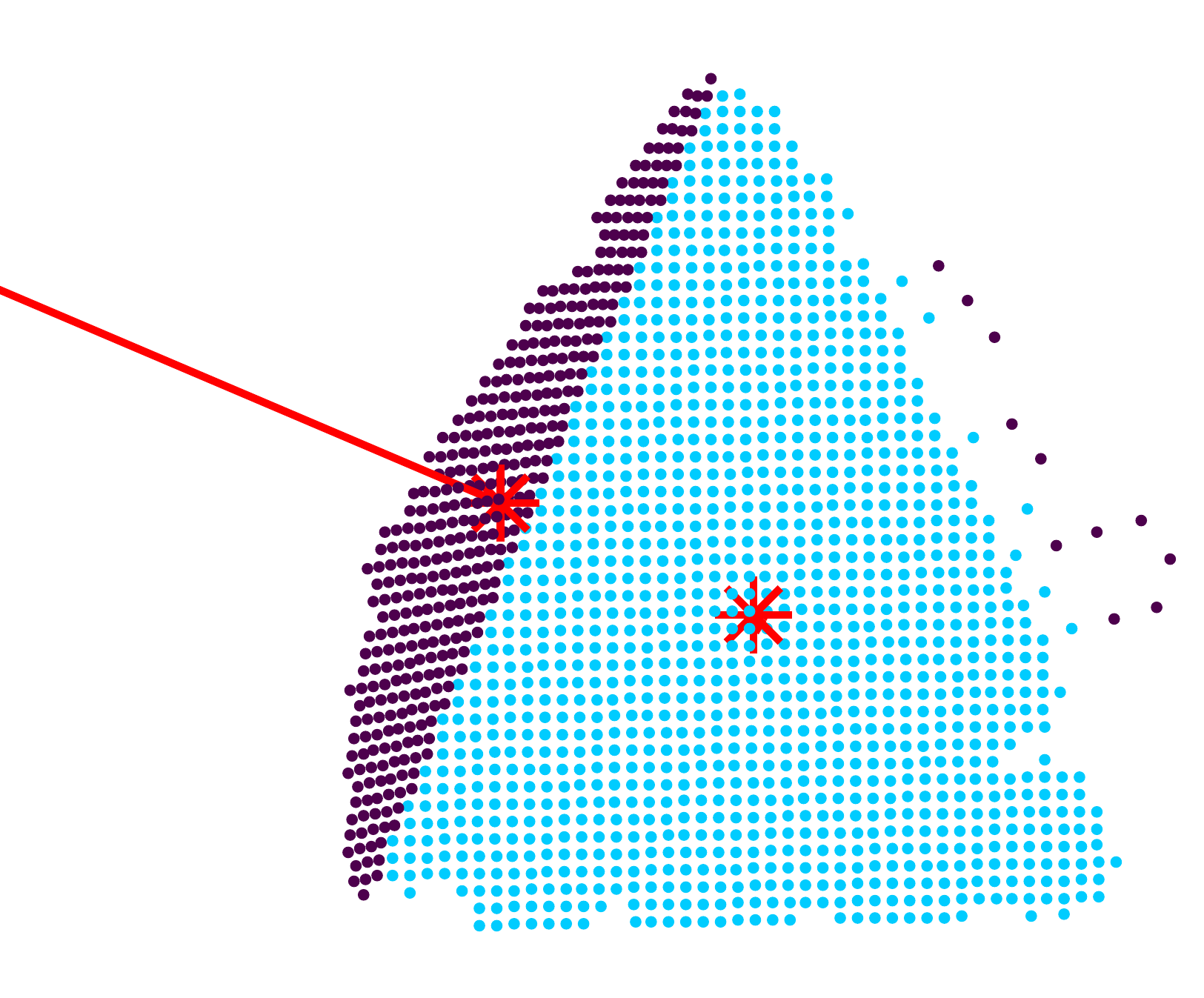}
      	\label{fig:exp:ransac_ini2}
	\end{subfigure}
			\begin{subfigure}[b]{0.2\textwidth} 	
      	\includegraphics[width=\textwidth]{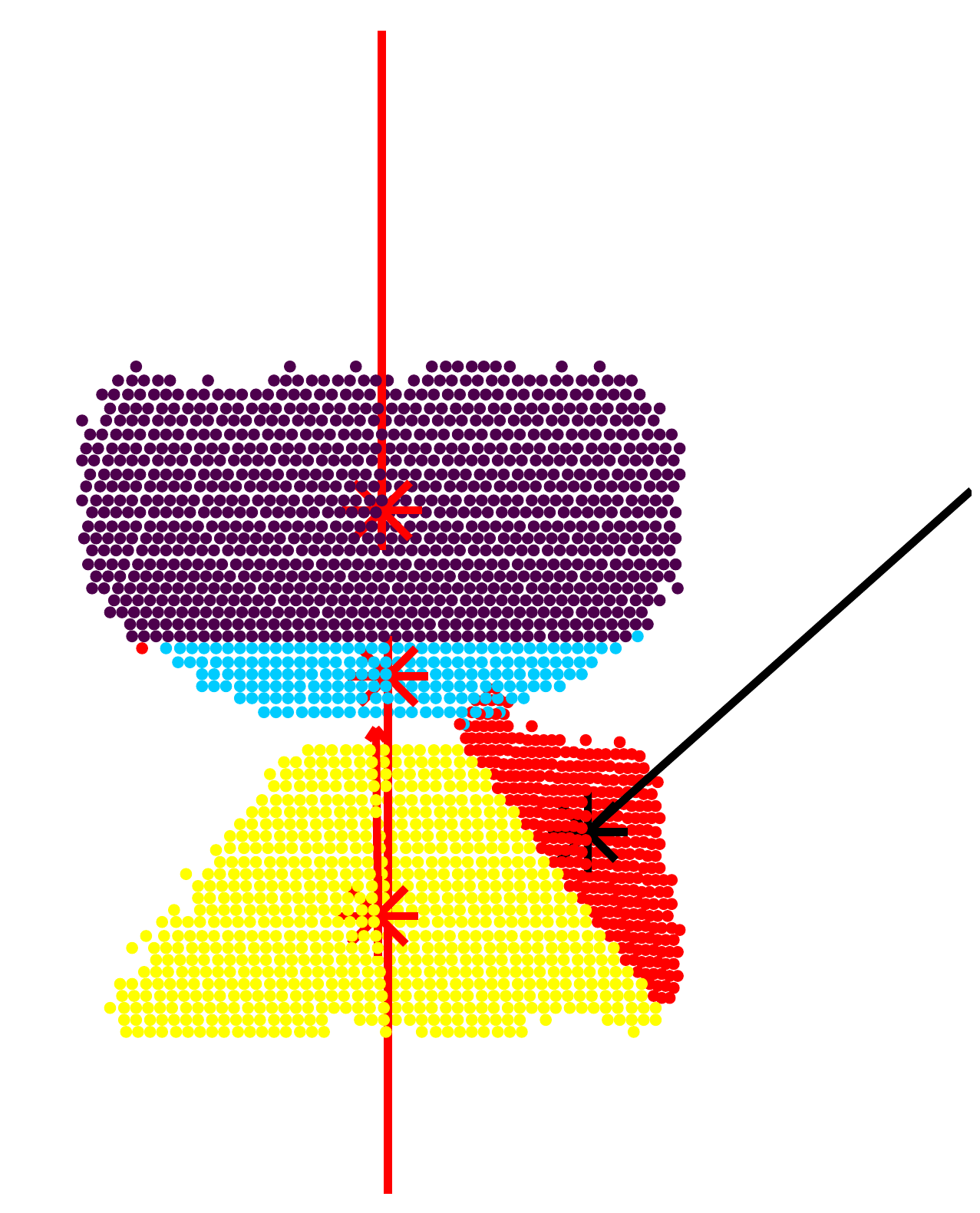}
      	\label{fig:exp:ransac_ini3}
	\end{subfigure}
	
	\begin{subfigure}[b]{0.2\textwidth} 	
      	\includegraphics[width=\textwidth]{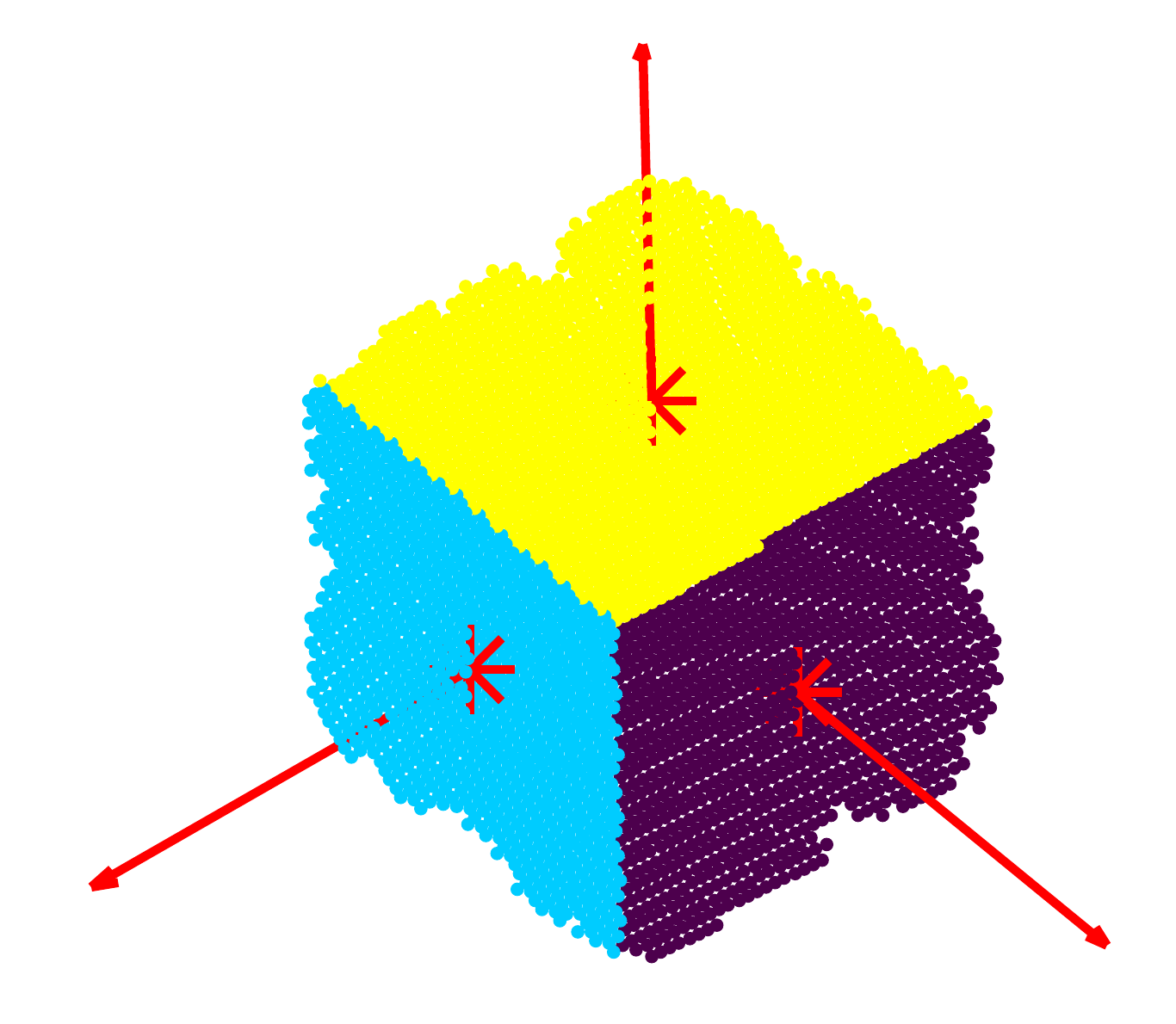}
      	\label{fig:exp:ransac_ini4}
	\end{subfigure}
	\begin{subfigure}[b]{0.2\textwidth} 	
      	\includegraphics[width=\textwidth]{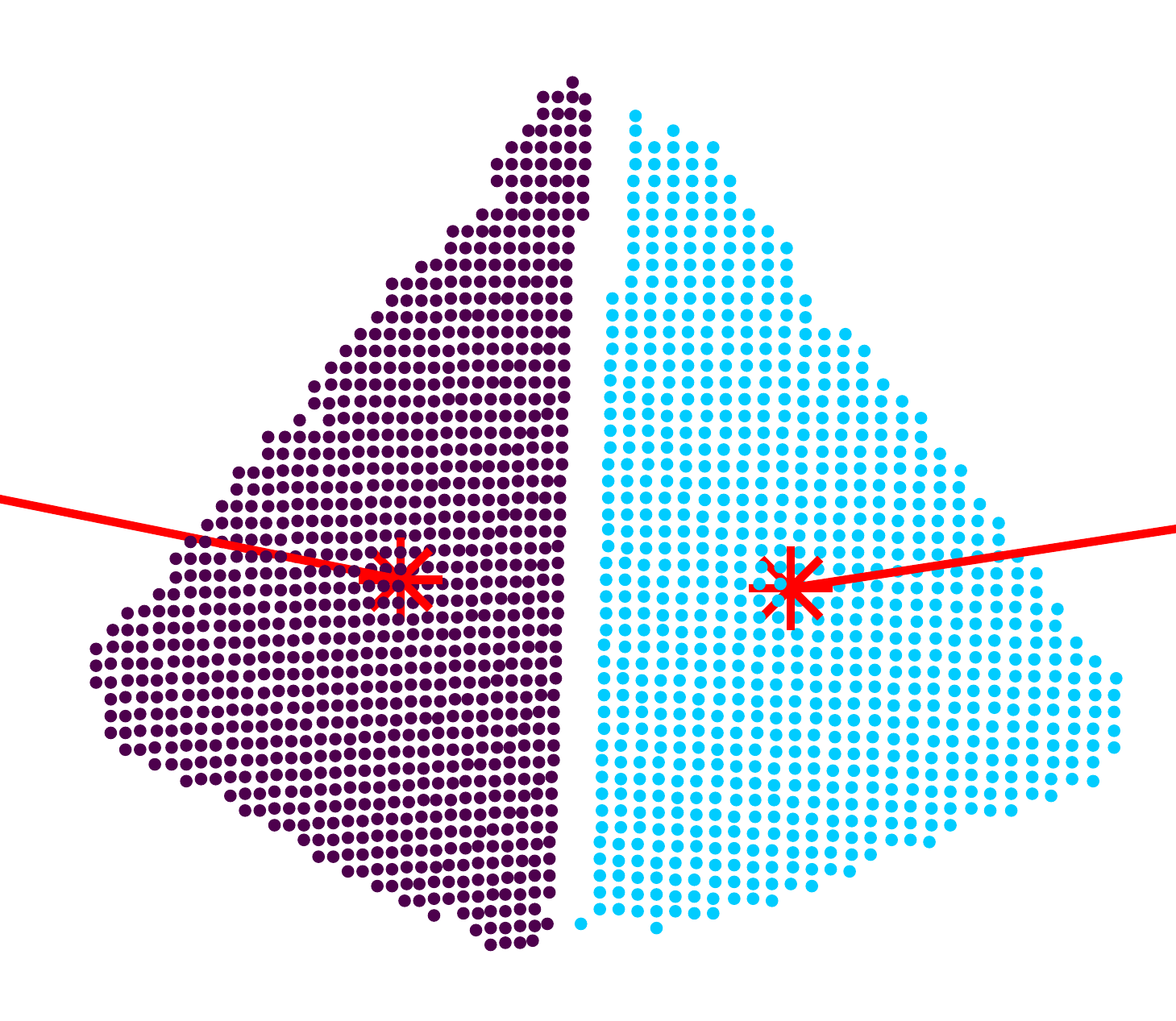}
      	\label{fig:exp:ransac_ini5}
	\end{subfigure}
	\begin{subfigure}[b]{0.2\textwidth} 	
      	\includegraphics[width=\textwidth]{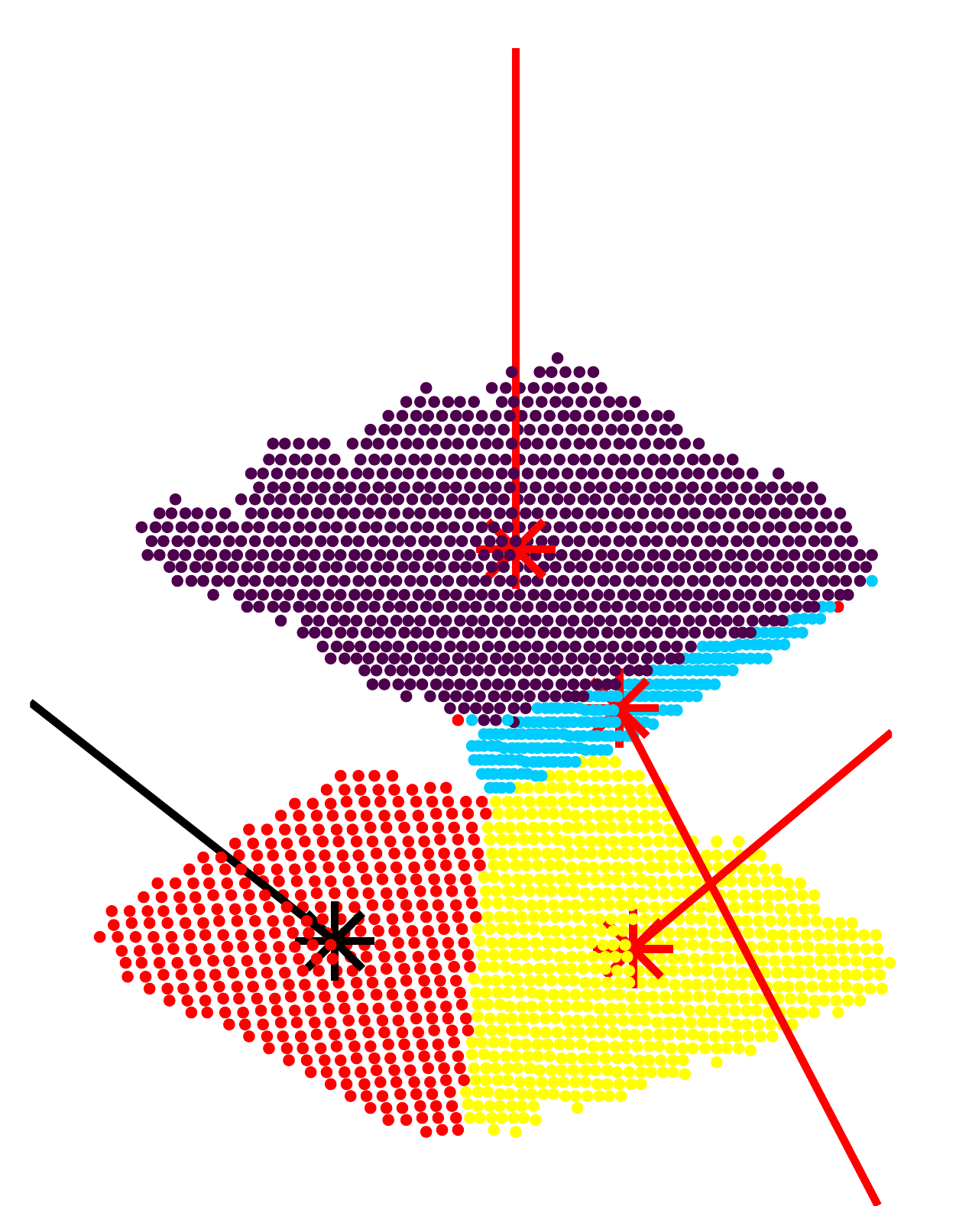}
      	\label{fig:exp:ransac_ini6}
	\end{subfigure}
	\caption{Results obtained by \textit{MC-RANSAC} with synthetic data.}
	\label{fig:exp:ransac_ini}
\end{figure}

The first experiment without noise allowed us to demonstrate the performance of the method. Figure \ref{fig:exp:ransac_ini} shows two of the eight views for each object in different positions. The coloured faces show the points used by MC-RANSAC to obtain the model. The asterisks represent the centroids of planes and the vectors are the normals. 

It is possible to evaluate the MC-RANSAC results visually. Furthermore, Table \ref{tab:exp:ransac_ang} shows quantitative values, which illustrate the correct results obtained by the MC-RANSAC method. For each object in the table, there are three columns: $\gamma$ is the mean difference between the angles of the estimated planes and the angles in the model (e.g., the cube has three planes, so in Fig.~\ref{fig:exp:ransac_ini}, with 90-88-91 degrees between the faces, $\gamma$ = $||90-90||+||90-88||+||90-91||/3 = (0+2+1)/3 = 1$), $\rho$ is the standard deviation of $\gamma$ and \textit{Planes} is the number of planes detected by the method for each view. The first eight rows show the values for each of the eight views. The final row shows the mean values for the angle means and standard deviations.  

\begin{table}[h!]
\centering
\begin{tabular}{|c||c|c|c|c|c|c|c|c|c|}
\hline
View & \multicolumn{3}{|c|}{Cube} & \multicolumn{3}{|c|}{Pyramid} & \multicolumn{3}{|c|}{Double Pyramid} \\
\hline
 & $\gamma$ & $\rho$ & Planes & $\gamma$ & $\rho$ & Planes & $\gamma$ & $\rho$ & Planes \\
\hline

1  	& 0.0003 & 0      & 2 & 0.9999 & 0 & 2  & 0.2012 & 0.2424 & 4\\

2 	& 0.0002 & 0      & 2 & 1      & 0 & 2  & 0.0918 & 0.1275 & 4\\

3  	& 0.0001 & 0      & 2 & 1.539  & 0 & 2  & 0.1118 & 0.1240 & 4\\

4 	& 0.2187 & 0.16   & 3 & 1.537  & 0 & 2  & 0.0989 & 0.1100 & 4\\
    
5 	& 0.0107 & 0.0139 & 3 & 1.5369 & 0 & 2  & 0.0310 & 0.0341 & 4\\
	
6   & 0.0048 & 0.0071 & 3 & 1.5369 & 0 & 2  & 0.0081 & 0.0152 & 4 \\
      
7 	& 0.0073 & 0.0028 & 3 & 1.537  & 0 & 2  & 0.0103 & 0.0106 & 4 \\
       
8 	& 0.0183 & 0.0238 & 3 & 1.5369 & 0 & 2  & 0.1189 & 0.1644 & 4\\

\hline
Mean & 0.0031 & 0.0259  & & 1.3805 & 0 & & 0.0530 & 0.1035 & \\
\hline
\end{tabular}
\caption{Mean angles in degrees between all of the planes for each view and the corresponding model. The number of planes detected is shown for each view and object.}
\label{tab:exp:ransac_ang}
\end{table}

These values demonstrate the excellent performance of the method. Next, the effects of different levels of Gaussian noise were evaluated based on comparisons with the ground truth clustering results. In order to compare our method, we performed experiments using the proposed MC-RANSAC method, clustered RANSAC and the original version. Clustered RANSAC is based on CC-RANSAC \cite{Gallo2011}, but the clusters obtained by the ground truth are used to compare the results. Traditional RANSAC estimates planes iteratively until no points are left. In many cases, this method returns incoherent planes due to its dependence on the initial random seed and the threshold used for detecting inliers. Therefore, the results presented in this subsection correspond only to MC-RANSAC and clustered RANSAC because in most cases, the number of planes obtained by the traditional RANSAC was not similar to that expected and their orientations were dissimilar. 

\begin{figure}[!htb]
	\centering
	\begin{subfigure}[b]{0.6\textwidth}
      	\includegraphics[width=\textwidth]{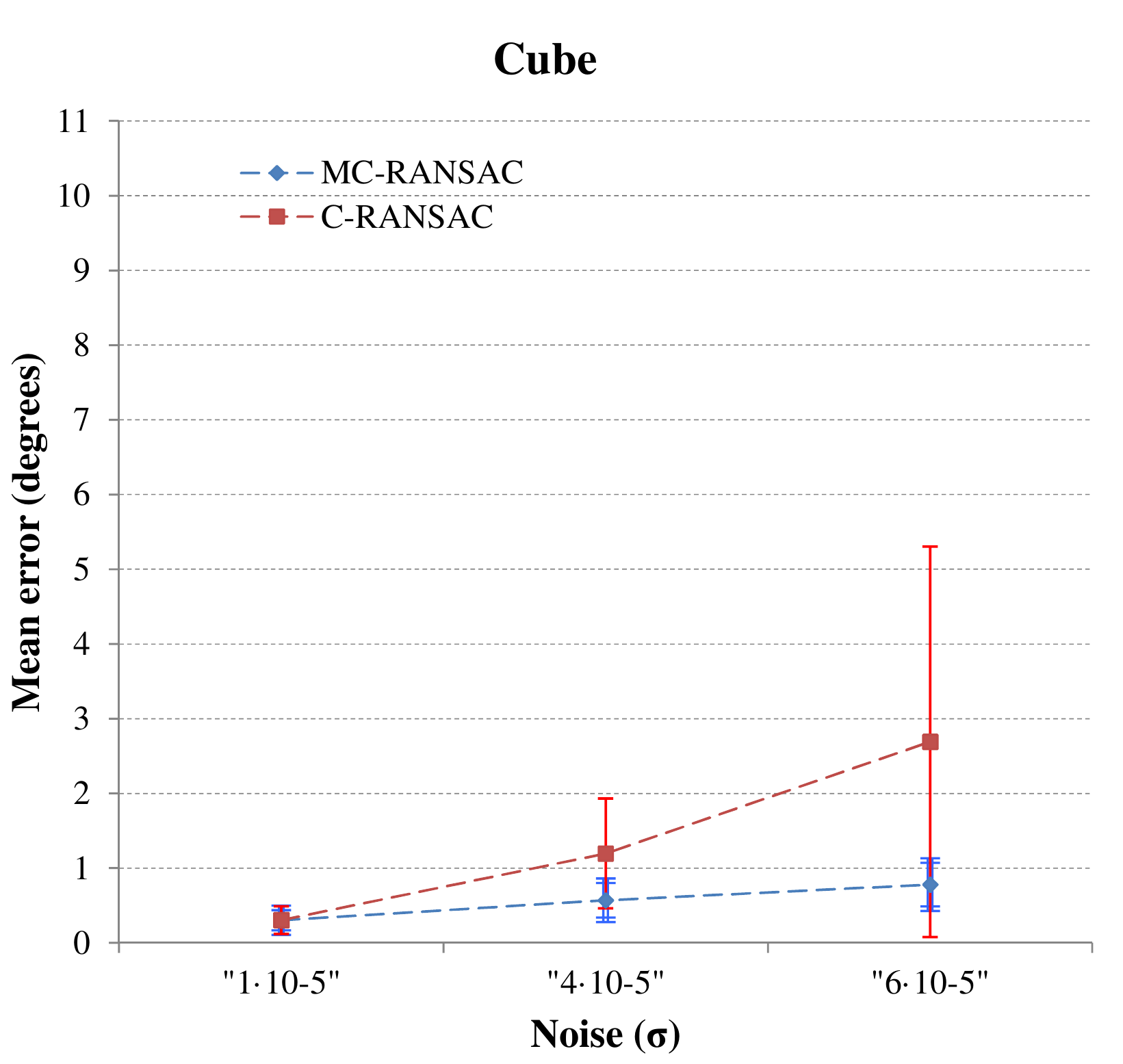} 
      	\label{fig:exp:ransac_error_11}
	\end{subfigure}
	
	\begin{subfigure}[b]{0.6\textwidth} 	
	\vspace{-1cm}
      	\includegraphics[width=\textwidth]{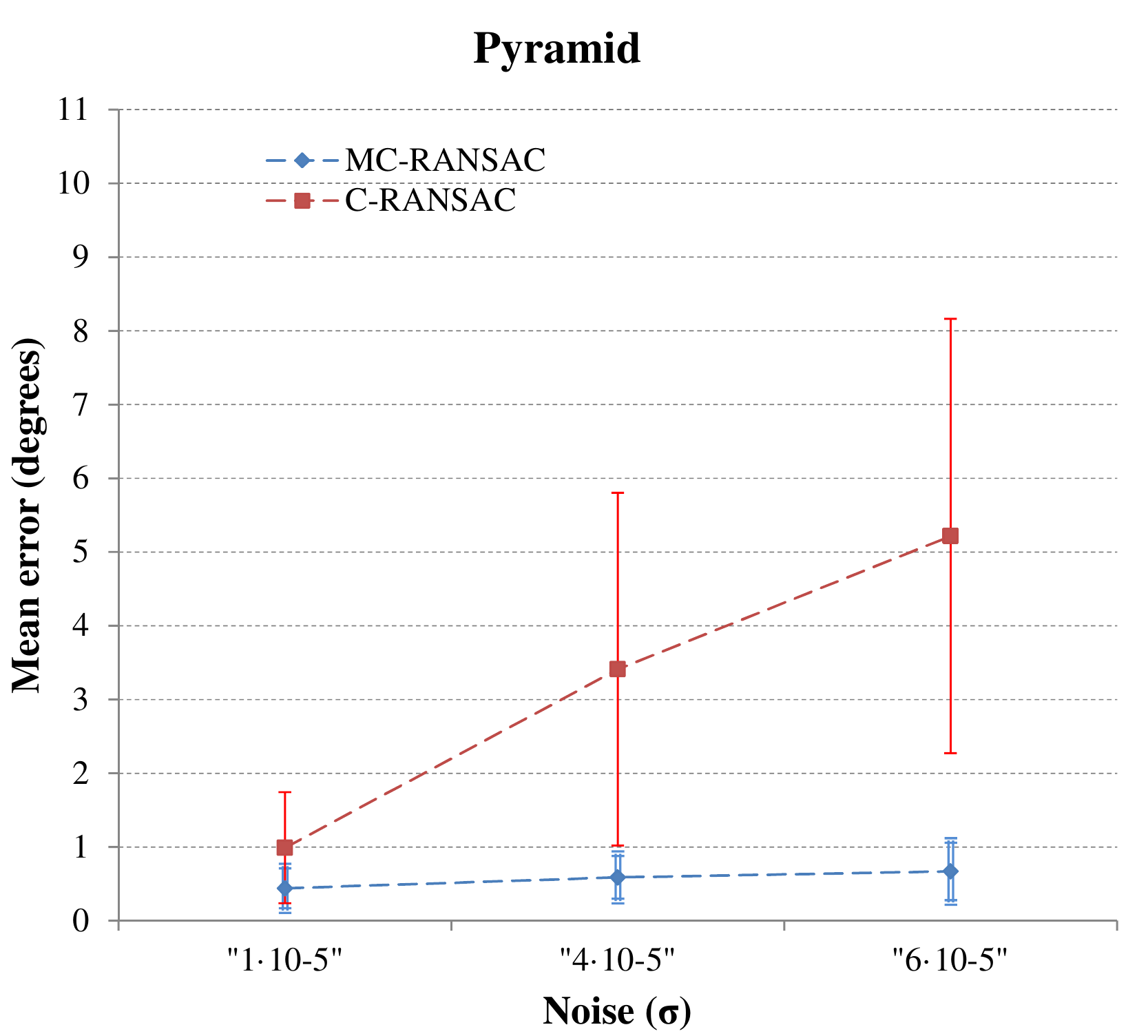}
      	\label{fig:exp:ransac_error_12}
	\end{subfigure}
	\vspace{-1cm}
\caption{Graphical representation of the errors in Table \ref{tab:exp:ransac_mcransac} for the cube and pyramid. The blue diamonds denote the errors with MC-RANSAC (first rows in each pair in the table) and red square marks denote those with C-RANSAC. The mean of the error and the associated standard deviation are presented for each level of noise.}
	\label{fig:exp:ransac_error_1}
\end{figure}

\begin{figure}[!htb]
	\centering
	\begin{subfigure}[b]{0.6\textwidth} 	
      	\includegraphics[width=\textwidth]{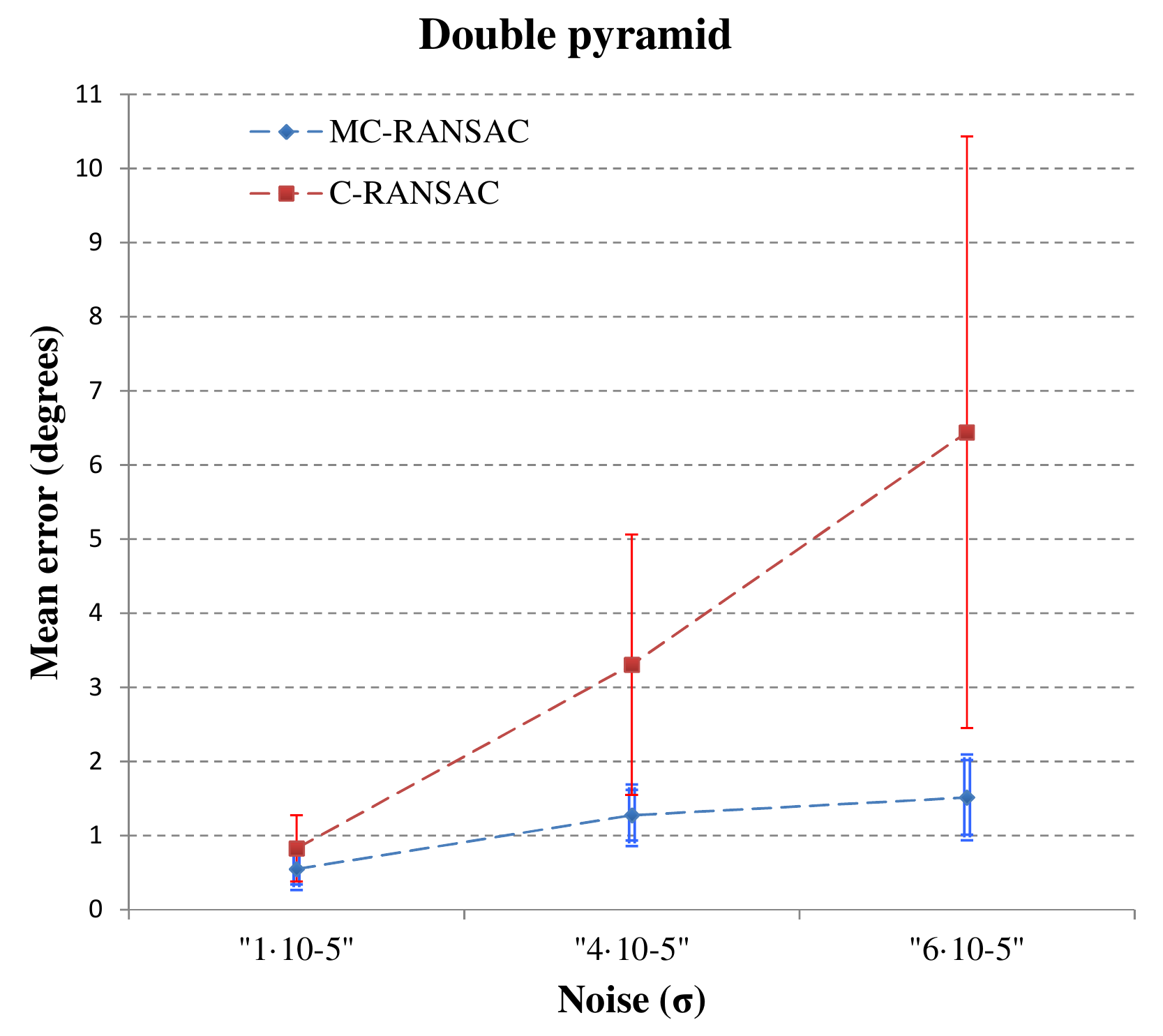}
      	\label{fig:exp:ransac_error_13}
	\end{subfigure}

\caption{Graphical representation of the errors in Table \ref{tab:exp:ransac_mcransac} for the double pyramid, as an extension of Figure~\ref{fig:exp:ransac_error_1}. The blue diamonds denote the errors with MC-RANSAC (first rows in each pair in the table) and red square marks denote those with C-RANSAC. The mean of the error and the associated standard deviation are presented for each level of noise.}
	\label{fig:exp:ransac_error_1b}
\end{figure}

In this experiment, we applied each algorithm five times with different levels of noise and various views. The results represent the mean values of all views based on five iterations. Table~\ref{tab:exp:ransac_mcransac} shows the mean values of the angles between the plane orientations and the model, which were extracted by MC-RANSAC and clustered RANSAC for each noise level (each pair of rows shows the MC-RANSAC error and clustered RANSAC error, respectively). The table shows that better results were obtained by the proposed method in terms of both the mean and standard deviation. Figures \ref{fig:exp:ransac_error_1} and \ref{fig:exp:ransac_error_1b} show the means and the standard deviations from the tables, where the improvement obtained using MC-RANSAC compared with clustered RANSAC is obvious. These figures show that the error increased slightly with the proposed method, which is represented by the blue line with diamonds, compared with clustered RANSAC (represented by the red line with squares), for which the error increased greatly with the noise. The accuracy of the proposed method decreased slightly whereas the accuracy of clustered RANSAC decreased greatly. The standard deviation is represented by the error bar for each point, which shows that the stability of the proposed MC-RANSAC was higher than that of clustered RANSAC, thereby demonstrating that more reliable and repeatable results were produced by MC-RANSAC. 

\begin{table}[htbp]
  \centering

    \begin{tabular}{lrrrrrrr}
    \hline
    Method & & \multicolumn{2}{c}{Cube} & \multicolumn{2}{c}{Pyramid} & \multicolumn{2}{c}{Double pyramid} \\
    \hline
     & $\sigma~noise$ & \multicolumn{1}{c}{$\gamma$} & \multicolumn{1}{c}{$\rho$} & \multicolumn{1}{c}{$\gamma$} & \multicolumn{1}{c}{$\rho$} & \multicolumn{1}{c}{$\gamma$} & \multicolumn{1}{c}{$\rho$} \\
    MC-R &    "$1\cdot10^{-5}$" & \textbf{0.30123} & \textbf{0.16632} & \textbf{0.43868} & \textbf{0.30187} & \textbf{0.5477} & \textbf{0.24398} \\
    Clust-R & "$1\cdot10^{-5}$" & 0.30324 & 0.18505 & 0.99062 & 0.75371 & 0.82737 & 0.44527 \\
    \hline
    MC-R &     "$4\cdot10^{-5}$" & \textbf{0.57041} & \textbf{0.26056} & \textbf{0.58772} & \textbf{0.32025} & \textbf{1.27398} & \textbf{0.37951} \\
     Clust-R & "$4\cdot10^{-5}$" & 1.19646 & 0.73382 & 3.41070 & 2.39174 & 3.30565 & 1.75562 \\
	\hline
    MC-R & "$6\cdot10^{-5}$" & \textbf{0.77933} & \textbf{0.32336} & \textbf{0.66831} & \textbf{0.41879} & \textbf{1.51525} & \textbf{0.54215} \\
     Clust-R & "$6\cdot10^{-5}$" & 2.69095 & 2.61422 & 5.21775 & 2.94591 & 6.44233 & 3.99118 \\
    \hline
    \end{tabular}%
   \caption{Results obtained by MC-RANSAC (MC-R) and Clustered RANSAC(Clust-R) in terms of model angle fitting, where the means (represented by $\gamma$) and standard deviations (represented by $\rho$) of the angles between the model and the planes were estimated for different levels of noise. For each pair of rows, the best result is marked in bold. Clearly, the proposed MC-RANSAC method obtained better results for each object and noise level in terms of both the mean and standard deviation.}
   \label{tab:exp:ransac_mcransac}%
\end{table}

\begin{figure}[h!]
  \centering
    \includegraphics[width=0.5\textwidth]{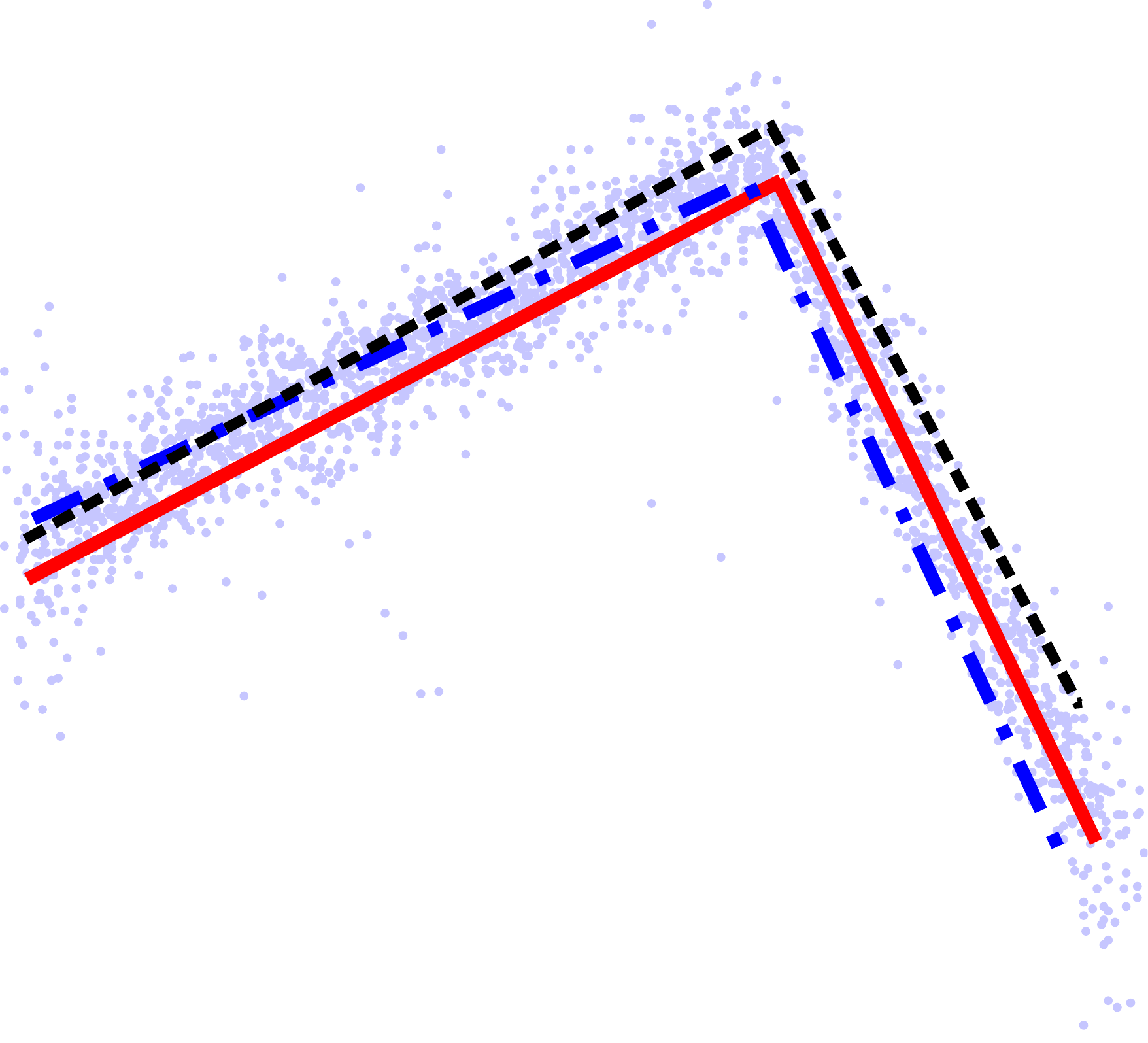}
  \caption{Example showing three pair of planes that satisfied the constraints of the cube model, but which were oriented in a slightly different manner.}
  \label{fig:exp:planes_example}
\end{figure}

The results of these experiments demonstrate how well the plane orientations satisfied the model constraints. However, for the same sample of points, many different combinations of planes could satisfy the constraints (as shown in Figure \ref{fig:exp:planes_example}, where three pair of planes satisfy the 90 degrees constraint on the points in the cube). 

Therefore, we performed further experiments to evaluate how well the planes agreed with the ground truth plane, i.e., instead of considering the constraints, we addressed the difference in the angle between a plane orientation estimated by MC-RANSAC and by the clustered RANSAC versus the same plane in the ground truth.

\begin{figure}[!htb]
	\centering
	\begin{subfigure}[b]{0.6\textwidth}
      	\includegraphics[width=\textwidth]{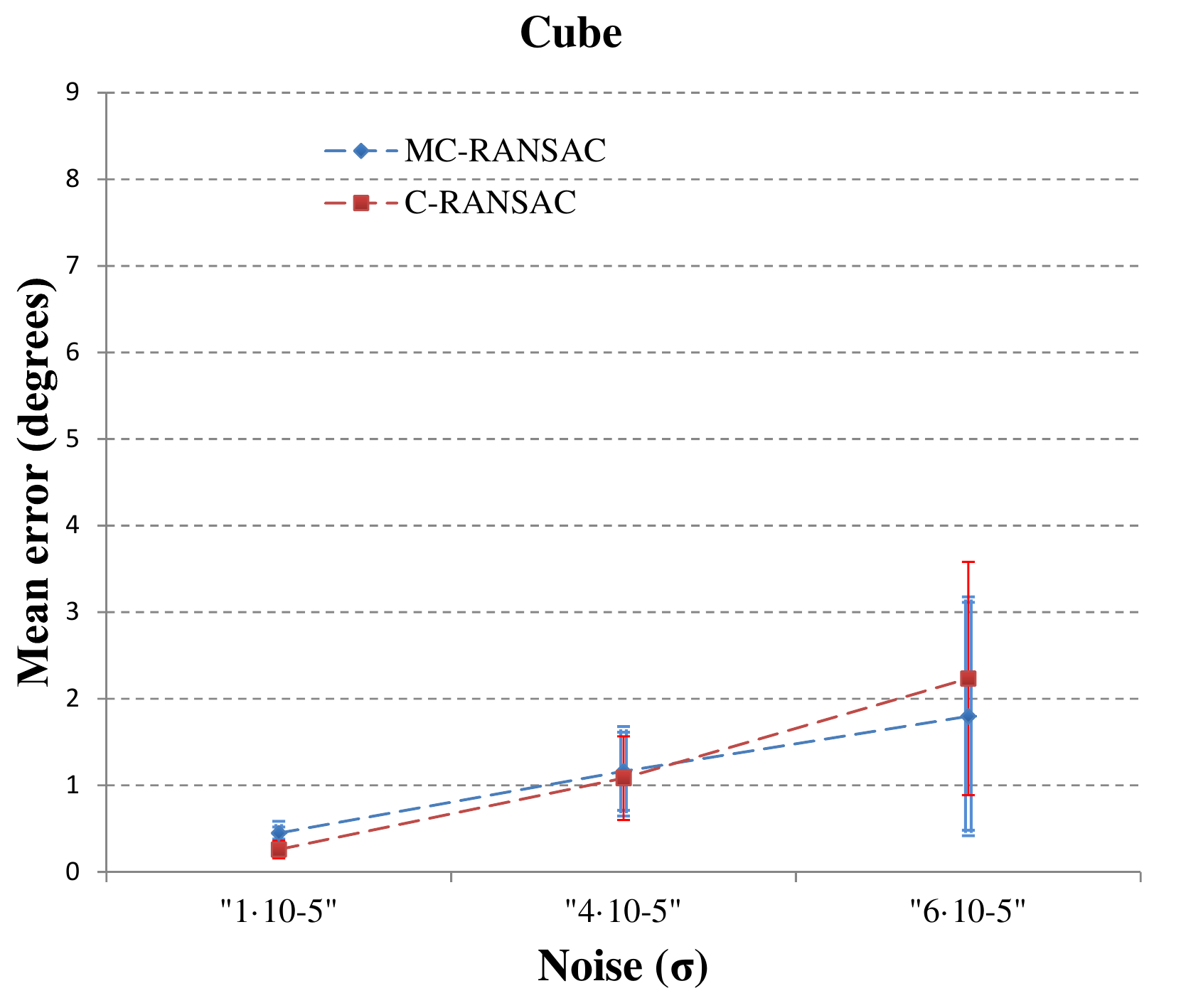} 
      	\label{fig:exp:ransac_error_21}  	
 	\end{subfigure}

	\begin{subfigure}[b]{0.6\textwidth} 
	\vspace{-1cm}	
      	\includegraphics[width=\textwidth]{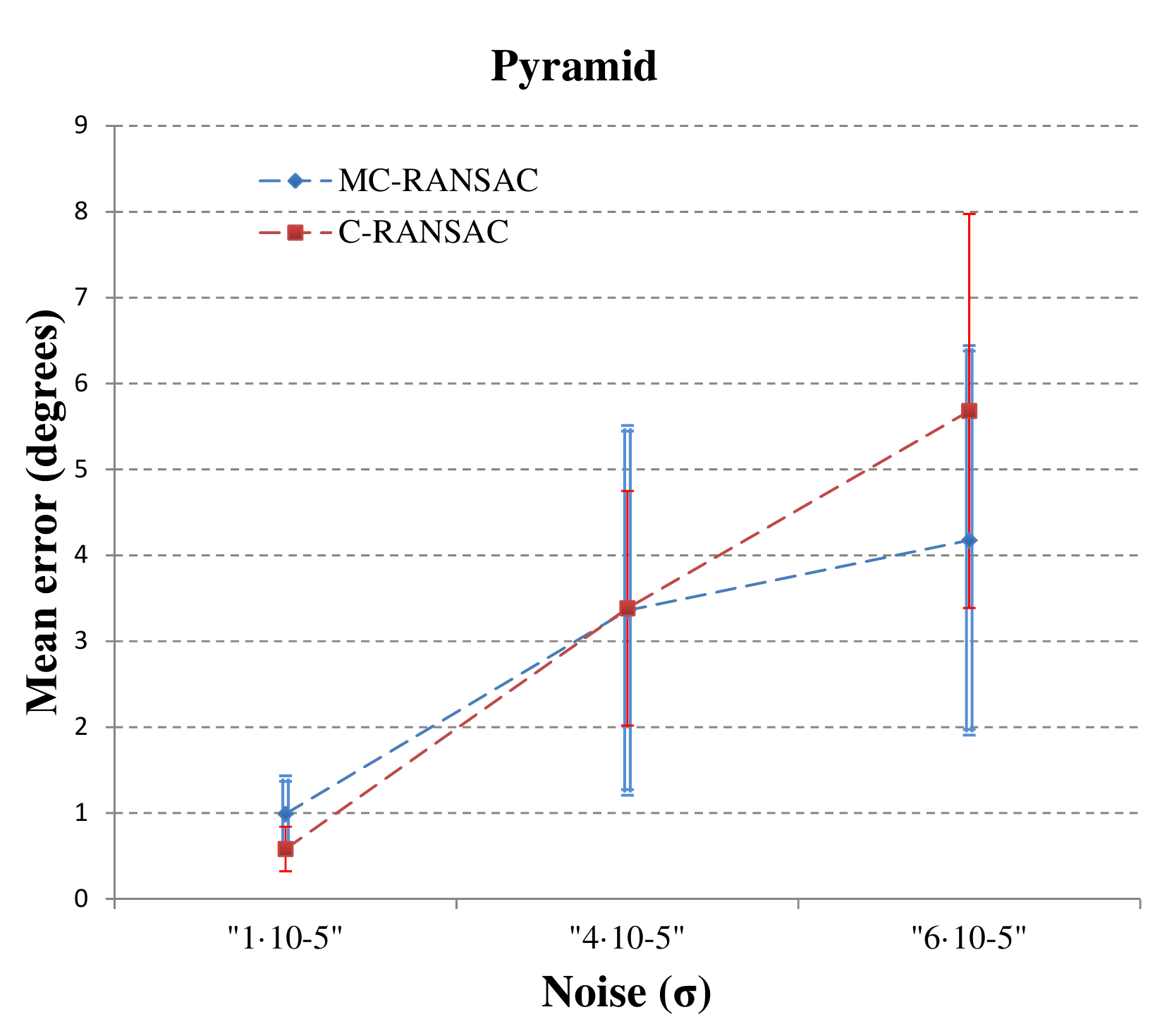}
      	\label{fig:exp:ransac_error_22}
	\end{subfigure}
\vspace{-1cm}		
	\caption{Graphical representation of the errors shown in Table \ref{tab:exp:mcransac_orient} for the cube and pyramid. The blue diamonds represents MC-RANSAC (first rows in each pair in the table) and the red squares denote clustered RANSAC. The mean of the error and the associated standard deviation are presented for each noise level.}
	\label{fig:exp:ransac_error_2}
\end{figure}

\begin{figure}[h]
	\centering	
	\begin{subfigure}[b]{0.6\textwidth} 	
      	\includegraphics[width=\textwidth]{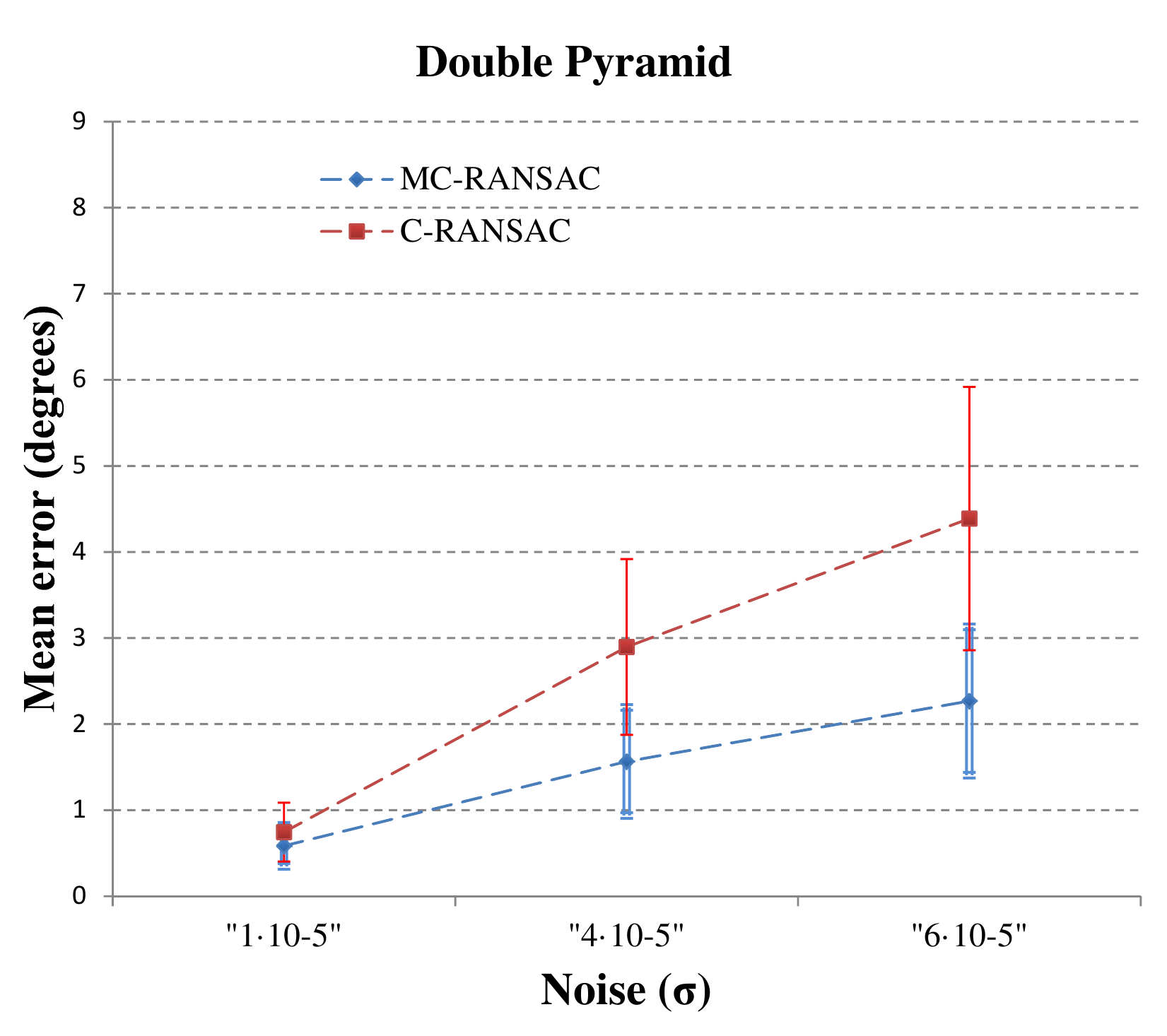}
      	\label{fig:exp:ransac_error_23}
	\end{subfigure}

	\caption{Graphical representation of the errors shown in Table \ref{tab:exp:mcransac_orient} for the double pyramid, as an extension of Figure~\ref{fig:exp:ransac_error_2}. The blue diamonds denote MC-RANSAC (first rows in each pair in the table) and the red squares denoted C-RANSAC. The mean of the error and the associated standard deviation are presented for each noise level.}
	\label{fig:exp:ransac_error_2b}
\end{figure}

Table~\ref{tab:exp:mcransac_orient} shows the mean values and the standard deviations of the angles between the planes and the corresponding angles in the ground truth for MC-RANSAC and clustered RANSAC. This table shows that clustered RANSAC obtained slightly better results when the number of planes was low (pyramid) and the noise level was also low because MC-RANSAC is more focused on providing a good fit to the constraints instead of only fitting the points. However, when the noise level was higher or the number of planes increased, the performance of MC-RANSAC was better (represented by the mean $\gamma$) and more stable (represented by the standard deviation $\rho$). Figures~\ref{fig:exp:ransac_error_2} and \ref{fig:exp:ransac_error_2b} show the information contained in the table in a graphical and intuitive manner. The red squares represent the mean error for each noise level with clustered RANSAC and the blue diamonds correspond to the results using MC-RANSAC. The standard deviation is plotted for each error value with error bars. In all cases, MC-RANSAC was more accurate as the noise increased.

\begin{table}[htbp]
  \centering
    \begin{tabular}{lrrrrrrr}
    \hline
     method&     & \multicolumn{2}{c}{Cube} & \multicolumn{2}{c}{Pyramid} & \multicolumn{2}{c}{Doublepyramid} \\
    \hline
     & $\sigma~noise$& \multicolumn{1}{c}{$\gamma$} & \multicolumn{1}{c}{$\rho$} & \multicolumn{1}{c}{$\gamma$} & \multicolumn{1}{c}{$\rho$} & \multicolumn{1}{c}{$\gamma$} & \multicolumn{1}{c}{$\rho$}  \\
    MC-R &$1\cdot10^{-5}$ &             0.44847 & 0.3220 & 0.99007 & 0.41041 & \textbf{0.58490} & \textbf{0.23920} \\
    Clust-R & $1\cdot10^{-5}$ & \textbf{0.25930} & \textbf{0.10268} & \textbf{0.58107} &\textbf{ 0.25939} & 0.74456 & 0.34151 \\
    \hline
    MC-R &$4\cdot10^{-5}$ & 1.16246 & 0.68947 & \textbf{3.35888} & 2.11948 & \textbf{1.56576} & \textbf{0.62812} \\
    Clust-R & $4\cdot10^{-5}$ & \textbf{1.08260} & \textbf{0.48365} & 3.38369 & \textbf{1.36565} & 2.89609 & 1.02107 \\
    \hline
    MC-R & $6\cdot10^{-5}$ & \textbf{1.79764} & \textbf{0.83456} & \textbf{4.17551} & \textbf{2.23639} & \textbf{2.26828} & \textbf{0.86087} \\ 
	Clust-R & $6\cdot10^{-5}$ & 2.23349 & 1.34749 & 5.68080 & 2.29415 & 4.38794 & 1.52995 \\
	\hline
    \end{tabular}%
	\caption{Results obtained by MC-RANSAC (MC-R) and clustered RANSAC (Clust-R) compared with the ground truth plane orientations. The means (represented by $\gamma$) and standard deviations (represented by $\rho$) for the angle orientations are shown for each element and noise level.}
  \label{tab:exp:mcransac_orient}%
\end{table}

The results presented in the tables and figures above demonstrate the superior performance of MC-RANSAC compared with state-of-the-art methods in terms of the accuracy when estimating the planes for various sets of points simultaneously. This improvement is very important for systems where the alignment of the planes must be highly precise, or in fine registration systems where accuracy is a key requirement. 

\subsection{PCC and MC-RANSAC}
The overall system based on PCC (Subsection~\ref{sec:PCC}) and MC-RANSAC (Subsection~\ref{sec:MCRANSAC}) can be used to estimate the best planes that satisfy a set of points, while preserving a group of constraints. PCC clusters each set of points and MC-RANSAC estimates the planar models.

The combination of both methods obtains the planar model, but it also helps to address other issues in various situations, particularly the detection of false positives during the PCC step. If we obtain an incorrect group of clusters due to an error in the PCC, as shown in Figure~\ref{fig:exp:k-means_wrong}, then clustered RANSAC will produce the wrong planes, whereas the constraints will  not be satisfied in MC-RANSAC during the checking step. This helps to determine whether the set of clustered points can form the expected model. 

\begin{figure}[!htb]
	\centering
    \includegraphics[width=0.5\textwidth]{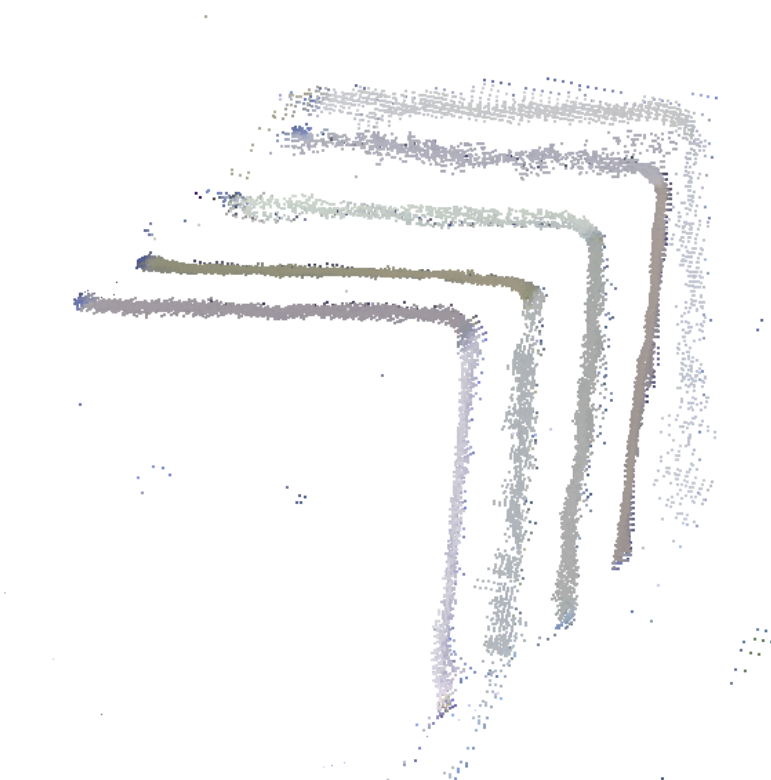} 
    \caption{Different views of two sides of a cube, where it is possible to appreciate the errors in the angles of the sides caused by the incorrect calibration of the camera.}
    \label{fig:exp:cube_calibration}
\end{figure}

Another interesting advantage of the proposed MME method is that it can reduce the effects of incorrect camera calibration in some applications. Figure~\ref{fig:exp:cube_calibration} shows the effects of incorrect camera calibration. These views correspond to the two sides of a cube from different viewpoints (they are aligned to facilitate visualization of the problem), where it is easy to see that the angle is not 90 degrees. This problem was caused by the incorrect calibration of the camera, which could be minimized using the proposed method because the constraints require that the results are within a specific range.

\section{Conclusions}
In this study, we proposed a novel method for planar model reconstruction from 3D point clouds. The MME method facilitates the accurate reconstruction of planar objects using prior knowledge related to the model. The MME method employs PCC and MC-RANSAC. First, the points that belong to each face of the object are evaluated, before estimating the planes that best fit the points and the constraints of the model. The PCC method uses a k-means algorithm to estimate the clusters and a tree search technique to refine the solutions while considering the prior constraints. The MC-RANSAC extends traditional RANSAC by considering pre-clustered input data as well as introducing a novel step that evaluates whether the inliers comply with the prior constraints.

We evaluated the proposed method for different objects (cube, pyramid and double pyramid) and different noise levels. The point clouds of the objects were obtained using a Microsoft Kinect sensor and a Blensor camera simulator. The PCC method obtained accurate clustering results, even with high noise levels and a complex planar object (i.e., the double pyramid). We compared the performance of MC-RANSAC with the state-of-the-art clustering version of RANSAC, where the results showed that the proposed method performed better in most of the experiments. The constraint fitting experiments obtained better results in all cases. In the plane orientation experiments, the proposed method outperformed clustered RANSAC as the amount of noise or the number of planes in the model increased.   

The main weakness of the MME is the processing time required for the PCC step, which is very interesting and future research should include an analysis based on a parallel design of the method. Moreover, experiments with different sets of constraints could be studied in future research, such as using coloured faces in the clustering step to improve the plane estimation. In our future research, we also aim to evaluate the effect of the proposed method in registering markers reconstructed using MME and mapping based on knowledge related to the main planes in a scene (e.g., the walls and floor).  

\section{References}
\textbf{}
\bibliographystyle{elsarticle-num}
\bibliography{./RANSACbib}

\section{Bio}
\textbf{Marcelo Saval-Calvo}

\begin{figure}[h!]
  \centering
    \includegraphics[width=0.3\textwidth]{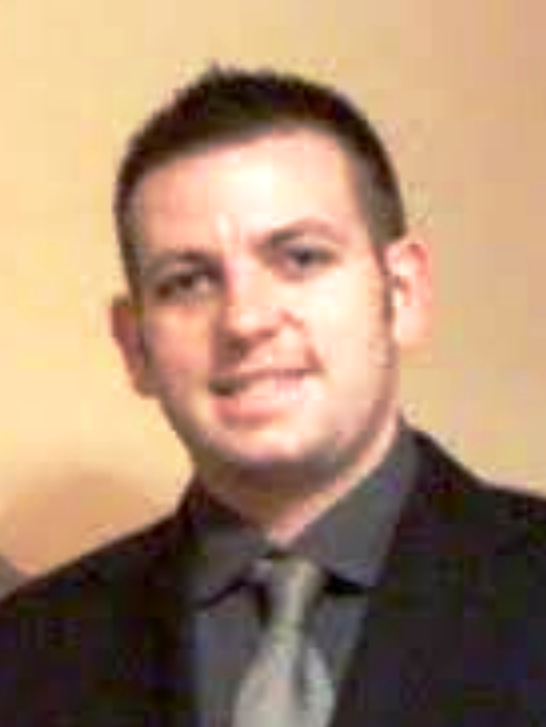}
  
  \label{fig:saval}
\end{figure}

Mr Marcelo Saval-Calvo received a degree in Computer Engineering in 2010 from the University of Alicante (Spain). In 2011, he received his MSc from the same university with a Master's Thesis in the area of Computer Vision. Currently, he is a PhD candidate working on 3D point cloud deformable registration funded by the Valencian Government. He was awarded a research fellowship by the Regional Ministry of Education, Culture and Sports of the Valencian Government for research at the University of Edinburgh. His current research interests include human behaviour analysis using computer vision, 3D deformable registration and model-based rigid registration.

\textbf{Jorge Azorin-Lopez}

\begin{figure}[h!]
  \centering
    \includegraphics[width=0.3\textwidth]{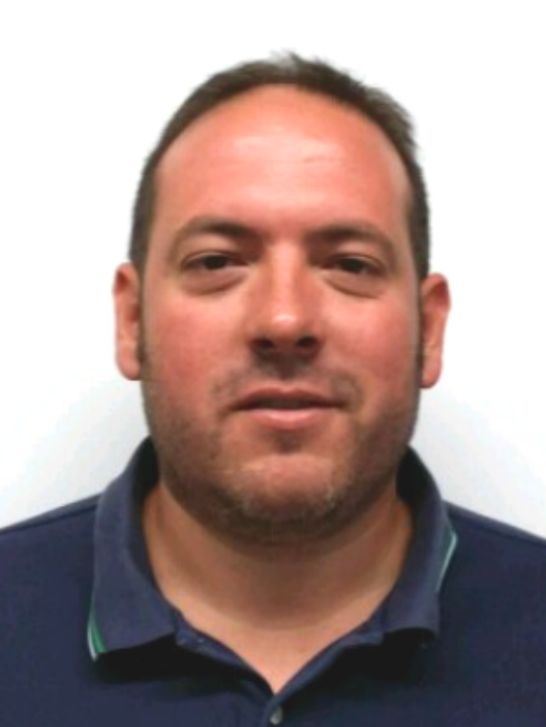}
  
  \label{fig:azorin}
\end{figure}

Dr Jorge Azorin-Lopez received a PhD degree in Computer Science from the University of Alicante (2007). Since 2001, he has been a faculty member in the Computer Technology department at the same university, where he is currently the Deputy Director of Research. He was awarded a post-doctoral research fellowship by the Spanish Government in 2009. He has worked on 14 research projects and published over 40 papers on computer vision in several journals, conferences and book chapters. He has served as a reviewer for numerous scientific journals and international conferences. His current areas of research include 3D vision systems.

\textbf{Andres Fuster-Guillo}
\begin{figure}[h!]
  \centering
    \includegraphics[width=0.3\textwidth]{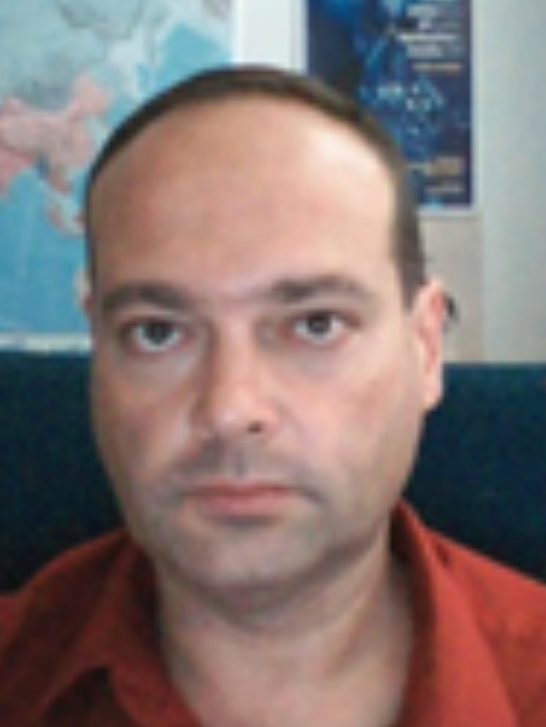}
  
  \label{fig:fuster}
\end{figure}

Dr Andres Fuster Guillo received his B.S. in Computer Science Engineering from the University of Valencia (1995) and his PhD in Computer Science from the University of Alicante (2003). Since 1997, he has been a member of the faculty in the Computer Technology department at the same university, where he is currently a professor. He has worked on more than 15 research projects and has published over 40 papers on computer vision and computer architecture. His current areas of research include vision systems for perception under adverse conditions, 3D vision systems, automated visual inspection and artificial neural networks.

\textbf{Jose Garcia-Rodriguez}

\begin{figure}[h!]
  \centering
    \includegraphics[width=0.3\textwidth]{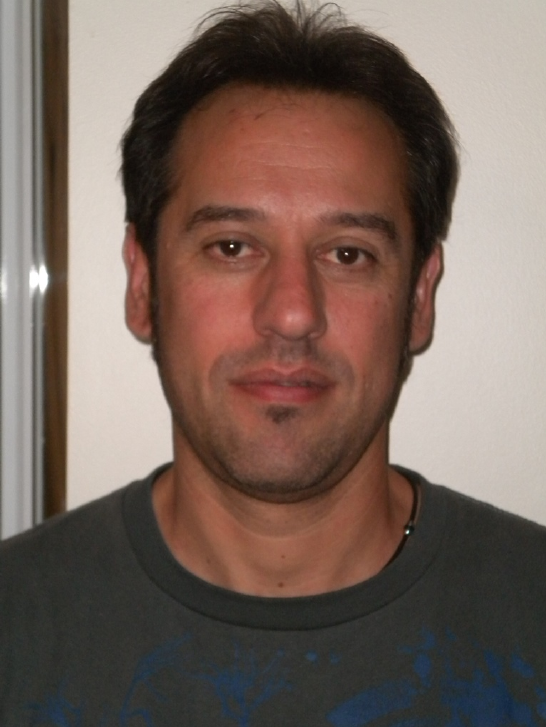}
  
  \label{fig:garcia}
\end{figure}
Dr Jose Garcia-Rodriguez received his Ph.D. degree, with specializations in Computer Vision and Neural Networks, from the University of Alicante (Spain). He is currently an Associate Professor at the Department of Computer Technology in the University of Alicante. His research areas include computer vision, computational intelligence, machine learning, pattern recognition, robotics, man-machine interfaces, ambient intelligence, computational chemistry, and parallel and multicore architectures. He has authored over 100 publications in journals and top conferences, and reviewed papers for several international journals, as well as chairing sessions in the last five editions of IJCNN and IWANN, and participating in the program committees of several international conferences.

\end{document}